\documentclass{article}

\usepackage{iclr2027_conference,times}
\usepackage{amsmath,amssymb,amsfonts,amsthm,mathtools,bm}
\usepackage{aliascnt}
\usepackage{booktabs,array,multirow,makecell}
\usepackage{graphicx,xcolor}
\usepackage{float}
\usepackage{enumitem}
\usepackage{microtype}
\usepackage{algorithm}
\usepackage{algorithmic}
\usepackage{tikz}
\usetikzlibrary{arrows.meta,calc,positioning}
\usepackage{hyperref}
\usepackage{url}
\usepackage[nameinlink,noabbrev]{cleveref}
\usepackage{needspace}
\usepackage{placeins}
\hypersetup{
  hidelinks,
  pdftitle={Protected Unbiased Sparsification: A Sharp Loewner Curve},
  pdfauthor={},
  bookmarksnumbered=true
}

\newcommand{\R}{\mathbb{R}}
\newcommand{\Q}{\mathbb{Q}}
\newcommand{\E}{\mathbb{E}}
\newcommand{\Prob}{\mathbb{P}}
\newcommand{\ones}{\mathbf{1}}
\newcommand{\supp}{\operatorname{supp}}
\newcommand{\Cov}{\operatorname{Cov}}
\newcommand{\Var}{\operatorname{Var}}
\newcommand{\diag}{\operatorname{diag}}
\newcommand{\tr}{\operatorname{tr}}
\newcommand{\rank}{\operatorname{rank}}
\newcommand{\poly}{\operatorname{poly}}
\newcommand{\OPT}{\operatorname{OPT}}
\newcommand{\MinUncut}{\operatorname{MinUncut}}
\newcommand{\uncut}{\operatorname{uncut}}
\newcommand{\simplex}{\Delta}
\newcommand{\Cp}{C_p}
\newcommand{\Cpa}{C_{p\mid A}}

\newcommand{\hKM}{h_{K,M}}
\newcommand{\Law}{\mathcal{L}}

\newcommand{\Slice}{\mathcal{F}}
\newcommand{\Ahat}{\widehat{A}}
\newcommand{\range}{\operatorname{range}}
\newcommand{\enc}{\operatorname{enc}}
\newcommand{\eps}{\varepsilon}
\newcommand{\psdle}{\preceq}

\newtheorem{theorem}{Theorem}
\newtheorem{proposition}{Proposition}
\newaliascnt{lemma}{proposition}
\newtheorem{lemma}[lemma]{Lemma}
\aliascntresetthe{lemma}
\newaliascnt{corollary}{proposition}
\newtheorem{corollary}[corollary]{Corollary}
\aliascntresetthe{corollary}
\newaliascnt{definition}{proposition}

\aliascntresetthe{definition}
\theoremstyle{remark}
\newaliascnt{remark}{proposition}
\newtheorem{remark}[remark]{Remark}
\aliascntresetthe{remark}
\crefname{theorem}{Theorem}{Theorems}
\crefname{proposition}{Proposition}{Propositions}
\crefname{lemma}{Lemma}{Lemmas}
\crefname{corollary}{Corollary}{Corollaries}
\crefname{definition}{Definition}{Definitions}
\crefname{remark}{Remark}{Remarks}
\crefname{equation}{Eq.}{Eqs.}
\crefname{figure}{Figure}{Figures}
\crefname{table}{Table}{Tables}
\crefname{section}{Section}{Sections}
\crefname{appendix}{Appendix}{Appendices}
\crefname{algorithm}{Algorithm}{Algorithms}

\title{Protected Unbiased Sparsification:\\
A Sharp Loewner Curve}

\author{Ziyang Wang$^{1}$ \quad Menghao Li$^{2}$\\
  {\normalfont\small $^{1}$Northwestern Polytechnical University, Xi'an
  \quad $^{2}$Tianjin University}}
\hypersetup{pdfauthor={Ziyang Wang and Menghao Li}}

\iclrfinalcopy
\begin{document}

\maketitle
\fancyhead{}
\fancyhead[L]{\small Protected Unbiased Sparsification}
\fancyhead[R]{\small Preprint}
\fancyfoot[C]{\thepage}

\begin{abstract}
Many learning systems replace a probability vector \(p\in\simplex_k\), such as
attention rows, data weights or expert gates, by a random vector \(Q\)
supported on at most \(m\ll k\) coordinates.  We ask that \(Q\) be unbiased,
that it preserve prescribed linear statistics \(AQ=Ap\) on every realization
(class proportions, block masses, means of known features), and that its
quality be certified for every downstream response, which is unknown when
\(Q\) is drawn.  We measure quality in Loewner order against \(\Cpa\), the
covariance of a single categorical draw after the protected statistics are
used as control variates.  With \(d\le k-2\) effective protected directions we
construct, for every budget \(d+1\le m\le k\), such a \(Q\) with
\(\Cov(Q)\psdle h\,\Cpa\), where \(h=(k-m)/((m-d)(k-d-1))\).  We prove that no
sparse protected estimator, biased or not, achieves a smaller coefficient
uniformly over instances, so in the worst case unbiasedness is free.  The
coefficient is the variance factor of drawing \(m-d\) of \(k-d\) units
without replacement, so each protected direction costs one unit of budget and
one unit of population.  VarOpt, equal-weight and ratio designs miss it by factors up
to \((k-1)/(k-m)\) or of order \(m\).  The new ingredient is a residual
covariance bridge that lifts a concurrent unprotected bound to arbitrary
pathwise protection.  Beyond the worst case, we determine the instance-optimal
constant for class protection with uniform weights inside classes and give a
factor that adapts to peaked weights.  An exact Las Vegas sampler attains the
curve within \(1+\eps\) in expected polynomial time, pivotal sampling inside
blocks is a fast sampler for block protection that is never worse than
sampling with replacement, and adapting the law to a revealed quadratic loss
is as hard as Minimum Uncut.  Experiments on synthetic weights and on GPT-2
and Qwen2.5 attention confirm the curve and the gap to classical designs.  In
decoding with Qwen2.5-7B-Instruct at \(0.4\%\) and \(1.6\%\) of a \(16\)k
context, the fast sampler keeps every block mass exact and stays
within the noise of VarOpt on RULER and on PG-19 perplexity.  Independent
draws raise the perplexity by up to \(44\%\), and top-\(m\) truncation
scores highest on RULER.

\end{abstract}

\section{Introduction}
\label{sec:intro}

Sparse surrogates of probability vectors appear throughout large-scale
learning.  Attention over a long context is restricted to a few retained keys
\citep{zhang2023h2o,xiao2024efficient}, a softmax gate over experts is truncated to its largest
entries \citep{shazeer2017outrageously}, and a training set is reweighted onto
a small coreset \citep{mirzasoleiman2020craig,killamsetty2021gradmatch}.
Deterministic truncation is biased.  A standard remedy is randomization: draw
a random probability vector \(Q\) with \(\E Q=p\) and \(|\supp(Q)|\le m\), so
that errors cancel on average over draws.  SANTA gives an unbiased post-softmax
attention estimator \citep{lee2026santa}.  MagicPIG studies unbiased oracle
sampling but implements a self-normalized estimator that is generally biased
\citep{chen2024magicpig}.  Unbiased compressors also arise in distributed
optimization \citep{alistarh2017qsgd,wangni2018gradient,wang2018atomo}.

\begin{figure*}[t]
  \centering
  \includegraphics[width=\textwidth]{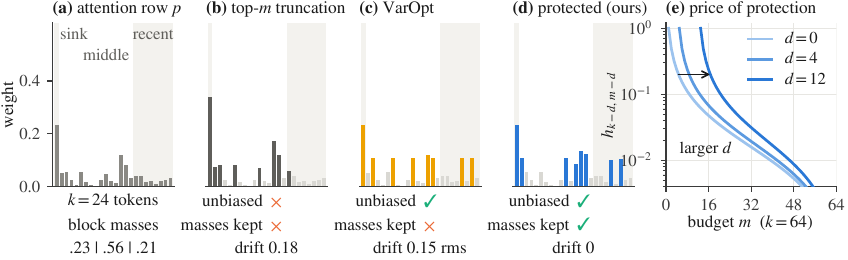}
  \caption{\textbf{Protected unbiased sparsification of a real attention row.}
  (a)~A last-position attention row of Qwen2.5-0.5B (\(k=24\)) with three
  blocks: the sink, the middle, and the eight most recent tokens.  (b) to
  (d)~Sparse weights with \(m=8\) atoms from top-\(m\) truncation, VarOpt
  and the law of \cref{thm:main} with the block masses protected (\(d=2\)),
  over the dense row (gray), with the drift of the block masses (root mean
  square over draws for VarOpt).  That law is unbiased, keeps the block masses
  on every draw, and satisfies \(\Cov(Q)\psdle h_{k-d,m-d}\Cpa\).  (e)~Each protected direction costs one
  atom of budget, so the curve \(h_{k-d,m-d}\) moves right as \(d\) grows.}
  \label{fig:hero}
\end{figure*}

We ask for three properties at once.  The law of \(Q\) should be
\emph{unbiased}, as above.  It should give \emph{exact protection}: some
linear statistics of \(p\) are known to matter and must be preserved on every
realization, not only on average.  Examples are the class proportions of a
coreset, the attention mass of each block of positions, and the means of a
few known features that serve as control variates.  We write these
constraints as \(AQ=Ap\).  Finally, its guarantee should be \emph{response
oblivious}: the sparse weights are drawn once and then consumed by responses
that are not known in advance, such as a gradient, a value projection, or a
whole training run, so the guarantee must hold for all responses
simultaneously.  \Cref{fig:hero} shows the three properties on an attention
row of a language model.

\paragraph{How to measure quality for all responses.}
For a response vector \(g\in\R^k\), the scalar \(g^\top Q\) is an unbiased
estimate of \(g^\top p\) with variance \(g^\top\Cov(Q)g\).  A bound
\(\Cov(Q)\psdle c\,B\) in Loewner order therefore bounds this variance for
every \(g\) at once, and also the error \(\E\|G(Q-p)\|_2^2\) of every
matrix-valued response \(G\).  The question is which \(B\) to use.  Exact
protection makes \(g^\top Q\) error free whenever \(g\) is a combination of
the protected statistics and the constant vector, so only the part of \(g\)
that the protected statistics cannot explain should count.  This leads to the
comparator
\begin{equation}
  \Cpa=\Cp-\Cp A^\top(A\Cp A^\top)^\dagger A\Cp,
  \qquad \Cp=\diag(p)-pp^\top,
  \label{eq:conditional-covariance}
\end{equation}
which satisfies \(g^\top\Cpa g=\min_\lambda\Var_{I\sim p}(g_I-\lambda^\top
Ae_I)\) (\cref{prop:comparator}).  In words, \(\Cpa\) is the covariance that a
single categorical draw \(e_I\), \(I\sim p\), retains after the protected
statistics are used as control variates with the best coefficients.  Every
unbiased, exactly protected law satisfies \(\Cov(Q)\psdle\Cpa\), so \(\Cpa\) is the natural
unit, and the question becomes: \emph{what fraction of \(\Cpa\) must an
unbiased, exactly protected, \(m\)-sparse law retain, uniformly over all
responses?}

\paragraph{Answer.}
Let \(d=\rank(A\Cp A^\top)\) be the number of protected directions that
actually bind on the simplex.  Redundant rows and the implicit constraint
\(\ones^\top Q=1\) do not count.  Assume \(d\le k-2\).  The answer depends on
\((k,m,d)\) only through the residual population \(K=k-d\) and the free
budget \(M=m-d\), and equals
\[
  \hKM=\frac{K-M}{M(K-1)} .
\]
This is the classical variance factor of sampling without replacement.  The mean of
\(M\) units drawn without replacement from \(K\) units with population
variance \(\sigma^2\) has variance \(\hKM\sigma^2\)
\citep{cochran1977sampling}, the \(1/M\) of independent draws times the
finite population correction \((K-M)/(K-1)\).  So
\cref{thm:main} says that, in the residual geometry, protected sparsification
can always do at least as well as sampling \(m-d\) of \(k-d\) units without
replacement, and that some instances force exactly this.  Each protected
direction costs one unit of budget and one unit of population.  The curve
equals one at the minimum feasible budget \(m=d+1\), decays like \(1/M\), and
vanishes at \(m=k\).

\paragraph{Worst case versus a given instance.}
The coefficient \(\hKM\) is a worst case over instances with the same
\((k,m,d)\).  For class-balanced reweighting with \(c\) equal classes of
uniform weights (\(A\) the class indicators, \(d=c-1\)), one independent draw
per class at \(m=c\) gives \(\Cov(Q)=\Cpa/c\), and \cref{sec:partition} shows
that the instance optimum is \((k-m)/(m(k-c))\) whenever \(c\) divides \(m\),
a factor \(m/(m-c+1)\) below the worst case.  The worst case comes from tiny
classes that consume support positions while carrying almost no mass.  Peaked
attention rows are far from it.  Keeping their heaviest
coordinates exactly gives a factor that adapts to the instance
(\cref{cor:heavy}), and pivotal sampling inside the blocks keeps every
block mass at the measured median error of VarOpt (\cref{prop:fast}).

\Needspace{5\baselineskip}
\paragraph{Contributions.}
\begin{enumerate}[leftmargin=*,itemsep=2pt,topsep=3pt]
  \item \textbf{Sharp protected Loewner curve (\cref{thm:main}).}  For every
  \(d\le k-2\) and \(d+1\le m\le k\) there is a finitely supported, unbiased,
  pathwise protected, \(m\)-sparse law with \(\Cov(Q)\psdle
  h_{k-d,m-d}\Cpa\), and no smaller coefficient holds uniformly over instances
  of effective rank \(d\).  The unprotected coefficient \(h_{k,m}\) and the
  boundary martingale appear in concurrent work \citep{zhang2026sharp}; v1
  also notes a single boundary law for all quadratic forms, whereas v2
  states the formwise bound.  The comparator of an
  intermediate state is nonlinear in the state.  A \emph{residual
  covariance bridge} (\cref{lem:bridge}),
  \(\Cov(P)+\E\,C_{P\mid A}\psdle\Cpa\), which follows from the concavity of
  the Schur complement, makes the recursion linear.  Our bound starts at
  \(m=d+1\) and improves the protected guarantee of
  \citet{haverbeck2026risk} by a factor above \(20/9\) at every common budget \(m<k\),
  on the same residual comparator obtainable from their law
  (\cref{cor:haverbeck,app:comparison}).
  \item \textbf{Unbiasedness is free in the worst case, and classical designs miss the curve
  (\cref{thm:universal,thm:designs}).}  No \(m\)-sparse protected estimator,
  biased or not, beats \(h_{k-d,m-d}\) in the worst case.  Equal-weight,
  VarOpt and ratio designs, each optimal for a criterion that ignores the
  response, lose a factor up to \((k-1)/(k-m)\) or of order \(m\).
  \item \textbf{What the comparator measures, and beyond the worst case
  (\cref{prop:comparator,prop:partition,cor:heavy}).}  We characterize
  \(\Cpa\) through control variates, determine the instance-optimal constant
  for partition constraints with uniform weights inside each class, and give a
  factor that adapts to peaked weights.
  \item \textbf{Algorithms and limits (\cref{thm:sampler,prop:fast,thm:minuncut}).}  For
  rational inputs, a Las Vegas algorithm draws an exactly protected, exactly
  unbiased \(Q\) with factor at most \((1+\eps)h_{k-d,m-d}\) in expected
  polynomial time.  For block protection, pivotal sampling inside the blocks is
  a fast sampler that is never worse than sampling with replacement, for every
  response.  If a quadratic loss
  \((Q-p)^\top W(Q-p)\) is revealed before the law is chosen, computing the
  best law as a finite list of atoms with rational probabilities admits no
  polynomial-time approximation scheme unless \(\mathsf P=\mathsf{NP}\).
\end{enumerate}
After the setup (\cref{sec:setup}) and the curve (\cref{sec:curve}),
\cref{sec:universal,sec:partition} compare it with other estimators and with
instances far from the worst case, \cref{sec:sampler,sec:hardness,sec:learning} treat
computation and learning, \cref{sec:numerics} reports experiments up to
long-context decoding, and
\cref{tab:notation} collects the notation.

\section{Problem, comparator, and examples}
\label{sec:setup}

Let \(\simplex_k=\{q\in\R_+^k:\ones^\top q=1\}\), and let
\(\simplex_k^\circ\) be its relative interior, the probability vectors whose
coordinates are all positive.  Fix \(p\in\simplex_k^\circ\) (zero coordinates
of \(p\) can be deleted first) and \(A\in\R^{a\times k}\).  A random
probability vector \(Q\) is \emph{admissible at budget \(m\)} if
\begin{equation}
  AQ=Ap\ \text{a.s.},\qquad \E Q=p,\qquad
  |\supp(Q)|\le m\ \text{a.s.}
  \label{eq:admissible}
\end{equation}
Every realization lies in the \emph{protected slice}
\(\Slice=\{q\in\simplex_k:Aq=Ap\}\), and every difference of two points of
\(\Slice\) lies in \(\ker\Ahat\), where \(\Ahat\in\R^{(a+1)\times k}\) stacks
the row \(\ones^\top\) on top of \(A\).  Rows of \(A\) that are redundant
given \(\ones^\top q=1\) do not restrict \(\Slice\) and must not be charged to
the budget.  The number of \emph{effective protected directions} is
\begin{equation}
 d=\rank(A\Cp A^\top)=\rank\Ahat-1,
 \qquad \Cp=\diag(p)-pp^\top
 \label{eq:effective-rank}
\end{equation}
(\cref{app:geometric}).  Protecting all \(c\) class totals gives \(d=c-1\),
because the last total is implied by \(\ones^\top Q=1\).  We assume
\(d\le k-2\), since \(d=k-1\) forces \(Q=p\), and \(m=d+1\) is the smallest
budget feasible on every instance, since a vertex of \(\Slice\) has at most
\(d+1\) nonzero coordinates.

\paragraph{The comparator.}
The law of \(Q\) is chosen when \((p,A,m)\) are known and before any
downstream response is revealed.  We certify all responses at once by a
Loewner bound \(\Cov(Q)\psdle c\,\Cpa\) against the comparator
\cref{eq:conditional-covariance}.  The next proposition, proved in
\cref{app:geometric}, explains this choice.

\begin{proposition}[What the comparator measures]
\label{prop:comparator}
Let \(I\sim p\) and write \(a_I=Ae_I\) for the protected statistics of a
single categorical draw.
\begin{enumerate}[label=(\alph*),leftmargin=*,itemsep=1pt,topsep=2pt]
  \item For every \(g\in\R^k\),
  \(g^\top\Cpa g=\min_{\lambda\in\R^a}\Var(g_I-\lambda^\top a_I)\).  Thus
  \(\Cpa\) is the covariance of \(e_I\) after the protected statistics are
  used as control variates with the best coefficients.
  \item For \(c\ge0\), \(\Cov(Q)\psdle c\,\Cpa\) holds if and only if
  \(\Var(g^\top Q)\le c\min_\lambda\Var(g_I-\lambda^\top a_I)\) for every
  \(g\), and if and only if \(\E\|G(Q-p)\|_2^2\le c\tr(G\Cpa G^\top)\) for
  every matrix \(G\) with \(k\) columns.
  \item Every admissible law, at any budget, satisfies
  \(\Cov(Q)\psdle\Cpa\).  At the minimum budget \(m=d+1\), the constant one
  cannot be improved uniformly over instances.
  \item \(\range(\Cpa)=\ker\Ahat\) and \(\Cpa\psdle\Cp\), with equality if and
  only if \(d=0\).  If \(g=A^\top\lambda+t\ones\), then \(g^\top Q=g^\top p\)
  on every realization and \(g^\top\Cpa g=0\).
\end{enumerate}
\end{proposition}

A bound against \(\Cp\) would charge the law for directions in which
protection already forces zero error.  For partition indicators,
\(\Cpa=\E\Cov(e_I\mid\text{class of }I)\) is the within-class covariance.

\paragraph{Examples of protected statistics.}
The user chooses \(A\) to encode statistics that must not be perturbed, and
\cref{sec:learning} treats an \(A\) that is only partly known.

\emph{Class-balanced reweighting.}  Points carry weights \(p\) and labels in
\(c\) classes, and the rows of \(A\) are the class indicators.  Every
realization keeps each class mass, \(d=c-1\), and \(\Cpa\) is the
within-class covariance.  The error of any gradient or loss computed from
\(Q\) then depends only on within-class dispersion (\cref{sec:learning}).

\emph{Attention with protected block masses and proxy features.}  Let \(p\)
be an attention row over \(k\) keys and \(V\in\R^{d_v\times k}\) the value
matrix, so that the attention output is \(Vp\).  The rows of \(A\) can be the
indicators of a few position blocks (attention sinks \citep{xiao2024efficient}, a local window, the
remote context) and \(r\) proxy features \(u_j^\top V\), the projections of
the values on fixed directions \(u_j\).  Then \(Q\) keeps every block mass,
\(VQ\) keeps every projection \(u_j^\top VQ\) exact, and
\(\E\|V(Q-p)\|_2^2\le c\,\tr(V\Cpa V^\top)\), where the trace is the residual
variance of the values after regression on the protected features.  If the \(u_j\) are the top \(r\) eigenvectors of \(V\Cp V^\top\),
this residual is at most the spectral tail of \(V\Cp V^\top\) beyond rank
\(r\) (\cref{app:geometric}), although on real attention rows such features
help little (\cref{sec:learning}).

\emph{Gates and control variates.}  The gate mass of each expert group, or the
means of features with known expectations \citep{lavenberg1981perspective},
can be protected in the same way.

\section{The sharp protected Loewner curve}
\label{sec:curve}

The main theorem makes the answer of the introduction precise.  For integers
\(K\ge2\) and \(1\le M\le K\) put
\begin{equation}
 h_{K,M}=\frac{K-M}{M(K-1)},
 \qquad K=k-d,\quad M=m-d .
 \label{eq:h-def}
\end{equation}
For a deterministic residual block, we set \(h_{1,1}=0\).

\begin{theorem}[Sharp protected sparsification curve]
\label{thm:main}
Assume \(p\in\simplex_k^\circ\), let
\(d=\rank(A\Cp A^\top)\le k-2\), and take \(d+1\le m\le k\).
There is a finitely supported admissible law satisfying
\begin{equation}
  \Cov(Q)\psdle h_{k-d,m-d}\,\Cpa
  =\frac{k-m}{(m-d)(k-d-1)}\,\Cpa .
  \label{eq:main-bound}
\end{equation}
For every feasible triple \((k,d,m)\), the coefficient in
\cref{eq:main-bound} is the supremum, over all instances \((p,A)\) of
effective protected rank \(d\), of the smallest constant \(c\) for which some
admissible law satisfies \(\Cov(Q)\psdle c\,\Cpa\).
\end{theorem}

\begin{remark}[Finite population benchmark]
\label{rem:srswor}
For \(d=0\) and uniform \(p\), simple random sampling of \(m\)-subsets has
\(\Cov(Q)=h_{k,m}\Cp\) exactly.  \Cref{thm:main} extends this benchmark to
arbitrary \(p\) and protection, with \(k\) and \(m\) shifted by \(d\).
\end{remark}

\subsection{Comparison with the closest results}
\label{sec:comparison}

\Cref{tab:prior} states the closest guarantees in common terms, at the same
total support \(m\).  \Cref{app:comparison} records the exact statements and
the versions that were checked.

\begin{table}[!htbp]
\centering
\footnotesize
\setlength{\tabcolsep}{3pt}
\renewcommand{\arraystretch}{1.1}
\caption{Closest guarantees at a common total support \(m\), as stated in the
cited works.
\(d_U=\dim U\) is the protected dimension of
\citet{haverbeck2026risk}, and \(d\le d_U\) in their
setting (\cref{cor:haverbeck}).}
\label{tab:prior}
\begin{tabular}{
 >{\raggedright\arraybackslash}p{2.05cm}
 >{\raggedright\arraybackslash}p{2.45cm}
 >{\raggedright\arraybackslash}p{2.6cm}
 >{\raggedright\arraybackslash}p{2.15cm}
 >{\raggedright\arraybackslash}p{1.7cm}
 >{\raggedright\arraybackslash}p{1.7cm}}
\toprule
Result & Output and protection & Bound at total support \(m\) & Comparator &
Least \(m\) & Construction \\
\midrule
\citet{haverbeck2026risk}, Prop.~A.3
& unbiased reweighting, \(U\)-part of the barycenter fixed pathwise
& \(\frac{20}{9(m-2d_U-2)}\), Loewner
& \((I-\Pi_U)\Sigma(I-\Pi_U)\)
& \(2d_U+3\)
& constrained randomized rounding \\
\addlinespace
\citet{zhang2026sharp}, Thm.~G.1 and Lem.~G.2 (v2)
& sparse convex combination, no protection
& \(h_{k,m}\), scalar, one quadratic form at a time
& weighted variance \(V_q\) (scalar)
& \(1\)
& existential boundary martingale \\
\addlinespace
\citet{tinarrage2026sharp}, Lem.~4
& unbiased equal-weight subset, needs \(p_i\le1/m\), no protection
& \(\frac1m(1-\|y\|^2)\), scalar
& unit-ball normalization
& \(1\)
& pivotal sampling \\
\addlinespace
\textbf{This paper}
& unbiased, arbitrary \(AQ=Ap\) pathwise
& \(h_{k-d,m-d}\le\frac1{m-d}\), Loewner, all responses at once, sharp
& \(\Cpa\), a Schur complement
& \(d+1\)
& finite law, exact-rational sampler \\
\bottomrule
\end{tabular}
\end{table}

\FloatBarrier
\Needspace{13\baselineskip}
\begin{corollary}[Comparison at a common budget]
\label{cor:haverbeck}
Let atoms \(x_1,\dots,x_k\in\R^D\) carry the weights \(p\), with covariance
\(\Sigma\) and \(X=[x_1,\dots,x_k]\), and let \(\Pi_U=UU^\top\) project onto a
subspace of dimension \(d_U\).  Protecting \(A=U^\top X\) keeps
\(U^\top\eta=0\) for the barycenter error \(\eta=X(Q-p)\), gives \(d\le d_U\),
and the law of \cref{thm:main} at total support \(m\ge d+1\) satisfies
\[
 \E[\eta\eta^\top]\psdle\frac1{m-d}\,(I-\Pi_U)\Sigma(I-\Pi_U).
\]
At the same total support, Proposition~A.3 of \citet{haverbeck2026risk}
requires \(m\ge2d_U+3\) and gives the factor \(20/(9(m-2d_U-2))\), larger by at
least \(20(m-d_U)/(9(m-2d_U-2))>20/9\).
\end{corollary}

The proof (\cref{app:geometric}) uses the residual comparator
\(X\Cpa X^\top\).  Residualizing the all-response bound in
Proposition~A.3 of \citet{haverbeck2026risk} yields the same comparator for
its law, so the improvement is in the coefficient and support budget.
Version~1 of \citet{zhang2026sharp} also notes a uniform boundary law in the
unprotected case; \cref{lem:isotropic} treats arbitrary pathwise protection
(\cref{app:comparison}).

\subsection{How the curve is attained}
\label{sec:overview}

\emph{Construction.}  Start at \(p\).  While the current state \(q\) has more
than \(m\) nonzero coordinates, move it at random, with conditional mean
\(q\), to the boundary of the protected slice restricted to its support, so
that at least one coordinate becomes zero.  Every move keeps
\(AQ=Ap\) and the mean, so the terminal state is admissible.  Two obstacles
remain, and each is removed by one lemma.

\emph{Obstacle 1: one step must be small in every direction at once.}  A
move is a law on the boundary of a polytope, and its covariance must be at
most \(\alpha\,C_{q\mid A}\) in Loewner order, that is, for all responses.

\begin{lemma}[Isotropic boundary contraction]
\label{lem:isotropic}
Let \(p_i>0\), \(\sum_i p_i=1\), \(\sum_i p_i a_i=0\), and
\(\sum_i p_i a_i a_i^\top=I_r\).  For the polytope
\(\mathcal K=\{x\in\R^r:1+a_i^\top x\ge0\ \forall i\}\), there is a finitely
supported random \(X\in\partial\mathcal K\) such that
\begin{equation}
  \E X=0,
  \qquad \Cov(X)\psdle \frac1{r^2}I_r.
  \label{eq:isotropic-boundary}
\end{equation}
\end{lemma}

\begin{proof}[Proof sketch]
For a positive definite test matrix \(H\), shoot a ray from the origin in
each direction \(-H^{-1}a_i\) to \(\partial\mathcal K\) and weight its
endpoint in proportion to \(p_i\) over its length (\cref{fig:radial}).
Isotropy centers this law, and the facet inequalities with the
arithmetic-harmonic mean inequality give \(\tr(H\Cov X)\le\tr(H)/r^2\).  The
payoff is bilinear in the law and in \(H\), so Sion's minimax theorem
\citep{sion1958general} gives one law that is good for every \(H\), and
replacing it on each facet by its barycenter leaves one atom per facet.
\end{proof}

At a positive state \(q\), let \(r=|\supp q|-1-\rank(AC_qA^\top)\) be the
slice dimension.  In suitable coordinates the protected slice restricted to
the support of \(q\) is the polytope \(\mathcal K\) of the lemma, and
\(C_{q\mid A}\) is the image of \(I_r\) (\cref{app:geometric}).  The lemma
therefore yields a mean-preserving protected move \(q\mapsto Q'\) that drops
at least one coordinate and obeys
\begin{equation}
 \Cov(Q'\mid q)\psdle \frac1{r^2}\,C_{q\mid A}.
 \label{eq:one-step}
\end{equation}

\emph{Obstacle 2: the comparator moves with the state.}  Step bounds are
relative to the random comparator \(C_{P_t\mid A}\) and cannot simply be
added.  The next lemma is a conservation law: the covariance already spent
plus what the state can still carry never exceeds the residual covariance at
\(p\).

\begin{lemma}[Residual covariance bridge]
\label{lem:bridge}
If a random \(P\in\simplex_k\) satisfies \(\E P=p\) and \(AP=Ap\) almost
surely, then
\begin{equation}
  \Cov(P)+\E\, C_{P\mid A}\psdle \Cpa .
  \label{eq:bridge}
\end{equation}
\end{lemma}

\begin{proof}
For positive semidefinite \(C\), the map
\(S(C)=C-CA^\top(ACA^\top)^\dagger AC\) is the Loewner minimum over \(L\) of
\((I-LA)C(I-LA)^\top\), so it is concave.  Unbiasedness gives
\(\E\,C_P=\Cp-\Cov(P)\), and pathwise protection gives \(\Cov(P)A^\top=0\), so
\(S(\Cp-\Cov(P))=\Cpa-\Cov(P)\).  Jensen's inequality for \(S\) gives
\(\E\,C_{P\mid A}=\E\,S(C_P)\psdle S(\E\,C_P)=\Cpa-\Cov(P)\).
\end{proof}

\emph{Assembling the recurrence.}  The two lemmas give a linear recursion
for \(B_t=\Cov(P_t)\).

\begin{proof}[Proof of \cref{thm:main}]
Start with \(P_k=p\).  For \(t=k,k-1,\ldots,m+1\), leave every state of
support at most \(t-1\) unchanged and apply \cref{eq:one-step} to every state
of support exactly \(t\).  Since \(\rank(AC_qA^\top)\le d\), the slice
dimension at such a state is at least \(t-d-1\ge1\), so with \(\alpha_t=(t-d-1)^{-2}\) every
branch satisfies \(\E[P_{t-1}\mid P_t]=P_t\) and
\(\Cov(P_{t-1}\mid P_t)\psdle \alpha_t C_{P_t\mid A}\).  Let
\(B_t=\Cov(P_t)\).  The law of total covariance and \cref{lem:bridge} give
\begin{equation}
 B_{t-1}\psdle \alpha_t\E C_{P_t\mid A}+B_t
 \psdle \alpha_t\Cpa+(1-\alpha_t)B_t .
 \label{eq:recurrence}
\end{equation}
Starting from \(B_k=0\), the scalar recurrence
\(\eta_{t-1}=\alpha_t+(1-\alpha_t)\eta_t\) has the closed form
\(\eta_t=(k-t)/((t-d)(k-d-1))\) (\cref{app:geometric}), so
\(B_m\psdle h_{k-d,m-d}\Cpa\).  The state \(P_m\) has support at most \(m\),
exact mean \(p\), and exact protection, and its law is finite because every
one-step law is.

\emph{Sharpness.}  Put mass \(\delta\) on \(d\) anchor coordinates, protect
each of them, and spread \(1-\delta\) uniformly on the other \(K\).  Every
realization keeps the anchors, so the free block is unprotected sampling of
\(M\) out of \(K\) uniform units.  There every unbiased \(M\)-sparse law has
\(\E\|Q-u\|_2^2\ge1/M-1/K\) for the uniform \(u\), which forces
\(c\ge(1-\delta)h_{K,M}\) (\cref{app:geometric}), and \(\delta\downarrow0\)
matches the upper bound.  This instance is a partition with \(d\) tiny
classes (\cref{sec:partition}).
\end{proof}

\section{Unbiasedness is free in the worst case, and classical designs miss the curve}
\label{sec:universal}

\Cref{thm:main} fixes the best constant among unbiased protected laws.  Can a
biased estimator do better, and does a classical design attain the curve?
Both answers are negative.

\begin{theorem}[The curve bounds every sparse protected estimator]
\label{thm:universal}
For every feasible \((k,d,m)\) and every \(c<h_{k-d,m-d}\) there is an
instance \((p,A)\) of effective protected rank \(d\) such that no random
\(m\)-sparse probability vector \(Q\) with \(AQ=Ap\) almost surely, unbiased
or not, satisfies \(\E[(Q-p)(Q-p)^\top]\psdle c\,\Cpa\).
\end{theorem}

Since the law of \cref{thm:main} is unbiased, it attains the minimax value of
the uniform guarantee among all \(m\)-sparse protected estimators, and in the
worst case unbiasedness costs nothing.  On a given
instance a biased estimator can still do better, as truncation does on short
attention rows (\cref{app:llm-64}).  On the extremal
instances every sparse realization has the same inner product with \(p\) on
the free block, so a bias cannot help (\cref{app:universal}).

For \(d=0\) write \(\lambda(Q)\) for the smallest \(c\) with
\(\Cov(Q)\psdle c\,\Cp\).  The designs below are unbiased, keep
\(\ones^\top Q=1\) on every realization, and are optimal for criteria that do
not look at the response.

\begin{theorem}[Classical designs miss the curve]
\label{thm:designs}
\begin{enumerate}[label=(\roman*),leftmargin=*,itemsep=1pt,topsep=2pt]
  \item \emph{Equal weights.}  If \(Q\) puts weight \(1/m\) on a random
  \(m\)-subset with inclusion probabilities \(mp_i\), then
  \(\lambda(Q)\ge h_{k,m}\); if \(m>1\), equality implies that \(p\) is uniform.
  \item \emph{VarOpt.}  For \(m<k\), if \(m-1\) coordinates have mass \(1/m\) and the
  other \(k-m+1\) have equal mass, every VarOpt design
  \citep{cohen2011efficient} and every equal-weight design has
  \(\lambda(Q)=1/m=\frac{k-1}{k-m}\,h_{k,m}\), as for independent sampling.
  \item \emph{Ratio design.}  The design of \citet{midzuno1951sampling} draws
  \(S\) with probability proportional to \(p(S)\) and sets
  \(Q=p|_S/p(S)\).  With one coordinate of mass \(1/2\) and \(k-1\) equal
  coordinates, \(\lambda(Q)\ge\frac{m(k-1)}{k+m-2}\,h_{k,m}\).
\end{enumerate}
At \(m=1\), the equal-weight law is categorical sampling and has
\(\Cov(Q)=\Cp\), hence \(\lambda(Q)=h_{k,1}=1\), for every \(p\).
At \(m=k\), VarOpt and the ratio design return \(Q=p\), so
\(\lambda(Q)=0\).  The full-subset equal-weight design is admissible only
for uniform \(p\), when it also returns \(p\).
\end{theorem}

Each design spends its randomness in directions that the uniform guarantee
penalizes.  VarOpt minimizes \(\sum_i\Var(Q_i)\) and puts all randomness
on a few light coordinates, whose contrasts then carry relative error
\(1/m\), the ratio design drops heavy coordinates too often, and independent
sampling forfeits the finite population correction.  The gap also appears on
realistic softmax weights (\cref{fig:numerics}(c)), and it closes when a few
coordinates carry almost all the mass.

\section{Beyond the worst case: partitions and peaked weights}
\label{sec:partition}

\Cref{thm:main} is uniform over instances, and its extremal instances have
protected classes of vanishing mass.  Balanced partitions and peaked weights
are far from them.  For a fixed instance, the \emph{instance-optimal
constant} \(\gamma^*(p,A,m)\) is the infimum of the \(c\) for which some
admissible law at budget \(m\) has \(\Cov(Q)\psdle c\,\Cpa\).  It can lie strictly below
\(h_{k-d,m-d}\), and for general \(A\) we do not know how to compute it
(\cref{rem:pointwise}).  For partitions it can be pinned down, and for peaked
weights it can be bounded by keeping the heaviest coordinates exactly.

\paragraph{Partitions.}
Let \([k]\) be partitioned into classes \(B_1,\dots,B_c\) with masses
\(\pi_j=p(B_j)\), sizes \(k_j=|B_j|\), and within-class weights
\(p^{(j)}=p|_{B_j}/\pi_j\), and let the rows of \(A\) be the class
indicators, so that \(d=c-1\).  A class that receives the budget \(x\) pays
\[
 u_j(x)=\pi_j\,\frac{\psi(x)-1/k_j}{1-1/k_j},
\]
where \(\psi\) interpolates \(s\mapsto1/s\) linearly between
integers, so \(u_j(x)=\pi_jh_{k_j,x}\) at integers \(x\) (and
\(u_j=0\) if \(k_j=1\)).  Let \(X_m=\{x\in\R^c:1\le x_j\le k_j,\
\sum_jx_j\le m\}\) be the fractional budgets.

\begin{proposition}[Partition protection]
\label{prop:partition}
For every partition instance and every \(c\le m\le k\),
\[
 \gamma^*(p,A,m)\le
 \min\Bigl\{h_{k-c+1,m-c+1},\ \min_{x\in X_m}\max_ju_j(x_j)\Bigr\}.
\]
The second term is attained by the \emph{block law}: round an optimal \(x\)
to integers \(s_j\in\{\lfloor x_j\rfloor,\lceil x_j\rceil\}\) with
\(\E s_j=x_j\) and \(\sum_js_j\le m\) on every realization, then sample the
classes independently, returning \(p^{(j)}\) if \(s_j=k_j\) and otherwise using
the unprotected law of \cref{thm:main} at budget \(s_j\).
The same minimax value with \(1/k_j\) replaced by
\(\|p^{(j)}\|_2^2\) is a lower bound on \(\gamma^*\), so if every \(p^{(j)}\)
is uniform, \(\gamma^*=\min_{x\in X_m}\max_ju_j(x_j)\).
\end{proposition}

The lower bound compares traces class by class, using
\(\|q\|_2^2\ge1/s\) for every \(s\)-sparse probability vector, and the upper
bound computes the block covariance directly (\cref{app:partition}).

\begin{corollary}[Equal classes]
\label{cor:equal-classes}
If the \(c\) classes have equal size \(n=k/c\) and \(p\) is uniform, then
\(\gamma^*=\frac1c\,\frac{n\psi(m/c)-1}{n-1}\).  If \(c\) divides \(m\),
\[
 \gamma^*=\frac{k-m}{m(k-c)}=\frac{m-c+1}{m}\,h_{k-c+1,m-c+1},
\]
and at \(m=c\) one independent draw per class gives \(\gamma^*=1/c\).
\end{corollary}

\paragraph{Peaked weights.}
Keeping a coordinate exactly spends one atom on it and removes its variance,
and the block argument applies to the singletons and the rest
(\cref{app:partition}).

\begin{corollary}[Heavy coordinates as singleton classes]
\label{cor:heavy}
Let \(0\le j<k\), and let \(L\) be the complement of a set of \(j\) coordinates, with mass
\(\pi_L\), weights \(p_L=p|_L/\pi_L\), protected rows \(A_L\) restricted to
\(L\), and \(d_L=\rank(A_LC_{p_L}A_L^\top)\).  For \(j+d_L+1\le m\le k\),
take \(Q=p\) when \(m=k\).  Otherwise \(d_L\le m-j-1\le k-j-2\), so keep
the \(j\) coordinates exactly and draw the rest from the law of
\cref{thm:main} for \((p_L,A_L)\) at budget \(m-j\).
In both cases the resulting law is admissible and satisfies
\[
 \Cov(Q)\psdle\pi_L\,h_{k-j-d_L,\,m-j-d_L}\,\Cpa .
\]
\end{corollary}

On peaked weights \(\pi_L\) is small, and \(j\) can be chosen
from \(p\) and \(A\) to minimize the factor.

\paragraph{Reading the results.}
The extremal instances of \cref{thm:main} are partitions into \(d\)
singletons of vanishing mass and one uniform class, and on balanced classes
protection is almost free, since \(m/(m-c+1)\to1\) as \(m\) grows.  The
block law is stratified sampling, and its allocation, which equalizes the
\(u_j(x_j)\), is minimax over all responses, whereas Neyman allocation
\citep{neyman1934two} is optimal for one known response.  The argument extends to rows of \(A\) supported inside single
classes (\cref{prop:block}).

\begin{remark}[Beyond partitions]
\label{rem:pointwise}
For general \(A\) the gap can also be strict: with \(k=3\), uniform \(p\),
\(A=(0,1,3)\) and \(m=2\) the only admissible law has factor \(7/9\) while
\(h_{2,1}=1\) (\cref{app:partition}).
\end{remark}

\section{Exact and fast sampling}
\label{sec:sampler}

\Cref{thm:main} is an existence result over the reals, and each of its steps
solves a minimax problem.  For rational inputs the same curve is realized by an
algorithm that uses only rational arithmetic and fair coin flips, with exact
mean and exact protection.

\begin{theorem}[Exact-rational random-path sampler]
\label{thm:sampler}
Let \(p\in\simplex_k^\circ\cap\Q^k\) and \(A\in\Q^{a\times k}\) have binary
encoding length \(L\), let \(d\le k-2\) and \(d+1\le m\le k\), and let \(0<\eps\le1\) be
rational with binary length \(\enc(\eps)\).  \Cref{alg:path} outputs a
rational \(Q\) satisfying \cref{eq:admissible} exactly and
\begin{equation}
  \Cov(Q)\psdle (1+\eps)\,h_{k-d,m-d}\,\Cpa .
  \label{eq:sampler-bound}
\end{equation}
One draw takes expected bit time \(\poly(k,a,L+\enc(\eps))\), and every
realized output has polynomial encoding length.
\end{theorem}

\Cref{alg:path} (\cref{app:sampler}) follows one root-to-leaf path of the
martingale and never lists the terminal law.  Each step finds an approximate
minimax law by an ellipsoid method over test matrices with radial
best responses, draws from it with fair bits, and rounds on a mesh to bound
the bit length.

\paragraph{A fast sampler for block protection.}
The minimax steps are slow, about \(1.5\) seconds per draw at \(k=64\) in our
implementation.  For block masses a classical design run inside the blocks
already gives a Loewner guarantee.  Pivotal
sampling \citep{deville1998splitting}, which is the dependent rounding of
\citet{srinivasan2001distributions}, draws a set of fixed size with prescribed
inclusion probabilities by letting two undecided units compete until one of
them is decided, in linear time.

\begin{proposition}[Fast sampler for block protection]
\label{prop:fast}
Let \(A\) be the indicators of classes \(B_1,\dots,B_c\), keep a set \(H\) of
coordinates exactly, let \(L_b=B_b\setminus H\) have mass \(\pi_b\), and let
integers \(1\le s_b<|L_b|\) satisfy \(|H|+\sum_bs_b\le m\) and
\(s_bp_i\le\pi_b\) on \(L_b\).  Drawing in each class, independently, \(s_b\)
coordinates of \(L_b\) by pivotal sampling in a random order with inclusion
probabilities \(s_bp_i/\pi_b\), each with weight \(\pi_b/s_b\), gives an
admissible law with
\[
 \Cov(Q)\psdle \max_b\frac{\pi_b}{s_b}\,\Cpa .
\]
\end{proposition}

Inside a class, \(s_b\) independent draws with probabilities \(p_i/\pi_b\)
and weights \(\pi_b/s_b\) have covariance exactly \(\pi_b/s_b\) times the
within-class covariance of \(L_b\), and pivotal sampling is never worse than these draws for
any response \citep{chauvet2017comparison}.  \Cref{app:partition} gives a
self-contained proof.  With budgets proportional to the light masses the
factor is close to \(\pi_L/(m-|H|)\), and on attention rows it is a median
\(0.94\) to \(1.36\) times the factor of \cref{cor:heavy}
(\cref{app:llm-fast}).  A draw takes \(O(k)\) time, about \(0.6\)
milliseconds at \(k=512\), after a setup that depends only on \(p\) and \(A\).
Our setup puts in \(H\) the \(j\) largest coordinates and any coordinate
whose inclusion probability would exceed one, keeps the \(j<m\) with the
smallest factor, and needs a median of \(53\) milliseconds at \(k=512\) and
\(m=128\).

\section{The query-aware computational boundary}
\label{sec:hardness}

The universal law of \cref{thm:main} is chosen before any response is known.
Suppose instead that a quadratic loss of the error,
\((Q-p)^\top W(Q-p)\) with \(W\succeq0\), is revealed before the law is
chosen.  We call \(W\) a \emph{quadratic query}, and its query-aware value
\(\OPT_W(p,A,m)=\inf_{\Law}\E_{\Law}(Q-p)^\top W(Q-p)\) is the infimum over
admissible finite laws.  An \emph{explicitly represented
law} is a finite list of admissible atoms with rational probabilities and
barycenter exactly \(p\).  The universal guarantee has the order of
quantifiers \(\exists\Law\ \forall W\), whereas \(\OPT_W\) allows a different
law for each \(W\).  The next theorem shows that
adapting to \(W\) is hard.  The sharpness of \cref{thm:main} does not rely
on this hardness.

\begin{theorem}[Exact Minimum-Uncut embedding]
\label{thm:minuncut}
Given a graph \(G=(V,E)\) on \(n\) vertices, one can construct in polynomial
time an instance with \(k=2n\), uniform \(p\), pair-mass protection \(A\) of
effective rank \(d=n-1\), budget \(m=n=d+1\), and an integer matrix
\(W_G\succeq0\) such that
\begin{equation}
 \OPT_{W_G}(p,A,n)=\frac{4\MinUncut(G)}{n^2}.
 \label{eq:minuncut-identity}
\end{equation}
For cubic \(G\), \(W_G\) has \(O(n)\) nonzeros and
\(\|W_G\|_{\mathrm{op}}\le12\).  Consequently, unless
\(\mathsf P=\mathsf{NP}\), there is no polynomial-time approximation scheme
for the query-aware problem whose output is an explicitly represented law.
\end{theorem}

\emph{Proof idea.}  Protecting the masses of \(n\) pairs makes every atom a
sign vector \(z\), with cost \(4\uncut_G(z)/n^2\) under the signless
Laplacian \(W_G\), and Minimum Uncut is APX-hard on cubic graphs
\citep{cilibrasi2007haplotyping,alimonti2000some} (\cref{app:hardness}).

\section{Learning consequences and scope}
\label{sec:learning}

\paragraph{Strongly convex ERM on the sparse weights.}
Let \(F_w(\theta)=\sum_iw_if_i(\theta)\) have minimizer \(\theta_w\), suppose
every realized weighted objective \(F_q\) is \(\mu\)-strongly convex, and let
\(F_p\) be \(\beta\)-smooth.  With
\(G_*=[\nabla f_1(\theta_p),\ldots,\nabla f_k(\theta_p)]\), the excess risk
is \(\E[F_p(\theta_Q)-F_p(\theta_p)]\le\beta h_{k-d,m-d}
\tr(G_*\Cpa G_*^\top)/(2\mu^2)\) (\cref{app:learning}, also for gradient
descent and quadratic tasks).  Under class balance
\(G_*\Cpa G_*^\top\) is the within-class covariance of the gradients, so exact
balance removes the between-class dispersion from the excess risk.

\paragraph{Choosing \(A\).}
Each protected direction costs one atom, so it pays only if the variance it
removes outweighs the atom.  Statistics that must be exact belong in \(A\),
and heavy coordinates are best kept exactly (\cref{cor:heavy}).  On
attention, protecting the block masses with the fast sampler costs nothing in
median error against VarOpt (\cref{sec:numerics}), whereas proxy features
help little.  On
Qwen2.5-0.5B rows of length \(64\), four value directions lower the median
value error only from \(1.15\) to \(1.07\) times that of VarOpt, and
directions from other text raise it (\cref{app:llm-tails}).

\paragraph{Unknown or approximate protection.}
A positive integer upper bound on \(d\) alone, without knowledge of the protected
directions, does not allow exact protection below \(m=k\), since protecting
every possible row forces \(Q=p\); when \(d=0\) is known, protection is
automatic.  If the protected statistics lie in the row space of a known
matrix \(S\), protecting \(S\) uses the larger rank \(d_S\ge d\) in the curve,
provided \(d_S+1\le m\le k\), with \(Q=p\) at \(m=k\).
Protecting some rows \(A'\) exactly keeps a
mean-square guarantee for all others, which interpolates between no protection
and full protection (\cref{prop:unknown-A}).  A pathwise tolerance
\(\|A(Q-p)\|\le\eps\) releases the extremal instances, and the curve under a
tolerance is open.

\section{Numerical illustration}
\label{sec:numerics}

\begin{figure}[t]
  \centering
  \includegraphics[width=\textwidth]{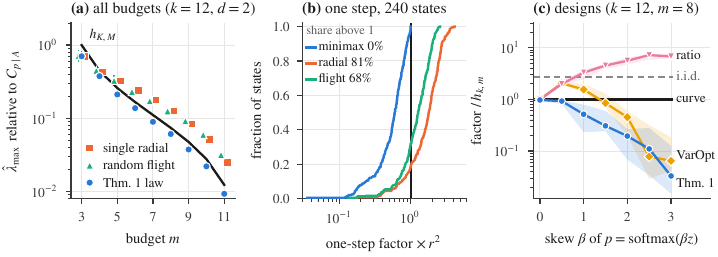}
  \caption{\textbf{The curve, the one-step lemma, and classical designs.}
  (a)~A protected instance with \(k=12\), \(d=2\): three exactly unbiased and
  protected step rules over all budgets (\(90\%\) bootstrap bars).  Only the minimax step of \cref{lem:isotropic}
  stays below \(h_{K,M}\).  (b)~Empirical distribution of the exact one-step
  factors, scaled by \(r^2\), at \(240\) random states.  The minimax step meets the bound of
  \cref{lem:isotropic} at every state.  (c)~Factor relative to
  \(h_{k,m}\) on weights \(p=\mathrm{softmax}(\beta z)\) with \(k=12\),
  \(m=8\), with medians and interquartile bands (\cref{app:numerics-revision}).}
  \label{fig:numerics}
\end{figure}

The experiments run the real-arithmetic construction of \cref{thm:main},
with the minimax step computed by matrix multiplicative weights
\citep{arora2016combinatorial} over the radial responses of
\cref{lem:rational-radial} and every one-step factor checked exactly.  All
runs follow protocols fixed in advance (\cref{app:numerics-revision,app:llm}).
Each experiment tests one claim.  \Cref{fig:numerics} tests the curve of
\cref{thm:main}, the one-step bound of \cref{lem:isotropic} and the gap of
\cref{thm:designs}.  The attention rows measure how far real weights are from
the worst case and what \cref{cor:heavy,prop:fast} deliver there, and decoding
measures the downstream cost of protection.

\paragraph{The curve and the minimax step.}
In \cref{fig:numerics}(a) our law lies below the curve at
every budget, while a single radial response and a cube-style random flight
\citep{deville2004cube} exceed it from \(m=5\) on, by up to a factor \(2.6\).  \Cref{fig:numerics}(b) shows why: the
minimax step has factor at most \(1.0001\), whereas the radial step
exceeds the bound of \cref{lem:isotropic} at \(81\%\) of states and the
flight at \(68\%\).

\paragraph{Classical designs.}
In \cref{fig:numerics}(c), as \cref{thm:designs} predicts, VarOpt rises above
the curve on mildly skewed weights, to a median of \(2.05\) at
\(\beta=0.5\), and the ratio design to \(7.2\), while our law does not.

\paragraph{Attention of a language model.}
On \(2688\) Qwen2.5-0.5B rows of length \(512\) (\cref{app:llm}), the
worst-case certificate, the bound of \cref{thm:main} on the value error with
the block masses and four value directions protected, is a median \(28\)
times the actual VarOpt error at \(m=64\).  Real attention is thus far from
the worst case, and keeping the heaviest tokens exactly (\cref{cor:heavy})
lowers the median certificate-to-VarOpt-error ratio to \(1.37\) on the
measured value response.  The certificate itself holds for every response
simultaneously.  At length \(64\), a random flight with the same protection
has about twice the measured error of the law of \cref{cor:heavy}; its
estimated trace ratio exceeds one on \(90\%\) of the cases, providing
empirical evidence that it can exceed the certified factor.  The fast sampler of
\cref{prop:fast} keeps every block mass exact with the median error of
VarOpt, lower than that of the minimax law, and at \(k=512\) its 99th
percentile stays below twice the VarOpt error, whereas VarOpt run inside the
blocks exceeds \(50\) times it (\cref{app:llm-fast}).

\paragraph{Downstream decoding.}
With Qwen2.5-7B-Instruct on prompts of about \(16\)k tokens and every attention
row replaced at decoding time by \(m=64\) or \(256\) sparse weights
(\(0.4\%\) or \(1.6\%\)), the fast sampler keeps every block mass exact and
stays within the noise of VarOpt on six RULER tasks and on PG-19 perplexity.
Independent draws raise the perplexity by \(44\%\) at \(m=64\), against at
most \(5.4\%\) for top-\(m\) truncation and the designs without replacement.
Truncation scores highest on RULER, so on these tasks unbiasedness does not
help (\cref{app:llm-downstream}, where two of six predictions failed).

\section{Related work and conclusion}
\label{sec:related}

Balanced sampling \citep{deville2004cube,tille2025balanced} and dependent
rounding \citep{srinivasan2001distributions,bansal2019generalization} keep
linear constraints along a martingale but do not attain the protected
Loewner curve (\cref{sec:comparison,app:related}), although inside blocks
pivotal sampling satisfies the with-replacement bound of \cref{prop:fast}.
Sampled attention \citep{chen2024magicpig,desai2026vattention,lee2026santa}
controls the error of one output at a time and does not keep block masses
(\cref{app:related}).

For unbiased, exactly protected, response-oblivious sparsification, the answer
is the variance factor of sampling \(m-d\) of \(k-d\) units without replacement.  An explicit law
attains it, no sparse protected estimator beats it in the worst case,
classical designs miss it, and an exact sampler realizes it.  Partitions and
peaked weights lie far below the worst case.  In long-context decoding, the
fast sampler keeps every block mass and performs within the noise of VarOpt,
although truncation scores higher on RULER.  Instance optima for general \(A\), lower
bounds for peaked weights, and the tasks on which unbiasedness pays
downstream, such as reweighted training, remain open.

\pdfbookmark[1]{Ethics statement}{ethics}
\section*{Ethics statement}

This is a theoretical paper on randomized sparsification of probability
vectors.  The numerical illustrations use three publicly released
language models (GPT-2 small, Qwen2.5-0.5B and Qwen2.5-7B-Instruct) on public
text (WikiText-2, PG-19 and fixed English sentences) and on the synthetic
tasks of RULER.  No human-subject data or personal
data are used, and we foresee no direct negative societal impact.  The
exact-protection guarantee could be used to enforce group-mass or
class-proportion constraints in data reweighting.  Whether a particular set
of protected statistics is appropriate is an application-level decision
that the mathematics does not settle.

\pdfbookmark[2]{AI use statement}{ai-use}
\subsection*{AI use statement}

Large language models assisted with writing and editing, literature search,
checking derivations, and coding the experiments and figures.  The authors
verified all content and take full responsibility for it.

\pdfbookmark[1]{Reproducibility statement}{reproducibility}
\section*{Reproducibility statement}

The theoretical results are proved in
\cref{app:geometric,app:universal,app:partition,app:sampler,app:hardness,app:learning,app:boundaries}.
The experiments of \cref{sec:numerics} and
\cref{app:numerics-large,app:numerics-revision,app:llm} follow seven
protocols written before the runs, and \cref{app:llm} marks every analysis
defined after a run and the one prediction added after a pilot.  The scripts
(NumPy and SciPy, and PyTorch with the public GPT-2 small,
Qwen2.5-0.5B and Qwen2.5-7B-Instruct checkpoints), the protocols, the fixed
seeds and the saved evaluation outputs accompany this preprint as arXiv
ancillary files.  The release README separates CPU checks of saved results
from model-based reruns, lists the figure and table commands, and identifies
omitted inputs that can be regenerated.  Reported experiment runtimes on
the original hardware are about three hours on eight CPU cores, plus about
three and a half hours on one A100 GPU for protocol 7; these are not the
runtimes of the saved-result checks.

\pdfbookmark[1]{References}{references}
\bibliography{references}
\bibliographystyle{iclr2027_conference}

\clearpage
\appendix
\crefalias{section}{appendix}
\crefalias{subsection}{appendix}
\section{Notation and terminology}
\label{app:notation}

\begin{table}[h]
\centering
\footnotesize
\setlength{\tabcolsep}{4pt}
\renewcommand{\arraystretch}{1.12}
\caption{Notation and terms used throughout.}
\label{tab:notation}
\begin{tabular}{@{}>{\raggedright\arraybackslash}p{3.1cm}>{\raggedright\arraybackslash}p{10.2cm}@{}}
\toprule
Symbol or term & Meaning \\
\midrule
\(\simplex_k\), \(\simplex_k^\circ\) & probability vectors on \(k\) coordinates, and those with all coordinates positive \\
\(A\in\R^{a\times k}\), \(\Ahat\) & protected statistics, and \(A\) with the row \(\ones^\top\) stacked on top \\
admissible law at budget \(m\) & law of a random \(Q\) with \(AQ=Ap\) a.s., \(\E Q=p\), \(|\supp Q|\le m\) a.s.\ (\cref{eq:admissible}) \\
protected slice \(\Slice\) & \(\{q\in\simplex_k:Aq=Ap\}\), the set of probability vectors with the same protected statistics as \(p\) \\
\(d\) & \(\rank(A\Cp A^\top)=\rank\Ahat-1\), the number of effective protected directions \\
\(K\), \(M\) & residual population \(k-d\) and free budget \(m-d\) \\
\(\Cp\) & \(\diag(p)-pp^\top\), covariance of one categorical draw \(e_I\), \(I\sim p\) \\
\(\Cpa\) & residual comparator \cref{eq:conditional-covariance}, covariance of \(e_I\) after regression on \(Ae_I\) (\cref{prop:comparator}) \\
\(C_{q\mid A}\) & the same object at an intermediate state \(q\), computed on the support of \(q\) \\
\(h_{K,M}\) & \((K-M)/(M(K-1))\), the curve \cref{eq:h-def} \\
\(\psdle\) & Loewner order, \(X\psdle Y\) if and only if \(Y-X\) is positive semidefinite \\
response-oblivious & the law is fixed before any downstream response is revealed \\
slice dimension \(r\) at \(q\) & \(|\supp q|-1-\rank(AC_qA^\top)\), the dimension of the protected slice restricted to the support of \(q\) \\
radial best response & the explicit law of \cref{lem:rational-radial}, which moves along rays to the boundary, for one test matrix \(H\) \\
minimax step & the one-step law of \cref{lem:isotropic}, valid for all test matrices at once \\
quadratic query \(W\) & a loss \((Q-p)^\top W(Q-p)\) revealed before the law is chosen (\cref{sec:hardness}) \\
explicitly represented law & a finite list of admissible atoms with rational probabilities and exact barycenter \(p\) \\
\(\gamma^*(p,A,m)\) & instance-optimal Loewner constant (\cref{sec:partition}) \\
\bottomrule
\end{tabular}
\end{table}

\section{Theorem-level comparison with the closest results}
\label{app:comparison}

This appendix records what the statements in \cref{tab:prior} prove, in their
own notation, and the versions that were checked.

\paragraph{\citet{haverbeck2026risk}, Proposition A.3 (arXiv v1, 1 July
2026).}  For a finitely supported probability measure \(\mu=\sum_jp_j
\delta_{x_j}\) on a Hilbert space, a subspace \(U\) of dimension \(d_U\) and
an integer \(m'\ge1\), a randomized constrained rounding returns
\(\hat\mu=\sum_j\hat p_j\delta_{x_j}\) supported on at most \(m'+2d_U+2\)
atoms almost surely, with \(\int x\,d(\hat\mu-\mu)\in U^\perp\) almost
surely, \(\E\hat\mu=\mu\), and
\(\E[\eta\otimes\eta]\psdle\frac{20}{9m'}(I-\Pi_U)\Sigma(I-\Pi_U)\) for the
barycenter error \(\eta\) and the covariance \(\Sigma\) of \(\mu\).  The
statement writes \(m\) for our \(m'\) and \(d\) for our \(d_U\).  The protection
is pathwise, unbiasedness and the covariance bound hold in expectation, and
no sharpness of the constant is claimed.  The proposition also bounds the
variance of every scalar response under the same output law.  For
\(Q=(\hat p_j)_j\) and the corresponding protection matrix \(A\), substitute
\(y_j=g_j-\lambda^\top Ae_j\) and minimize over \(\lambda\), using
\(A(Q-p)=0\), to obtain \(\Cov(Q)\psdle\frac{20}{9m'}\Cpa\).
The improvement in \cref{cor:haverbeck} concerns the coefficient and support
budget; its proof is in \cref{app:geometric}.

\paragraph{\citet{zhang2026sharp} (arXiv v1, 31 August 2026, and v2, 2
October 2026).}  The unprotected coefficient appears as Theorem 6.1 and
Lemma 6.2 of v1 and as Theorem G.1 and Lemma G.2 of v2.  For points
\(a_1,\dots,a_s\) with weights \(q\) and every \(1\le t\le s\), some
probability vector \(\lambda\) with at most \(t\) nonzero entries satisfies
\(\|\sum_i\lambda_ia_i-\bar a_q\|^2\le\frac{s-t}{t(s-1)}V_q\), where
\(V_q=\sum_iq_i\|a_i-\bar a_q\|^2\).  The constant is sharp for the uniformly
weighted regular simplex.  The statement is deterministic.  Its proof
iterates a one-step boundary law that is unbiased and satisfies the scalar
bound for one fixed quadratic form, and the law may depend on the form.
Version 1 also states that a single law for all forms exists, with a
convex-separation justification; version 2 omits this remark.
The one-step inequality is proved through the
full-dimensional Grace-Danielsson inequality in v1 and by an elementary
argument in v2.  Neither version develops this sparsification bound under
arbitrary pathwise linear protection.
\Cref{lem:isotropic} gives a proof of the uniform one-step bound in the
protected setting, by a different argument (radial responses and a minimax
theorem).

\paragraph{\citet{tinarrage2026sharp} (arXiv v1, 24 June 2026).}  Theorem 3
gives the sharp radius
\(R_{d,t}=\sqrt{s(d+1-s)/(dt^2)}\), \(s=\min\{t,\lfloor(d+1)/2\rfloor\}\),
for equal-weight barycenters of \(t\)-tuples of vertices of a simplex in the
unit ball (the paper writes \(k\) for our \(t\)).  Lemma 4 gives, for a point
\(y\) whose barycentric coordinates are at most \(1/r\), an unbiased random
\(r\)-subset with
\(\E\|\frac1r\sum_{i\in R}v_i-y\|^2\le\frac1r(1-\|y\|^2)\), by ordered pivotal
sampling.  Both statements are scalar and impose no linear constraint beyond
the subset size.

\subsection{Further related work}
\label{app:related}

\paragraph{Balanced sampling and dependent rounding.}
Unequal-probability sampling
without replacement goes back to \citet{horvitz1952generalization}.  The cube
method draws samples with prescribed inclusion probabilities while keeping
auxiliary totals exactly or nearly balanced
\citep{deville1998splitting,deville2004cube,tille2025balanced}, and its
variance is approximated through residuals of a regression on the balancing
variables \citep{deville2005variance}, a scalar relative of \(\Cpa\).  For
uniform \(p\), equal inclusion probabilities \(m/k\) and \(d+1\) balancing
variables, their variance approximation with the correction for the loss of
degrees of freedom is \(\frac{k-m}{m(k-d-1)}\,r^\top\Cpa r\) for the
Horvitz-Thompson estimate of \(r^\top p\).  This is \(h_{k-d,m-d}\,r^\top\Cpa r\)
with \(m\) in place of \(m-d\).  The correction shifts the population by
\(d\), and the worst case shifts the budget as well.
Dependent and pipage rounding \citep{srinivasan2001distributions,%
gandhi2006dependent} and sub-isotropic rounding walks
\citep{bansal2019generalization} move a fractional point to a vertex by a
martingale that keeps selected linear constraints and controls correlations.
These works establish pathwise balance and unbiasedness, but not the exact
matrix profile of the covariance that must remain relative to \(\Cpa\) at
every budget.  For pivotal sampling, \citet{chauvet2017comparison} show that
the Horvitz-Thompson estimator is never worse than sampling with replacement,
and \cref{prop:fast} turns this into a Loewner guarantee for block protection,
without the finite population factor.  \Cref{sec:numerics} shows that a generic random-direction
flight does not attain it.

\paragraph{Sparse representations and unbiased compression.}
Tchakaloff-type
results give finite positive representations of moments
\citep{bayer2006tchakaloff}, approximate Carathéodory theorems give sparse
convex combinations with sharp \(\ell_2\) radius
\citep{mirrokni2017tight,tinarrage2026sharp}, and efficient unbiased
sparsification characterizes optimal laws for separable objectives
\citep{barnes2024efficient}.  Unbiased compressors in distributed optimization
control the variance of one vector at a time
\citep{alistarh2017qsgd,wangni2018gradient,wang2018atomo}, and
variance-optimal stream sampling keeps a fixed-size sample whose subset sums
have minimal total variance \citep{cohen2011efficient}.  Our atoms carry
arbitrary weights, satisfy pathwise affine constraints, and are controlled in
Loewner order for all responses at once.

\paragraph{Sparse and sampled attention.}
Heavy-hitter and streaming caches keep a few keys deterministically
\citep{zhang2023h2o,xiao2024efficient}.  MagicPIG samples keys by
locality-sensitive hashing, keeps the sink and local tokens, and estimates the
output with self-normalized importance weights \citep{chen2024magicpig}.
vAttention keeps the sink, a local window and predicted top-\(k\) keys,
samples the remaining keys uniformly, and chooses the budget of each query and
head so that the relative error of the output is at most \(\eps\) with
probability at least \(1-\delta\) \citep{desai2026vattention}.  SANTA
computes the scores exactly and averages value rows drawn from the softmax
weights, independently or by stratified or systematic sampling of the
cumulative weights, and reports faster decoding kernels
\citep{lee2026santa}.  The guarantees of vAttention and SANTA concern the
output of one query and head, and none of these methods keeps the mass of
every position block on each draw.  \Cref{prop:fast} keeps every block mass
and bounds the covariance for all value projections at once by that of
independent draws inside the blocks, which is the estimator of SANTA run in
each block.  Protocol 7 compares such designs downstream
(\cref{app:llm-downstream}).

\paragraph{Closest results.}
\Cref{sec:comparison} compares our bound with the
protected guarantee of \citet{haverbeck2026risk}, whose construction also
drives risk bounds and KV-cache rounding algorithms, and with the concurrent
unprotected bound of \citet{zhang2026sharp}.  We treat the unprotected
coefficient, the boundary martingale and the v1 uniform-law remark as
concurrent prior work.  Our
contribution is the lift to arbitrary pathwise protection through
\cref{lem:bridge}, the protected uniform one-step lemma, the sharpness analysis in the
protected regime, the instance-optimal constants for partitions, and the
rational bit-model realization.

\paragraph{Spectral sparsification.}
Loewner-order guarantees are central to spectral sparsification, which keeps
and reweights a few rank-one terms of a positive semidefinite sum so that the
sparse sum stays within a constant factor of the original in Loewner order
\citep{spielman2011graph,batson2012twice}.  There the guarantee bounds the
sparse matrix itself.  Here the sparse object is a random probability vector
that must be unbiased and exactly protected, and the Loewner bound concerns its
covariance, which governs the error of every linear response.

\paragraph{Query-aware design and learning subsets.}
Sparse standard quadratic
optimization shows the difficulty of optimizing one quadratic objective over
a sparse simplex \citep{bomze2025tractable}.  Our fixed-barycenter problem
differs because every atom must be sparse and feasible while the mixture must
reproduce \(p\).  CRAIG and Grad-Match relate weighted gradient subsets to
optimization stability \citep{mirzasoleiman2020craig,killamsetty2021gradmatch}.
We use strong convexity only to interpret the protected coefficient and do
not propose a new subset-selection method.

\section{Geometric and covariance proofs}
\label{app:geometric}

\subsection{Tangent factorization}

Fix a positive state \(q\) on a support of size \(s\).  Put
\(u=\sqrt q\), \(P_0=I-uu^\top\), and
\(R=P_0\diag(\sqrt q)A^\top\).  Then
\(R^\top R=AC_qA^\top\).  For \(v\in\ker(u^\top)\) one has
\(A\diag(\sqrt q)v=R^\top v\), so the subspace
\(\ker(u^\top)\cap\ker(A\diag(\sqrt q))\) is the orthogonal complement of
\(\range(R)\) inside \(\range(P_0)\).  If \(U\) is an orthonormal basis of this
subspace, orthogonal projection gives
\[
 UU^\top=P_0-R(R^\top R)^\dagger R^\top.
\]
Multiplication on both sides by \(\diag(\sqrt q)\) yields
\[
 \diag(\sqrt q)UU^\top\diag(\sqrt q)
 =C_q-C_qA^\top(AC_qA^\top)^\dagger AC_q
 =C_{q\mid A}.
\]
Since \(q>0\) on its support, \(\range(C_q)=\ker(\ones^\top)\) there, so
\(\rank(AC_qA^\top)=\dim A(\ker\ones^\top)=\rank\Ahat-1\).  This is the
identity \(d=\rank\Ahat-1\) used in \cref{sec:setup}.  A vector \(v\) lies in
the subspace above if and only if \(\diag(\sqrt q)v\in\ker\Ahat\).  Hence for
a rational full-column-rank basis \(T\) of \(\ker\Ahat\),
\[
 C_{q\mid A}
 =T\bigl(T^\top\diag(q)^{-1}T\bigr)^{-1}T^\top ,
\]
and in particular \(\range(C_{q\mid A})=\ker\Ahat\) on the support of \(q\).
Every point of the protected slice with support in that of \(q\) is
\(q+\diag(\sqrt q)Ux\) for some \(x\in\R^r\), which is the parametrization
\(q(x)_i=q_i(1+a_i^\top x)\) with \(a_i=u_i/\sqrt{q_i}\) used in
\cref{sec:curve}.  The identities \(\sum_iq_ia_i=U^\top\sqrt q=0\) and
\(\sum_iq_ia_ia_i^\top=U^\top U=I_r\) follow from the choice of \(U\).

\subsection{\texorpdfstring{Proof of \cref{prop:comparator}}{Proof of Proposition 1}}

(a)~For \(\lambda\in\R^a\),
\(\Var(g_I-\lambda^\top a_I)=(g-A^\top\lambda)^\top\Cp(g-A^\top\lambda)\).
This convex quadratic is minimized where
\(A\Cp A^\top\lambda=A\Cp g\), which is solvable because
\(\range(A\Cp)=\range(A\Cp A^\top)\).  The minimum value is
\(g^\top\Cp g-g^\top\Cp A^\top(A\Cp A^\top)^\dagger A\Cp g=g^\top\Cpa g\).

(b)~Loewner order is the order of quadratic forms, and
\(\Var(g^\top Q)=g^\top\Cov(Q)g\), so the first equivalence follows from (a).
For a matrix \(G\),
\(\E\|G(Q-p)\|_2^2=\tr(G\Cov(Q)G^\top)\), and
\(\tr(G(c\Cpa-\Cov(Q))G^\top)\ge0\) whenever \(c\Cpa\succeq\Cov(Q)\).  The
converse follows by taking \(G=g^\top\).

(c)~\Cref{lem:bridge} with \(P=Q\) gives
\(\Cov(Q)\psdle\Cpa-\E C_{Q\mid A}\psdle\Cpa\).  At \(m=d+1\),
\(h_{K,1}=1\), and the sharpness part of \cref{thm:main} shows that one cannot
be replaced by a smaller constant uniformly over instances.

(d)~The range identity is the tangent factorization at \(q=p\).  The matrix
\(\Cp-\Cpa=(A\Cp)^\top(A\Cp A^\top)^\dagger(A\Cp)\) is positive
semidefinite, and it vanishes if and only if \(A\Cp=0\), that is, if and only
if \(d=\rank(A\Cp A^\top)=0\).  If \(g=A^\top\lambda+t\ones\), then
\(g^\top Q=\lambda^\top AQ+t\ones^\top Q=\lambda^\top Ap+t=g^\top p\) on every
realization, and (a) with the coefficient \(\lambda\) gives
\(g^\top\Cpa g\le\Var(t)=0\).

When the rows of \(A\) are the indicators of a partition, the linear span of
the constant and the protected statistics consists of all functions of the
class of \(I\).  The least-squares projection of \(e_I\) onto it is the
conditional expectation given the class, so
\(\Cpa=\E\Cov(e_I\mid\text{class of }I)\).

\paragraph{Spectral tail in the attention example.}
Adding rows to \(A\) can only decrease the comparator, since the minimum in
(a) is then taken over a larger set.  Let \(\Sigma_V=V\Cp V^\top\) and let
the columns of \(U\) be its top \(r\) eigenvectors, with eigenvalues
\(\Lambda_r\).  For the rows \(U^\top V\) alone,
\(V C_{p\mid U^\top V}V^\top=\Sigma_V-\Sigma_VU(U^\top\Sigma_VU)^\dagger
U^\top\Sigma_V=\Sigma_V-U\Lambda_rU^\top\), whose trace is the spectral tail
of \(\Sigma_V\) beyond rank \(r\).  Additional block indicators decrease it
further.

\subsection{Proof of the isotropic boundary lemma}

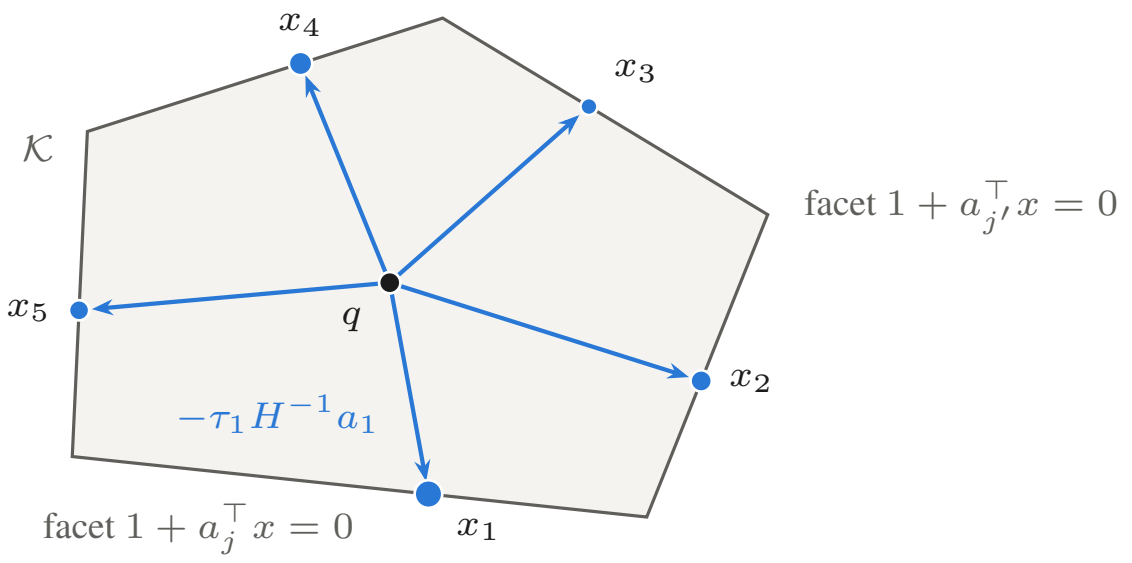
\begin{figure}[t]
  \centering
\definecolor{psA}{HTML}{2A78D6}
\definecolor{psB}{HTML}{EB6834}
\definecolor{psInk}{HTML}{1B1B1B}
\definecolor{psMuted}{HTML}{5F5E5A}
\definecolor{psFill}{HTML}{F4F3EF}
\begin{tikzpicture}[x=1cm,y=1cm,>={Stealth[length=4pt,width=3pt]},font=\scriptsize,
  every node/.style={text=psInk}]
  \coordinate (v1) at (-2.1,-1.15);
  \coordinate (v2) at ( 1.7,-1.55);
  \coordinate (v3) at ( 2.5, 0.45);
  \coordinate (v4) at ( 0.35, 1.75);
  \coordinate (v5) at (-2.0, 1.0);
  \fill[psFill] (v1)--(v2)--(v3)--(v4)--(v5)--cycle;
  \draw[psMuted,line width=0.6pt] (v1)--(v2)--(v3)--(v4)--(v5)--cycle;
  \node[anchor=north,text=psMuted] at ($(v1)!0.22!(v2)+(0,-0.1)$) {facet $1+a_j^\top x=0$};
  \node[anchor=west,text=psMuted] at ($(v3)+(0.12,0.05)$) {facet $1+a_{j'}^\top x=0$};
  \node[anchor=south east,text=psMuted] at ($(v5)!0.15!(v1)+(-0.08,0)$) {$\mathcal K$};
  \coordinate (o) at (0,0);
  \coordinate (x1) at ($(v1)!0.62!(v2)$);
  \coordinate (x2) at ($(v2)!0.45!(v3)$);
  \coordinate (x3) at ($(v3)!0.55!(v4)$);
  \coordinate (x4) at ($(v4)!0.40!(v5)$);
  \coordinate (x5) at ($(v5)!0.55!(v1)$);
  \foreach \i in {1,...,5}{
    \draw[->,psA,line width=0.8pt,shorten >=2.4pt,shorten <=2.2pt] (o) -- (x\i);
  }
  \foreach \i/\r in {1/2.6,2/2.0,3/1.6,4/2.2,5/1.9}{
    \filldraw[fill=psA,draw=white,line width=0.5pt] (x\i) circle (\r pt);
  }
  \node[anchor=north west] at ($(x1)+(0.06,-0.04)$) {$x_1$};
  \node[anchor=west] at ($(x2)+(0.06,0)$) {$x_2$};
  \node[anchor=south west] at ($(x3)+(0.04,0.04)$) {$x_3$};
  \node[anchor=south] at ($(x4)+(0,0.06)$) {$x_4$};
  \node[anchor=east] at ($(x5)+(-0.06,0)$) {$x_5$};
  \filldraw[fill=psInk,draw=white,line width=0.5pt] (o) circle (2pt);
  \node[anchor=north east] at ($(o)+(-0.06,-0.04)$) {$q$};
  \node[text=psA,anchor=east] at ($(o)!0.62!(x1)+(-0.1,0)$) {$-\tau_1H^{-1}a_1$};
\end{tikzpicture}
  \caption{\textbf{The radial best response behind \cref{lem:isotropic}.}
  In slice coordinates the protected slice through \(q\) is the polytope
  \(\mathcal K\) with facets \(1+a_i^\top x=0\), one per support coordinate,
  and \(q\) sits at the origin.  For a test matrix \(H\succ0\), a ray is shot
  from the origin in direction \(-H^{-1}a_i\) to its first boundary point
  \(x_i\), and the endpoint receives weight proportional to \(p_i/\tau_i\)
  (inverse ray length).  Isotropy makes the law centered and gives
  \(\tr(H\Cov X)\le 1/\tr(H^{-1})\).  A minimax over \(H\) produces one law
  valid in every direction.  Every endpoint lies on a facet, so one support
  coordinate is dropped.}
  \label{fig:radial}
\end{figure}

\Cref{fig:radial} illustrates the construction.  The isotropy assumptions
imply that zero is in the interior of \(\mathcal K\) and that \(\mathcal K\)
is bounded.  Indeed, along every nonzero direction \(w\) the values
\(a_i^\top w\) have mean zero and positive second moment under the weights
\(p_i\), so some value is negative and the ray meets a facet.

Fix \(H\succ0\).  For every nonzero \(a_i\), move from zero along
\(-H^{-1}a_i\) to the first boundary point
\[
 x_i=-\tau_iH^{-1}a_i,
 \qquad \tau_i>0.
\]
The \(i\)-th facet inequality at \(x_i\) gives
\(\tau_i a_i^\top H^{-1}a_i\le1\).  Define
\[
 Z=\sum_{a_i\ne0}\frac{p_i}{\tau_i},
 \qquad
 \pi_i=\frac{p_i/\tau_i}{Z}.
\]
Zero normals may be omitted because they contribute to neither isotropic
moment.  The boundary law is centered:
\[
 \sum_i\pi_i x_i
 =-\frac{H^{-1}}Z\sum_i p_i a_i=0.
\]
Since \(\sum_ip_i=1\), its cost obeys
\[
 \sum_i\pi_i x_i^\top Hx_i
 =\frac1Z\sum_i p_i\tau_i a_i^\top H^{-1}a_i
 \le \frac1Z\sum_ip_i\le\frac1Z .
\]
Moreover,
\[
 Z\ge \sum_i p_i a_i^\top H^{-1}a_i=\tr(H^{-1}).
\]
The arithmetic-harmonic mean inequality therefore gives
\[
 \tr\!\left(H\Cov(X)\right)
 \le \frac1{\tr(H^{-1})}
 \le \frac{\tr(H)}{r^2}.
\]

Let \(\mathfrak L\) be the set of probability laws on the compact set
\(\partial\mathcal K\) with mean zero.  It is convex and weakly compact.  Let
\(\mathfrak H=\{H\succeq0:\tr(H)=1\}\).  For \(\nu\in\mathfrak L\) the map
\(H\mapsto\tr(H\Cov_\nu X)\) is linear, and for fixed \(H\) the map
\(\nu\mapsto\tr(H\Cov_\nu X)=\int x^\top Hx\,d\nu(x)\) is linear and weakly
continuous.  Sion's minimax theorem \citep{sion1958general} therefore gives
\[
 \min_{\nu\in\mathfrak L}\max_{H\in\mathfrak H}
 \tr(H\Cov_\nu(X))
 =
 \max_{H\in\mathfrak H}\min_{\nu\in\mathfrak L}
 \tr(H\Cov_\nu(X)) .
\]
For \(H\succ0\) in \(\mathfrak H\) the radial law shows that the inner
minimum is at most \(r^{-2}\).  For singular \(H\in\mathfrak H\) and
\(\epsilon>0\), put \(H_\epsilon=(H+\epsilon I)/(1+r\epsilon)\in\mathfrak
H\).  Since \(H\psdle(1+r\epsilon)H_\epsilon\), the radial law for
\(H_\epsilon\) has \(H\)-cost at most \((1+r\epsilon)/r^2\), and letting
\(\epsilon\downarrow0\) bounds the inner minimum by \(r^{-2}\).  Hence the
right side is at most \(r^{-2}\).  The left side is
\(\min_\nu\lambda_{\max}(\Cov_\nu X)\), so a minimax law satisfies
\(\Cov X\psdle r^{-2}I_r\).  Finally, some minimax law puts at most one
atom on each facet \(\{x\in\mathcal K:1+a_i^\top x=0\}\).  Assign every
boundary point to one facet that contains it and replace the conditional law
of \(X\) on each facet by its barycenter.  This keeps the mean, stays on the
facet because the facet is convex, and does not increase \(\Cov X\) in
Loewner order, because a second moment dominates the outer product of its
mean.  The resulting law has at most one atom per index \(i\), and it can be
computed by a semidefinite program with one block of size \(r+1\) per facet.

\subsection{Proof of the residual bridge}

The proof in \cref{sec:curve} is complete.  The following regression argument
gives the same bound and explains it as a split of residual variance.
Conditionally on \(P\), draw \(I\sim P\).  Marginally \(I\sim p\).  Set
\[
 X=e_I-p,
 \quad Y=AX,
 \quad L_p=\Cp A^\top(A\Cp A^\top)^\dagger,
 \quad R=X-L_pY.
\]
Then \(\Cov(R)=\Cpa\).  Since \(AP=Ap\) pathwise,
\(\E[R\mid P]=P-p\).  For \(P=q\), the centered conditional residual is
\((I-L_pA)(e_I-q)\).  With \(S_q=AC_qA^\top\) and
\(L_q=C_qA^\top S_q^\dagger\), completion of squares gives, for every
matrix \(L\),
\[
 (I-LA)C_q(I-LA)^\top
 =C_{q\mid A}
 +(L-L_q)S_q(L-L_q)^\top.
\]
The identity uses \(S_qS_q^\dagger AC_q=AC_q\), which holds because
\(\range(AC_q)=\range(AC_qA^\top)\).  Thus the conditional covariance of the
residual given \(P=q\) dominates \(C_{q\mid A}\), and the law of total
covariance yields
\[
 \Cpa=\Cov(R)
 =\Cov(\E[R\mid P])+\E\Cov(R\mid P)
 \succeq \Cov(P)+\E C_{P\mid A}.
\]

\subsection{Recurrence identity and sharpness}

For \(K=k-d\) and \(M=t-d\), the update
\(\eta_{t-1}=\alpha_t+(1-\alpha_t)\eta_t\) with
\(\alpha_t=(t-d-1)^{-2}\) is equivalent to
\[
 1-\eta_{t-1}
 =\left(1-\frac1{(M-1)^2}\right)(1-\eta_t).
\]
Multiplying these factors and using
\[
 \prod_{j=M}^{K-1}\left(1-\frac1{j^2}\right)
 =\frac{(M-1)K}{M(K-1)}
\]
gives \(\eta_m=1-(M-1)K/(M(K-1))=h_{K,M}\).

\paragraph{Unprotected lower bound.}
For \(d=0\) and uniform \(p=\ones/k\), any unbiased \(m\)-sparse \(Q\) has
\(\|Q\|_2^2\ge1/m\) pathwise, so
\(\tr\Cov(Q)=\E\|Q\|_2^2-1/k\ge1/m-1/k\), while
\(\tr\Cp=1-1/k\).  If \(\Cov(Q)\psdle c\,\Cp\) then taking traces gives
\(c\ge(1/m-1/k)/(1-1/k)=h_{k,m}\).  The uniform \(m\)-subset law has
\(\Cov(Q)=h_{k,m}\Cp\) exactly, by symmetry and the trace identity.

\paragraph{Protected lower bound.}
For \(d>0\), take
\[
 p_i=\delta/d\quad(i\le d),
 \qquad
 p_i=(1-\delta)/K\quad(i>d),
 \qquad A=[I_d\ \ 0],
\]
so that \(A\Cp A^\top=\diag(p_{1:d})-p_{1:d}p_{1:d}^\top\) has rank \(d\)
because \(\sum_{i\le d}p_i=\delta<1\).  Every admissible \(Q\) has
\(Q_i=p_i>0\) for \(i\le d\) pathwise, so \(d\) support positions are
consumed and the free block \(Q_F=(Q_i)_{i>d}\) is \(M\)-sparse with
\(\ones^\top Q_F=1-\delta\).  Then \(\widetilde Q=Q_F/(1-\delta)\) is an
unbiased \(M\)-sparse probability vector on \(K\) coordinates with uniform
mean, and \(\Cov(Q)\) is \((1-\delta)^2\Cov(\widetilde Q)\) embedded in the
free block.  The regression of \(e_I\) on the anchor indicators is the
conditional expectation given the partition into the anchors and the free
block, hence
\(\Cpa=(1-\delta)\,C_{\rm unif,K}\) embedded in the free block, with
\(C_{\rm unif,K}=(I-\ones\ones^\top/K)/K\).  If \(\Cov(Q)\psdle c\,\Cpa\),
taking traces on the free block and using the unprotected bound for
\(\widetilde Q\) gives
\((1-\delta)^2h_{K,M}\tr C_{\rm unif,K}\le c\,(1-\delta)\tr C_{\rm unif,K}\),
that is \(c\ge(1-\delta)h_{K,M}\).  Letting \(\delta\downarrow0\) shows that
the supremum over instances of effective rank \(d\) is at least \(h_{K,M}\),
which matches the upper bound of \cref{thm:main}.

\subsection{\texorpdfstring{Proof of \cref{cor:haverbeck}}{Proof of the comparison corollary}}

The barycenter error is \(\eta=X(Q-p)\), so \(U^\top\eta=A(Q-p)=0\) on every
realization, and \(d=\rank(A\Cp A^\top)\le d_U\) because \(A\) has \(d_U\)
rows.  \Cref{thm:main} gives
\(\E[\eta\eta^\top]=X\Cov(Q)X^\top\psdle h_{k-d,m-d}X\Cpa X^\top\).  With
\(\Sigma=X\Cp X^\top\), \(X\Cp A^\top=\Sigma U\) and
\(A\Cp A^\top=U^\top\Sigma U\), so
\(X\Cpa X^\top=\Sigma-\Sigma U(U^\top\Sigma U)^\dagger U^\top\Sigma\), the
Schur complement of \(\Sigma\) with respect to \(U\).  For every
\(v\in\R^D\),
\[
 v^\top\bigl(\Sigma-\Sigma U(U^\top\Sigma U)^\dagger U^\top\Sigma\bigr)v
 =\min_{\lambda}(v-U\lambda)^\top\Sigma(v-U\lambda)
 \le(v-\Pi_Uv)^\top\Sigma(v-\Pi_Uv),
\]
by the choice \(\lambda=U^\top v\), which proves the second inequality
together with \(h_{K,M}\le1/M\).  At total support \(m\), Proposition A.3 of
\citet{haverbeck2026risk} applies with \(m'=m-2d_U-2\ge1\) and constant
\(20/(9m')\).  The ratio of the two constants is at least
\(20(m-d)/(9(m-2d_U-2))\ge20(m-d_U)/(9(m-2d_U-2))>20/9\).

\section{Proofs for Section~\ref{sec:universal}}
\label{app:universal}

\subsection{\texorpdfstring{Proof of \cref{thm:universal}}{Proof of Theorem 2}}

For \(d=0\) take \(p\) uniform.  Every \(m\)-sparse probability vector \(q\)
has \(\|q\|_2^2\ge1/m\) and \(q^\top p=1/k\), so
\(\|q-p\|_2^2=\|q\|_2^2-1/k\ge1/m-1/k\) for every realization, whatever the
law.  If \(\E[(Q-p)(Q-p)^\top]\psdle c\,\Cp\), taking traces gives
\(c\ge(1/m-1/k)/(1-1/k)=h_{k,m}\).

For \(d>0\) use the anchor instance of \cref{app:geometric}: \(d\) anchors of
mass \(\delta/d\) protected by \(A=[I_d\ 0]\), and \(K\) free coordinates of
mass \((1-\delta)/K\).  Protection fixes every anchor on every realization,
so the free part \(q_F\) of a realization has total mass \(1-\delta\) and at
most \(M\) nonzero entries.  Hence \(\|q_F\|_2^2\ge(1-\delta)^2/M\), and since
\(p_F\) is constant on the free block,
\(q_F^\top p_F=(1-\delta)^2/K=\|p_F\|_2^2\).  Therefore
\[
 \|q_F-p_F\|_2^2=\|q_F\|_2^2-\frac{(1-\delta)^2}{K}
 \ge(1-\delta)^2\Bigl(\frac1M-\frac1K\Bigr)
\]
for every realization, biased laws included.  The comparator is
\(\Cpa=(1-\delta)C_{\rm unif,K}\) on the free block, with trace
\((1-\delta)(K-1)/K\).  If \(\E[(Q-p)(Q-p)^\top]\psdle c\,\Cpa\), the traces
on the free block give \(c\ge(1-\delta)h_{K,M}\), and \(\delta\downarrow0\)
proves the claim.

\subsection{\texorpdfstring{Proof of \cref{thm:designs}}{Proof of Theorem 3}}

\emph{(i) Equal weights.}  Let \(\pi_i=mp_i\le1\) and let \(\Delta\) be the
covariance of the inclusion indicators, so that \(\Cov(Q)=\Delta/m^2\) and
\(\Delta_{ii}=\pi_i(1-\pi_i)\).  With \(D=\diag(1/p)\),
\[
 \tr(D\Cov(Q))=\frac1{m^2}\sum_i\frac{\pi_i(1-\pi_i)}{p_i}=\frac{k-m}{m},
 \qquad
 \tr(D\Cp)=\sum_i(1-p_i)=k-1 .
\]
If \(\Cov(Q)\psdle c\,\Cp\), then
\(c\ge\tr(D\Cov(Q))/\tr(D\Cp)=h_{k,m}\).  If \(c=h_{k,m}\), then
\(h_{k,m}\Cp-\Cov(Q)\) is positive semidefinite with zero \(D\)-weighted
trace and \(D\succ0\), so \(\Cov(Q)=h_{k,m}\Cp\).  Comparing diagonals,
\(h_{k,m}p_i(1-p_i)=p_i(1-mp_i)/m\) for every \(i\), or equivalently
\((m-1)(kp_i-1)=0\).  If \(m>1\), this forces \(p_i=1/k\).
If \(m=1\), the law is categorical sampling with probabilities \(p_i\),
so \(\Cov(Q)=\Cp\) and \(\lambda(Q)=h_{k,1}=1\) for every \(p\).

\emph{(ii) VarOpt.}  Assume \(m<k\) and let \(L=k-m+1\ge2\), so that the light coordinates have mass
\(1/(mL)\).  The VarOpt threshold is \(\tau=1/m\): the heavy coordinates have
inclusion probability one and the light ones \(1/L\), which sum to \(m\).
Every VarOpt design therefore keeps all heavy coordinates and exactly one
light coordinate, chosen uniformly, with weight \(\tau=1/m\).  An equal-weight
design with inclusion probabilities \(mp_i\) is the same law.  Its covariance
lives on the light block and equals \(\frac1{m^2L}(I-\ones\ones^\top/L)\).
For a nonzero light contrast \(x\) (supported on the light block with
\(\ones^\top x=0\)) the Rayleigh quotient is
\(\frac{\|x\|^2/(m^2L)}{\|x\|^2/(mL)}=1/m\).  Conversely, for every \(x\),
\(x^\top\Cp x=\Var_{I\sim p}(x_I)\ge\Prob(I\ \text{light})\,
\Var(x_I\mid I\ \text{light})=\frac1{mL}\|P x_L\|^2=m\,x^\top\Cov(Q)x\),
where \(P\) projects onto light contrasts.  Hence \(\lambda(Q)=1/m\), and
\(\frac1m\big/h_{k,m}=\frac{k-1}{k-m}\).
When \(m=k\), VarOpt selects every coordinate and returns \(Q=p\), hence
\(\lambda(Q)=0\).  A full-subset equal-weight design is admissible only
for uniform \(p\), when it also returns \(p\).

\emph{(iii) Ratio design.}  Let the heavy coordinate have mass \(b\) and the
others \(v=(1-b)/(k-1)\).  Since \(\Prob(S)=p(S)/\binom{k-1}{m-1}\),
\[
 P_1:=\Prob(1\in S)=b+(m-1)v ,
\]
and on \(\{1\in S\}\) the realization has \(p(S)=P_1\) and \(Q_1=b/P_1\).
Thus \(\E Q_1^2=b^2/P_1\) and \(\Var(Q_1)=b^2(1-P_1)/P_1\).  The Rayleigh
quotient at \(x=e_1\), where \(x^\top\Cp x=b(1-b)\), gives
\[
 \lambda(Q)\ge\frac{b(1-P_1)}{(1-b)P_1}
 =\frac{b(k-m)}{b(k-1)+(m-1)(1-b)} ,
\]
which at \(b=1/2\) equals \(\frac{k-m}{k+m-2}=\frac{m(k-1)}{k+m-2}\,h_{k,m}\).
At \(m=k\), \(S=[k]\) almost surely, so \(Q=p\) and \(\lambda(Q)=0\).

\emph{Independent sampling.}  For \(Q=\frac1m\sum_{t\le m}e_{I_t}\) with
independent \(I_t\sim p\), \(\Cov(Q)=\Cp/m\) exactly.

\section{Instance-adaptive bounds and scope statements}
\label{app:partition}

\subsection{Block-separable comparator}

\begin{lemma}[Block decomposition]
\label{lem:block-comparator}
Let \([k]\) be partitioned into classes \(B_1,\dots,B_c\) with masses
\(\pi_j\) and within-class weights \(p^{(j)}\).  Suppose the rows of \(A\)
consist of the class indicators and of rows each supported in a single
class, and let \(A_j\) collect the rows supported in \(B_j\), restricted to
\(B_j\).  Then
\(\Cpa=\sum_j\pi_j\,C_{p^{(j)}\mid A_j}\), each term embedded in the block
\(B_j\times B_j\), where \(C_{p^{(j)}\mid A_j}=C_{p^{(j)}}\) if \(A_j\) is
empty.
\end{lemma}

\begin{proof}
By \cref{prop:comparator}(a), \(\Cpa\) is the covariance of the residual of
\(e_I\) after least-squares projection, in \(L_2(p)\), onto the span of the
constant and the protected statistics.  This span is the orthogonal direct sum
over \(j\) of the spans of \(\ones_{B_j}(I)\) and of the protected features of
class \(j\) multiplied by \(\ones_{B_j}(I)\), since functions supported on
disjoint classes are orthogonal.  For a coordinate \(i\in B_j\), the function
\(\ones[I=i]\) vanishes outside \(B_j\), so its projection and its residual
\(R_i\) vanish outside \(B_j\) too.  Residuals of coordinates in different
classes therefore have zero covariance.  Within \(B_j\),
\(\E[R_iR_{i'}]=\pi_j\E[R_iR_{i'}\mid I\in B_j]\), and conditionally on
\(I\in B_j\) the projection is the regression on the constant and the rows of
\(A_j\) under \(p^{(j)}\), whose residual covariance is
\(C_{p^{(j)}\mid A_j}\).
\end{proof}

\subsection{\texorpdfstring{Proof of \cref{prop:partition}}{Proof of the partition proposition}}

For class indicators, \cref{lem:block-comparator} gives
\(\Cpa=\sum_j\pi_jC_{p^{(j)}}\) with
\(C_{p^{(j)}}=\diag(p^{(j)})-p^{(j)}p^{(j)\top}\) embedded in \(B_j\).  Write
\[
 \ell_j(x)=\pi_j\,\frac{\psi(x)-\|p^{(j)}\|_2^2}{1-\|p^{(j)}\|_2^2}
\]
for the function \(u_j\) with \(1/k_j\) replaced by \(\|p^{(j)}\|_2^2\), and
\(\ell_j=0\) if \(k_j=1\).  The lower bound of \cref{prop:partition} is
\(\gamma^*\ge\min_{x\in X_m}\max_j\ell_j(x_j)\).

\emph{Lower bound.}  Let \(Q\) be admissible at budget \(m\) with
\(\Cov(Q)\psdle\gamma\Cpa\).  Protection gives \(\sum_{i\in B_j}Q_i=\pi_j\)
on every realization, so \(Q^{(j)}=Q|_{B_j}/\pi_j\) is a random probability
vector on \(B_j\) with \(\E Q^{(j)}=p^{(j)}\).  Let
\(s_j=|\supp(Q)\cap B_j|\).  Then \(1\le s_j\le k_j\) and \(\sum_js_j\le m\)
on every realization, so \(x=(\E s_j)_j\in X_m\).  Fix \(j\) with
\(k_j\ge2\).  The principal block of the Loewner inequality on \(B_j\) reads
\(\pi_j^2\Cov(Q^{(j)})\psdle\gamma\pi_jC_{p^{(j)}}\), and taking traces gives
\[
 \pi_j\bigl(\E\|Q^{(j)}\|_2^2-\|p^{(j)}\|_2^2\bigr)
 \le\gamma\bigl(1-\|p^{(j)}\|_2^2\bigr).
\]
A probability vector with \(s\) nonzero coordinates has squared norm at least
\(1/s\) by the Cauchy-Schwarz inequality.  Since \(\psi\) is convex and
\(\psi(s)=1/s\) at integers, Jensen's inequality gives
\(\E\|Q^{(j)}\|_2^2\ge\E[1/s_j]=\E\psi(s_j)\ge\psi(x_j)\).  Hence
\(\gamma\ge\ell_j(x_j)\) for every \(j\), and so
\(\gamma\ge\max_j\ell_j(x_j)\ge\min_{x'\in X_m}\max_j\ell_j(x'_j)\).

\emph{Upper bound.}  The law of \cref{thm:main} attains
\(h_{k-c+1,m-c+1}\).  For \(x\in X_m\), write \(x_j=\lfloor
x_j\rfloor+f_j\), draw \(U\) uniform on \([0,1)\), put \(F_j=\sum_{i\le
j}f_i\), and set \(s_j=\lfloor x_j\rfloor+\lfloor F_j+U\rfloor-\lfloor
F_{j-1}+U\rfloor\).  This systematic rounding gives
\(s_j\in\{\lfloor x_j\rfloor,\lceil x_j\rceil\}\), \(\E s_j=x_j\), and
\(\sum_js_j\le\lceil\sum_jx_j\rceil\le m\) on every realization.  Given
\(s\), set \(Q^{(j)}=p^{(j)}\) if \(s_j=k_j\); otherwise draw \(Q^{(j)}\) from
the unprotected law of \cref{thm:main} for \(p^{(j)}\) at budget \(s_j\).
Make the choices independently across classes and set
\(Q=\sum_j\pi_jQ^{(j)}\) with each term embedded in \(B_j\).
Then \(Q\) is admissible.  The conditional means given \(s\) do not
depend on \(s\), so
\[
 \Cov(Q)=\E\Cov(Q\mid s)
 \psdle\sum_j\pi_j^2\,\E\bigl[h_{k_j,s_j}\bigr]C_{p^{(j)}} .
\]
For \(k_j\ge2\), \(h_{k_j,s}=(k_j/s-1)/(k_j-1)\) is affine in \(1/s\), and \(s_j\) takes the
two integers next to \(x_j\) with mean \(x_j\),
so \(\E[1/s_j]=\psi(x_j)\) and \(\pi_j\E[h_{k_j,s_j}]=u_j(x_j)\).
Singleton classes contribute zero.  Therefore
\(\Cov(Q)\psdle\sum_ju_j(x_j)\pi_jC_{p^{(j)}}\psdle\max_ju_j(x_j)\,\Cpa\).
Minimizing over the compact set \(X_m\) gives the upper bound.

\emph{Equality.}  If \(p^{(j)}\) is uniform, then
\(\|p^{(j)}\|_2^2=1/k_j\) and \(\ell_j=u_j\).

\paragraph{Proof of \cref{cor:equal-classes}.}  All \(u_j\) equal
\(u(x)=\frac1c\frac{n\psi(x)-1}{n-1}\), which decreases in \(x\).  Every
\(x\in X_m\) has \(\min_jx_j\le m/c\), so \(\max_ju(x_j)\ge u(m/c)\), with
equality at \(x_j=m/c\).  If \(c\) divides \(m\), then \(\psi(m/c)=c/m\) and
\(u(m/c)=(k-m)/(m(k-c))\).

\subsection{Block-separable protection}

\begin{proposition}[Block-separable protection]
\label{prop:block}
In the setting of \cref{lem:block-comparator}, let
\(d_j=\rank(A_jC_{p^{(j)}}A_j^\top)\) and let integers \(s_j\) satisfy
\(d_j+1\le s_j\le k_j\) and \(\sum_js_j\le m\).
Take \(Q^{(j)}=p^{(j)}\) when \(s_j=k_j\), and otherwise draw \(Q^{(j)}\)
from the law of \cref{thm:main} at budget \(s_j\).
Independent choices across classes give an admissible law with
\[
 \Cov(Q)\psdle\max_j\pi_jh_{k_j-d_j,s_j-d_j}\,\Cpa,
\]
where \(h_{k_j-d_j,s_j-d_j}=0\) if \(s_j=k_j\).
\end{proposition}

\begin{proof}
For every block with \(s_j<k_j\), one has
\(d_j\le s_j-1\le k_j-2\), so \cref{thm:main} applies.
Each \(Q^{(j)}\) keeps the class mass and \(A_jQ^{(j)}=A_jp^{(j)}\), so
\(Q=\sum_j\pi_jQ^{(j)}\) is admissible.  Independence and
\cref{lem:block-comparator} give
\(\Cov(Q)=\sum_j\pi_j^2\Cov(Q^{(j)})\psdle
\sum_j\pi_j^2h_{k_j-d_j,s_j-d_j}C_{p^{(j)}\mid A_j}\psdle
\max_j\pi_jh_{k_j-d_j,s_j-d_j}\Cpa\).
\end{proof}

\subsection{\texorpdfstring{Proof of \cref{cor:heavy}}{Proof of the heavy-coordinates corollary}}

\begin{proof}
If \(m=k\), take \(Q=p\); both sides of the stated bound are zero.
Assume \(m<k\).
Every row \(a\) of \(A\) splits as \(a|_H+a|_L\), and \(a|_H\) is a
combination of the indicators of the singletons \(\{i\}\), \(i\in H\).  Let
\(A'\) stack these indicators, the indicator of \(L\), and the rows of
\(A_L\) embedded in \(L\).  The law keeps every coordinate in \(H\) and
satisfies \(A_LQ_L=A_Lp_L\), so it protects \(A'\) and hence \(A\).  The
singletons and \(L\) partition \([k]\), and every row of \(A'\) is a class
indicator or is supported in \(L\).  \Cref{prop:block} with \(s_i=1\) for the
singletons and \(s_L=m-j\) gives
\(\Cov(Q)\psdle\pi_Lh_{k-j-d_L,m-j-d_L}C_{p\mid A'}\).  The protected
statistics of \(A\) are linear in those of \(A'\), so \(C_{p\mid A'}\psdle\Cpa\)
as in \cref{prop:unknown-A}(ii).
\end{proof}

\subsection{\texorpdfstring{Proof of \cref{prop:fast}}{Proof of the fast-sampler proposition}}

\paragraph{Pivotal sampling.}
Let \(x\in(0,1]^n\) have an integer sum \(s\).  Pivotal sampling
\citep{deville1998splitting} visits the coordinates in a given order and
keeps one current coordinate with a fractional value.  When the current
coordinate \(i\) meets the next fractional coordinate \(j\), the pair
\((x_i,x_j)\) becomes \((x_i+x_j,0)\) or \((0,x_i+x_j)\) with probabilities
proportional to \(x_i\) and \(x_j\) if \(x_i+x_j\le1\), and \((1,x_i+x_j-1)\)
or \((x_i+x_j-1,1)\) with probabilities proportional to \(1-x_j\) and
\(1-x_i\) otherwise, and the coordinate that is still fractional becomes
current.  Each step keeps the mean and the sum and decides at least one
coordinate, so after at most \(n-1\) steps the result is the indicator of a
set \(S\) of exactly \(s\) coordinates with \(\Pr(i\in S)=x_i\).  The step is
the dependent rounding step of \citet{srinivasan2001distributions}, and our
sampler visits the coordinates in a uniformly random order.

\paragraph{Pivotal sampling never loses to sampling with replacement.}
The following claim is Theorem~1 of \citet{chauvet2017comparison}.  We include
a complete proof.  For every \(z\in\R^n\) and every order,
\begin{equation}
 \Var\Bigl(\sum_{i\in S}z_i\Bigr)\le W(x,s),\qquad
 W(x,s)=\sum_ix_iz_i^2-\frac1s\Bigl(\sum_ix_iz_i\Bigr)^2,
 \label{eq:pivotal-wr}
\end{equation}
and \(W(x,s)\) is the variance of \(\sum_{t=1}^sz_{I_t}\) for \(s\) independent
draws with \(\Pr(I_t=i)=x_i/s\).  A random order is a mixture of fixed orders
under which the total has the same mean, so its variance is the average of
theirs, and it suffices to fix the order.  We argue by induction on \(s\).
For \(s=1\) every comparison involves a total mass at most one, so \(S\) is a
single draw with probabilities \(x_i\), and equality holds.

Let \(s\ge2\) and \(C_i=x_1+\dots+x_i\).  Call coordinate \(i\)
\emph{cross-border} if \(C_{i-1}<\ell\le C_i\) for an integer \(\ell<s\), and
call a maximal stretch of the other coordinates a \emph{run}.  The current
coordinate that enters a run and the coordinates of the run have a total mass
at most one, so every comparison inside the run is of the first kind, and the
coordinate of the run that carries its mass is drawn with probabilities
proportional to \(x\), independently of everything else, and plays no further
role in the probabilities.  Pivotal sampling is therefore two-stage
\citep{chauvet2012characterization}.  The first stage is pivotal sampling of
the clustered population in which each run is one coordinate, with the mass
\(\phi\) of the run and the \(x\)-weighted mean of \(z\) over the run as its
value, and the second stage draws one coordinate in each selected run with
probabilities proportional to \(x\).  Sampling with replacement has the same
two-stage form with independent first-stage draws.  For both designs the
variance is the first-stage variance plus the same term
\(\sum_{\text{runs}}\sum_{i\in\text{run}}x_i(z_i-\bar z_{\text{run}})^2\), so we
may assume that runs and cross-border coordinates alternate and that every
run is a single coordinate.

The first comparison is then between the first run, of mass \(1-a\) and value
\(Y_1\), and the first cross-border coordinate, of mass \(a+b\) and value
\(Y_2\), where \(0<a\le a+b\le1\) and \(b<1\).  With probability
\((1-a-b)/(1-b)\) the run is selected and the cross-border coordinate goes on
with mass \(b\), and otherwise the roles are exchanged.  The rest is pivotal
sampling of \(s-1\) coordinates from the remaining population, with masses
\(\phi'\).  The two outcomes give conditional means of the total that differ
by \((1-b)(Y_1-Y_2)\), so the law of total variance and the induction
hypothesis give
\[
 \Var\Bigl(\sum_{i\in S}z_i\Bigr)\le a(1-a-b)(Y_1-Y_2)^2+\E\,W(\phi',s-1).
\]
Let \(R\) be the mass-weighted sum of the values after the first two
coordinates, \(u=Y_1-Y_2\), \(w=R-(s-1-b)Y_2\) and
\(\kappa=a(s-1+b)/(1-b)-(s-1)\).  Expanding \(W\) for the clustered
population gives
\begin{align*}
 &W(\phi,s)-a(1-a-b)u^2-\E\,W(\phi',s-1)\\
 &\qquad=\frac{(w+\kappa u)^2}{s(s-1)}
 +\frac{ab(1-a-b)\bigl(2(s-1)-b(s-2)\bigr)}{(1-b)^2(s-1)}\,u^2\ge0,
\end{align*}
which proves \cref{eq:pivotal-wr}.  The supplementary material checks this
identity symbolically.

\paragraph{Proof of \cref{prop:fast}.}
Every coordinate of \(L_b\) is sampled with probability \(s_bp_i/\pi_b\le1\)
and gets weight \(\pi_b/s_b\), so \(\E Q_i=p_i\), and a sample of exactly
\(s_b\) coordinates carries the mass \(\pi_b\).  The kept coordinates and the
class masses are therefore exact, the support is at most
\(|H|+\sum_bs_b\le m\), and the law is admissible.  Fix a class, let
\(p_{L_b}=p|_{L_b}/\pi_b\), and apply \cref{eq:pivotal-wr} with
\(x_i=s_bp_i/\pi_b\) and \(z_i=(\pi_b/s_b)g_i\).  The total
\(\sum_{i\in S}z_i\) is \(\sum_{i\in L_b}g_iQ_i\), and
\[
 W(x,s_b)=\frac{\pi_b}{s_b}\Bigl(\sum_{i\in L_b}p_ig_i^2
 -\frac1{\pi_b}\Bigl(\sum_{i\in L_b}p_ig_i\Bigr)^2\Bigr)
 =\frac{\pi_b}{s_b}\,\pi_b\,g^\top C_{p_{L_b}}g ,
\]
so \(\Cov(Q|_{L_b})\psdle(\pi_b/s_b)\,\pi_bC_{p_{L_b}}\).  The classes are
sampled independently and the kept coordinates do not vary, so summing over
\(b\) and applying \cref{lem:block-comparator} to the partition into the
singletons of \(H\) and the sets \(L_b\) gives
\(\Cov(Q)\psdle\max_b(\pi_b/s_b)\,C_{p\mid A'}\), where the rows of \(A'\) are
the indicators of these classes.  The class totals of \(A\) are linear in
those of \(A'\), so \(C_{p\mid A'}\psdle\Cpa\) as in
\cref{prop:unknown-A}(ii).  Each step of pivotal sampling decides a
coordinate, so a draw takes \(O(k)\) time.

\subsection{\texorpdfstring{What must be known about \(A\), and partial protection}{What must be known about A, and partial protection}}

\begin{proposition}[Knowledge of \(A\)]
\label{prop:unknown-A}
\begin{enumerate}[label=(\roman*),leftmargin=*,itemsep=1pt,topsep=2pt]
  \item If a random probability vector satisfies \(aQ=ap\) almost surely for
  every row vector \(a\) in a set \(\mathcal A\), it does so for every \(a\)
  in the span of \(\mathcal A\).  If this span is \(\R^{1\times k}\), then
  \(Q=p\) almost surely.
  \item If every row of \(A\) is a linear combination of the rows of \(S\)
  and \(\ones^\top\), then \(C_{p\mid S}\psdle\Cpa\).  Put
  \(d_S=\rank(SC_pS^\top)\ge d\).  For \(d_S+1\le m\le k\), take \(Q=p\)
  when \(m=k\); otherwise use the law of \cref{thm:main} for \(S\).
  In both cases, \(\Cov(Q)\psdle h_{k-d_S,m-d_S}\Cpa\).
  For fixed \(k\) and \(m<k\),
  \(h_{k-d,m-d}=(k-m)/((m-d)(k-d-1))\) increases with \(d<m\).
  \item The law of \cref{thm:main} with \(A=0\) satisfies
  \(\E\|A(Q-p)\|_2^2\le h_{k,m}\tr(A\Cp A^\top)\) for every \(A\).
  \item If \(A'\) consists of some rows of \(A\) and
  \(d'=\rank(A'\Cp A'^\top)\), the law of \cref{thm:main} for \(A'\)
  satisfies \(A'Q=A'p\) pathwise and
  \(\E\|A(Q-p)\|_2^2\le h_{k-d',m-d'}\tr(AC_{p\mid A'}A^\top)\).
\end{enumerate}
\end{proposition}

\begin{proof}
(i)~Linear constraints that hold almost surely are closed under linear
combinations, and \(a=e_i^\top\) gives \(Q_i=p_i\).  (ii)~If the protected
statistics of \(A\) are linear functions of those of \(S\) and of the
constant, the minimum in \cref{prop:comparator}(a) for \(S\) runs over a
larger set, so \(g^\top C_{p\mid S}g\le g^\top\Cpa g\) for every \(g\).
If \(m=k\), then \(Q=p\) gives the bound, including \(d_S=k-1\) with
\(h_{1,1}=0\).  Otherwise \(d_S\le m-1\le k-2\), so \cref{thm:main}
applies to \(S\); its law also protects \(A\), giving the bound.  The
denominator \((m-d)(k-d-1)\) decreases in \(d\).  (iii) and (iv) follow from
\(\E\|A(Q-p)\|_2^2=\tr(A\Cov(Q)A^\top)\) and \cref{thm:main}.
\end{proof}

\subsection{\texorpdfstring{The three-coordinate example of \cref{rem:pointwise}}{The three-coordinate example}}

With
\[
 k=3,\quad p=(1/3,1/3,1/3),\quad A=(0,1,3),\quad m=2,
\]
the two endpoints of the slice are
\(q^{(1)}=(5/9,0,4/9)\) and \(q^{(2)}=(0,5/6,1/6)\).  The only centered
endpoint law uses probabilities \((3/5,2/5)\).  For \(w=(2,-3,1)^\top\),
\[
 \Cov(Q)=\frac1{54}ww^\top,
 \qquad C_{p\mid A}=\frac1{42}ww^\top,
 \qquad \Cov(Q)=\frac79C_{p\mid A},
\]
whereas \(h_{2,1}=1\).

\section{Rational sampler details}
\label{app:sampler}

\paragraph{Roadmap.}
\Cref{alg:path} is the procedure of \cref{thm:sampler}.  The proof has three
layers.  \Cref{lem:rational-categorical,lem:rational-radial} show that a
single radial response is an explicit rational law that can be sampled
exactly.  \Cref{lem:rational-minimax} turns radial responses into a law that
is valid for all test matrices, by a rational ellipsoid method that queries
the radial response at each center.  \Cref{lem:rational-reset,lem:residual-floor}
keep the bit length of the state bounded at a small, controlled covariance
cost.  The proof of \cref{thm:sampler} at the end of this appendix adds the
two error sources to the recursion of \cref{thm:main}.  The parameters are
\(\theta=\eps/8\) (accuracy of each minimax step), \(R=k-m\) (maximal number
of steps), \(\beta=\eps h_{K,M}/(4R)\) (covariance added by one reset), and
the mesh \(1/N\) with \(N=\lceil(k+1)/(\beta\underline\lambda)\rceil\), where
\(\underline\lambda\) is a computable lower bound on the nonzero spectrum of
\(\Cpa\).

\begin{algorithm}[!htbp]
\caption{Implicit protected random path (one draw)}
\label{alg:path}
\begin{algorithmic}[1]
\REQUIRE Rational \((p,A)\), budget \(m\), accuracy \(\eps\in(0,1]\)
\STATE Compute \(d=\rank(A\Cp A^\top)\), \(K=k-d\), \(M=m-d\).
\IF{\(m=k\)} \RETURN \(p\) \ENDIF
\IF{\(M=1\)}
  \STATE Decompose \(p\) into basic feasible vertices of the slice
  (\cref{lem:rational-reset}), sample one vertex exactly with fair bits
  (\cref{lem:rational-categorical}), \mbox{and \textbf{return} it.}
\ENDIF
\STATE \(R\leftarrow k-m\), \(\theta\leftarrow\eps/8\),
  \(\beta\leftarrow\eps h_{K,M}/(4R)\).
\STATE Compute a rational basis \(T_0\) of \(\ker\Ahat\) and a
  rational \(\underline\lambda>0\) with
  \(\Cpa\succeq\underline\lambda P_{\range(T_0)}\) (\cref{lem:residual-floor}).
\STATE \(N\leftarrow\lceil(k+1)/(\beta\underline\lambda)\rceil\),
  \(q\leftarrow p\).
\WHILE{\(|\supp(q)|>m\)}
  \STATE \textit{Minimax step:} run the rational ellipsoid and
  column-generation routine of \cref{lem:rational-minimax} with the radial
  response of \cref{lem:rational-radial} to obtain a \((1+\theta)\)-valid
  one-step mixture of boundary laws.
  \STATE Sample the mixture index and then the radial endpoint exactly with
  fair bits, and set \(q\) to the endpoint (one coordinate becomes zero).
  \STATE \textit{Reset:} decompose \(q\) into at most \(k\) basic feasible
  vertices of its face and replace it by the conditionally unbiased
  mesh-\(1/N\) systematic reset of \cref{lem:rational-reset}.
\ENDWHILE
\RETURN \(q\)
\end{algorithmic}
\end{algorithm}
\FloatBarrier

This appendix proves the bit-model claims in \cref{thm:sampler}.  All
arithmetic is exact over the rationals.  Randomness consists only of fair
bits, so expected polynomial time does not mean a worst-case bound on the
number of random bits.

\subsection{Exact categorical choices and radial responses}

\begin{lemma}[Exact rational categorical draw]
\label{lem:rational-categorical}
For rational probabilities
\(\rho_j=a_j/b_j\ge0\), \(b_j>0\), with \(\sum_{j=1}^J\rho_j=1\), there is
a fair-bit Las Vegas algorithm that returns \(j\) with probability exactly
\(\rho_j\).  Its expected bit and arithmetic cost is polynomial in the
aggregate encoding length of the distribution.
\end{lemma}

\begin{proof}
Take the unreduced common denominator
\[
 D=\prod_{j=1}^J b_j,\qquad A_j=a_jD/b_j.
\]
Then \(\sum_jA_j=D\) and \(\log_2D\le\sum_j\log_2b_j\).  For \(D>1\), draw
\(\ell=\lceil\log_2D\rceil\) bits as a uniform integer
\(U\in\{0,\ldots,2^\ell-1\}\), reject when \(U\ge D\), and otherwise return
the interval containing \(U\) among the cumulative \(A_j\)'s.  Acceptance
has probability \(D/2^\ell>1/2\).  Thus fewer than two trials are needed in
expectation, and the accepted output has the prescribed law exactly.  The
case \(D=1\) is deterministic.
\end{proof}

\begin{lemma}[Rational radial response]
\label{lem:rational-radial}
Let \(q\in\Q_{>0}^{s}\), let \(T\in\Q^{s\times r}\) have full column rank
with \(\ones^\top T=0\) and \(AT=0\), and write \(b_i^\top\) for row \(i\)
of \(T\).  Put
\[
 G=T^\top\diag(q)^{-1}T.
\]
For rational \(H\succ0\), omit zero rows and define
\[
 u_i=H^{-1}b_i,\qquad
 \lambda_i=\max_{j\in\supp(q)}
       \frac{b_j^\top u_i}{q_j},\qquad
 z_i=-u_i/\lambda_i,\qquad
 \pi_i=\frac{\lambda_i}{\sum_{h:b_h\ne0}\lambda_h}.
\]
Then \(q+Tz_i\) is a rational boundary point of the same protected slice,
\(\sum_i\pi_i z_i=0\), and, for
\(V_H=\sum_i\pi_i z_iz_i^\top\),
\begin{equation}
 \langle H,V_H\rangle
 \le\frac1{\tr(H^{-1}G)}
 \le\frac{\tr(HG^{-1})}{r^2}.
 \label{eq:app-radial-response}
\end{equation}
The law and all its probabilities have encoding length polynomial in
\(\operatorname{bits}(q,T,H)\) and are computable in rational polynomial
time.
\end{lemma}

\begin{proof}
The \(j=i\) term shows \(\lambda_i>0\).  Maximality of \(\lambda_i\) gives
\(q_j+b_j^\top z_i\ge0\) for every \(j\), with equality for a maximizing
coordinate.  Moreover,
\[
 \sum_i\pi_i z_i
 =-\frac{H^{-1}}{\sum_h\lambda_h}\sum_i b_i=0.
\]
Writing \(Z=\sum_i\lambda_i\), the same maximality gives
\[
 \langle H,V_H\rangle
 =\frac1Z\sum_i\frac{b_i^\top H^{-1}b_i}{\lambda_i}
 \le\frac1Z,
 \qquad
 Z\ge\sum_i\frac{b_i^\top H^{-1}b_i}{q_i}
 =\tr(H^{-1}G).
\]
The second inequality in \cref{eq:app-radial-response} is the
arithmetic-harmonic mean inequality applied to
\(H^{1/2}G^{-1}H^{1/2}\).  Fraction-free elimination computes
\(H^{-1}b_i\), and exact cross multiplication finds every maximum.
Determinant bounds for rational linear systems give polynomial output
height.  Finally, \cref{lem:rational-categorical} samples the law exactly.
\end{proof}

\subsection{A rational spectrahedral minimax lemma}

\begin{lemma}[Rational one-step minimax]
\label{lem:rational-minimax}
Let \(r\ge1\), \(G\in\Q^{r\times r}\) be symmetric positive definite, and
\(0<\theta\le1\) be rational.  Set
\[
 c=\frac1{r^2},\qquad
 \delta=\frac{\theta}{4r},\qquad
 \gamma=\frac{\theta c}{32},
\]
\[
 \mathcal A_G=\{X\in\mathbb S^r:\langle G,X\rangle=1\},
 \qquad
 \mathcal K_\delta
 =\{X\in\mathcal A_G:X\succeq\delta G^{-1}\}.
\]
Suppose a rational query \(X\in\mathcal K_\delta\) returns, in time and
height polynomial in the static input and \(\operatorname{bits}(X)\), a
rational \(V_X\succeq0\) such that
\begin{equation}
 \langle X,GV_XG\rangle\le c.
 \label{eq:oracle-payoff}
\end{equation}
Then a deterministic rational Turing algorithm makes polynomially many
queries and returns rational weights \(\alpha_j\) and queried responses such
that
\begin{equation}
 \bar V=\sum_j\alpha_jV_{X_j}
 \preceq (1+\theta)cG^{-1}.
 \label{eq:rational-minimax-bound}
\end{equation}
The number of calls, runtime, and output height are polynomial in the static
input length and \(\operatorname{bits}(\theta)\).
\end{lemma}

\begin{proof}
The case \(r=1\) has
\(\mathcal K_\delta=\{G^{-1}\}\), so one oracle call gives
\(V_X\preceq G^{-1}\).  Assume \(r\ge2\) and let
\[
 D=\frac{r(r+1)}2-1,\quad
 U=\max_i\sum_j|G_{ij}|,\quad
 \ell=\frac{\det G}{U^{r-1}},\quad
 X_0=\frac1rG^{-1}.
\]
The positive rationals \(U,\ell\) obey
\(\lambda_{\max}(G)\le U\) and \(\lambda_{\min}(G)\ge\ell\).  In the
\(D\)-dimensional affine space \(\mathcal A_G\), with relative Frobenius
balls,
\begin{equation}
 B_{\mathcal A_G}(X_0,\rho_0)
 \subseteq\mathcal K_\delta
 \subseteq B_{\mathcal A_G}(X_0,R_0),
 \qquad
 \rho_0=\frac{1/r-\delta}{2U},\quad R_0=\frac2\ell .
 \label{eq:minimax-balls}
\end{equation}
Indeed, a tangent perturbation \(Z\) satisfies
\(\langle G,Z\rangle=0\), and the first inclusion follows from
\[
 X_0+Z-\delta G^{-1}
 \succeq\left((1/r-\delta)/U-\|Z\|_F\right)I,
\]
and the second follows by conjugating \(X\) with \(G^{1/2}\), obtaining a
positive semidefinite matrix of trace one.  A rational basis for
\(\{Z:\langle G,Z\rangle=0\}\) is obtained by eliminating any nonzero
coordinate of the vectorization of \(G\).  Its Gram matrix and inverse have
polynomial height, so \cref{eq:minimax-balls} yields rational coordinate
radii with polynomial logarithmic ratio.

For rational \(X\in\mathcal A_G\), fraction-free
\(LDL^\top\)/Bareiss elimination either certifies
\(M=X-\delta G^{-1}\succeq0\) or returns a rational \(v\) with
\(v^\top Mv<0\).  Besides a negative pivot, a zero diagonal with nonzero
off-diagonal \(a=M_{ij}\) and \(b=M_{jj}\ge0\) is exposed by
\(v=e_i-ae_j/(b+1)\), for which
\(v^\top Mv=-a^2(b+2)/(b+1)^2<0\).  Schur-complement witnesses lift by
rational back substitution.  The latter gives the exact separating
inequality
\[
 \langle vv^\top,Y\rangle
 \ge\delta v^\top G^{-1}v,\qquad Y\in\mathcal K_\delta.
\]
All pivots and lifted negative-curvature vectors are quotients of input
minors and therefore have polynomial height.

Write \(S_X=GV_XG\succeq0\).  From
\cref{eq:oracle-payoff} and \(X\succeq\delta G^{-1}\),
\begin{equation}
 \tr(GV_X)\le c/\delta,\qquad
 \|S_X\|_F\le \mathsf L,\qquad
 \mathsf L=\frac{Uc}{\delta}.
 \label{eq:column-width}
\end{equation}
Run the rational outward-rounded central-cut ellipsoid method
\citep{groetschel1988geometric} in \(\mathcal A_G\).  Besides PSD cuts,
retain each oracle column with its fixed-gap constraint
\begin{equation}
 \mathcal H_j=\{X\in\mathcal A_G:
                    \langle X,S_j\rangle\ge c+2\gamma\}.
 \label{eq:retained-gap}
\end{equation}
At a center satisfying the spectrahedron and all retained constraints,
query the oracle and add its new column, and use only the central relaxation
\(\langle Y-X,S_j\rangle\ge0\) for the ellipsoid update.  Before adding it,
test exactly whether \(S_j\in\operatorname{span}\{G\}\).  In that special
case its payoff is constant on \(\mathcal A_G\), is at most \(c\) at the
query, and the single column is already valid.  Otherwise its relative cut
normal is nonzero.

For the volume certificate define
\begin{equation}
 a_0=\frac{1/r-\delta}{U},\qquad
 \eta_0=\frac{\gamma}{2(c+4\gamma)},\qquad
 \rho_*=\min\left\{\frac{\eta_0a_0}{2},
                         \frac{\gamma}{2\mathsf L}\right\}.
 \label{eq:volume-radius}
\end{equation}
If some \(X^*\in\mathcal K_\delta\) met every retained column at level
\(c+4\gamma\), then
\(\widehat X=(1-\eta_0)X^*+\eta_0X_0\) would have PSD slack at least
\(\eta_0a_0I\) and payoff at least \(c+7\gamma/2\).  Hence the relative
\(\rho_*\)-ball about \(\widehat X\) would satisfy all constraints in
\cref{eq:retained-gap}.  After
\[
 O\!\left(D^2\log\frac{R_0}{\rho_*}\right)
\]
rounded central cuts, the containing ellipsoid has smaller volume than that
ball, a contradiction.  Consequently the polynomially many retained
columns satisfy
\begin{equation}
 \max_{X\in\mathcal K_\delta}\min_j\langle X,S_j\rangle
 \le c+4\gamma.
 \label{eq:finite-game}
\end{equation}
The rounded central-cut ellipsoid method of
\citet[Thm.~3.2.1]{groetschel1988geometric}, run in the \(D\)-dimensional
coordinates of \(\mathcal A_G\) with outer radius \(R_0\) and volume target
\(\operatorname{vol}B_{\mathcal A_G}(\,\cdot\,,\rho_*)\), fixes its
iteration count \(N_{\rm ell}=O(D^2\log(R_0/\rho_*))\) and its truncation
precision \(O(N_{\rm ell})\) binary digits from these two rationals alone.
Both are polynomial in the input length because \(\log(R_0/\rho_*)\) is
polynomial in \(\operatorname{bits}(G,\theta)\).  Thus stored centers and
shape matrices have one static polynomial height bound.  The raw height of a
processed cut affects runtime, not the precision of subsequent centers.

Finite minimax turns \cref{eq:finite-game} into a rational SDP.  For
\(S(\alpha)=\sum_j\alpha_jS_j\) and
\(\tau=1-r\delta\),
\[
 \max_{X\in\mathcal K_\delta}\langle X,S(\alpha)\rangle
 =\delta\langle G^{-1},S(\alpha)\rangle+
   \tau\lambda_{\max}(G^{-1/2}S(\alpha)G^{-1/2}),
\]
so it suffices to solve
\begin{equation}
\begin{aligned}
 \min_{\alpha,\mu}\quad&
 \delta\langle G^{-1},S(\alpha)\rangle+\tau\mu\\
 \text{s.t.}\quad&
 \alpha\ge0,\quad\ones^\top\alpha=1,\quad
 \mu G-S(\alpha)\succeq0.
\end{aligned}
\label{eq:finite-rational-sdp}
\end{equation}
If \(N\) columns were retained, strict Slater data are
\(\alpha_j^0=1/N\) and \(\mu^0=2\mathsf L/\ell\), and the LMI slack is at least
\(\mathsf L I\).  An optimum has \(\mu\le\mathsf L/\ell\), so adding
\(0\le\mu\le3\mathsf L/\ell\) makes the feasible set bounded.

For completeness, an explicit inner radius for the target with objective at
most \(c+5\gamma\) is obtained by mixing an optimum with this Slater point
by
\[
 \zeta=\min\left\{\frac14,\,
 \frac{\gamma}{4(1+c+4\gamma+3\mathsf L/\ell)}\right\},
\]
and taking
\begin{equation}
 \sigma=\min\left\{
 \frac{\zeta}{2N},\
 \frac{\zeta\mathsf L}{2(U+N\mathsf L)},\
 \frac{\gamma}{4(1+cN)},\
 \frac{\zeta\mathsf L}{\ell},\
 \frac{\mathsf L}{2\ell}
 \right\}.
 \label{eq:sdp-inner-radius}
\end{equation}
The five terms preserve, respectively, positive weights, half the LMI
slack, the objective slack, and the lower and upper bounds on \(\mu\).
After eliminating one simplex coordinate, a safe inner radius is
\(\rho_{\rm SDP}=\sigma/N\), while an outer radius is
\(R_{\rm SDP}=2+3\mathsf L/\ell\).  Rational weak optimization with exact
scalar and \(LDL^\top\) separation therefore returns an exactly feasible
rational \((\alpha,\mu)\) of polynomial height and objective at most
\(c+5\gamma\).

Finally put \(W=G^{1/2}\bar VG^{1/2}\).  The SDP objective bound implies
\[
 (1-r\delta)\lambda_{\max}(W)\le c+5\gamma,
\]
and hence
\[
 \lambda_{\max}(W)
 \le\frac{1+5\theta/32}{1-\theta/4}\,c
 \le(1+\theta)c.
\]
This is exactly \cref{eq:rational-minimax-bound}.  Every parameter and
radius displayed above is rational and has polynomial binary length, which
also proves the claimed Turing bounds.
\end{proof}

To instantiate \cref{lem:rational-minimax} at a rational state of the slice,
use \(H=GXG\) in \cref{lem:rational-radial}.  The floor
\(X\succeq\delta G^{-1}\) is what makes \(H\succ0\) there.  Since
\(\tr(HG^{-1})=\langle G,X\rangle=1\),
\[
 \langle X,GV_HG\rangle=\langle H,V_H\rangle\le1/r^2.
\]
The radial height bound supplies the oracle-height hypothesis.  Mixing the
queried radial laws preserves the exact conditional mean and protected
slice, and every outcome loses a support coordinate.

\subsection{A bounded-height unbiased reset}

\begin{lemma}[Rational face reset]
\label{lem:rational-reset}
On a fixed support \(S\), let
\[
 F_S=\{x\in\R_+^S:\ones^\top x=1,\ Ax=Ap\}.
\]
Given \(x\in F_S\cap\Q^S\), one can compute in rational polynomial time
\[
 x=\sum_{j=1}^J\lambda_jv^j,\qquad J\le|S|,
\]
where the \(v^j\) are basic feasible vertices of \(F_S\) and
\(\supp(v^j)\subseteq\supp(x)\).  For any integer \(N\ge1\), there is an
exact rational random reset \(\widetilde x\) satisfying
\begin{equation}
 \E[\widetilde x\mid x]=x,\qquad
 \supp(\widetilde x)\subseteq\supp(x),\qquad
 \|\widetilde x-x\|_2\le J/N\le k/N.
 \label{eq:reset-properties}
\end{equation}
Every realized reset has coordinate height at most
\[
 B_{\rm reset}\le\log N+JB_{\rm vert}+O(J\log J),
\]
where \(B_{\rm vert}=\poly(k,a,L)\) depends only on the original rational
face description, not on the height of \(x\).
\end{lemma}

\begin{proof}
Work in the smallest face containing \(x\).  Choose a basic feasible vertex
\(v\), extend the ray from \(v\) through \(x\) to the next boundary, and
recurse on the lower-dimensional face.  At most \(|S|-1\) dimension drops
give at most \(|S|\) vertices.  Exact ratio tests implement every step.
More explicitly, at one drop write the current point as \(y\), and let
coordinate \(i\) be the first coordinate hit by the ray.  The new boundary
point has coordinates
\[
 y'_j=\frac{v_i y_j-y_i v_j}{v_i-y_i}.
\]
Writing \(v=V/D_v\) and \(y=Y/D_y\) with shared denominators, the factor
\(D_vD_y\) cancels before reduction and all \(y'_j\) share denominator
\(V_iD_y-Y_iD_v\).  Thus
\(B(y')\le B(y)+B(v)+O(1)\): one drop adds only polynomially many bits, and
the at-most-\(|S|-1\) drops do not create a repeated-squaring height
recurrence.
After redundant equality rows are removed, Cramer's rule applied to
\(\Ahat v=(1,(Ap)^\top)^\top\) gives the uniform bound \(B_{\rm vert}\).

Put \(a_j=\lfloor N\lambda_j\rfloor\),
\(f_j=N\lambda_j-a_j\), \(F_0=0\), and
\(F_j=\sum_{\ell\le j}f_\ell\).  For \(u\in[0,1)\), define
\[
 \xi_j(u)=\lfloor F_j+u\rfloor-\lfloor F_{j-1}+u\rfloor.
\]
Here
\[
 h_0:=F_J=\sum_j f_j=N-\sum_j a_j\in\mathbb Z,
 \qquad
 \xi_j(u)\in\{0,1\},
 \qquad
 \sum_j\xi_j(u)=h_0,
\]
and \(\int_0^1\xi_j(u)\,du=f_j\).  Consequently the coefficients
\((a_j+\xi_j)/N\) are nonnegative and sum to one for every realized
\(u\), hence \(\widetilde x\) below remains in \(F_S\).
This vector changes only at the rational breakpoints
\((-F_j)\bmod1\), so \cref{lem:rational-categorical} samples one of its at
most \(J\) constant intervals by exact rational length.  Since
\(\E\xi_j=f_j\), the random point
\[
 \widetilde x=\sum_j\frac{a_j+\xi_j}{N}v^j
\]
is conditionally unbiased.  Also
\(\sum_j|a_j+\xi_j-N\lambda_j|\le J\), and since
\(\|v^j\|_2\le\|v^j\|_1=1\), this proves
\cref{eq:reset-properties}.  Its realized coefficients have denominator
\(N\), and incoming decomposition denominators occur only in the categorical
probabilities.  This gives the stated height bound.
\end{proof}

\subsection{Spectral and covariance ledgers}

\begin{lemma}[Computable residual spectral floor]
\label{lem:residual-floor}
Let \(r_0\ge1\), and let \(T_0\in\Q^{k\times r_0}\) be a full-column-rank basis of the global
slice direction space, and put
\[
 R_T=T_0^\top T_0,\qquad
 G_0=T_0^\top\diag(p)^{-1}T_0,\qquad
 C_0=T_0G_0^{-1}T_0^\top=\Cpa.
\]
With
\[
 \Lambda_T=\max_i\sum_j|(R_T)_{ij}|,\qquad
 \Lambda_G=\max_i\sum_j|(G_0)_{ij}|,
\]
define
\[
 \underline\lambda
 =\min\left\{1,\frac{\det R_T}
 {\Lambda_T^{r_0-1}\Lambda_G}\right\}.
\]
Then \(C_0\succeq\underline\lambda P_{\range(T_0)}\), and
\(\log(1/\underline\lambda)\) is polynomial in the rational input length.
If \(\widetilde x\) is the reset of \cref{lem:rational-reset} with
\[
 N=\left\lceil\frac{k+1}{\beta\underline\lambda}\right\rceil,
 \qquad 0<\beta\le1,
\]
then
\begin{equation}
 \Cov(\widetilde x\mid x)\preceq\beta C_0.
 \label{eq:reset-covariance}
\end{equation}
\end{lemma}

\begin{proof}
The nonzero eigenvalues of \(C_0\) are those of
\(G_0^{-1/2}R_TG_0^{-1/2}\).  Gershgorin's bound and the determinant identity
give
\[
 \lambda_{\min}^{+}(C_0)
 \ge\frac{\lambda_{\min}(R_T)}{\lambda_{\max}(G_0)}
 \ge\frac{\det R_T}{\Lambda_T^{r_0-1}\Lambda_G}.
\]
Clearing rational denominators makes the nonzero determinant an integer
divided by a denominator of polynomial bit length, proving the height
claim.  Reset errors lie in \(\range(T_0)\), so
\[
 \Cov(\widetilde x\mid x)
 \preceq\frac{k^2}{N^2}P_{\range(T_0)}
 \preceq\frac{k^2}{N^2\underline\lambda}C_0
 \preceq\beta C_0.
\]
\end{proof}

\paragraph{Proof of \cref{thm:sampler}.}
If \(m=k\), return \(p\) before dividing by \(R=k-m\).  If \(K=1\), this is
the only feasible case.  If \(M=1\), use the vertex decomposition in
\cref{lem:rational-reset} without rounding and sample one basic feasible
vertex.  Since
\[
 \rank\begin{bmatrix}\ones^\top\\A\end{bmatrix}=d+1,
\]
every such vertex has support at most \(d+1=m\), and the bridge
\cref{lem:bridge} gives \(\Cov(Q)\preceq C_0=h_{K,1}C_0\).

Now assume \(m<k\) and \(M\ge2\).  Set
\[
 R=k-m,\qquad \theta=\eps/8,\qquad
 \beta=\frac{\eps h_{K,M}}{4R}.
\]
Here \(K>M\ge2\), so the global tangent rank is \(r_0=K-1\ge2\).
Choose the global reset denominator from
\cref{lem:residual-floor}.  At every active state, instantiate
\cref{lem:rational-minimax} with the rational radial oracle, sample its two
categorical levels using \cref{lem:rational-categorical}, and then apply the
reset.  A radial outcome drops support and a reset never enlarges it, so
every path has at most \(R\) active stages.  At support \(t>m\), with
effective protected rank \(d_t\le d\), the tangent dimension obeys
\[
 r_t=t-d_t-1\ge t-d-1\ge M\ge2.
\]

Let \(j=t-d-1\).  The one-step coefficient is at most
\(\alpha_j=(1+\theta)/j^2\).  Total covariance and
\cref{lem:bridge,eq:reset-covariance} give the scalar ledger
\[
 \eta'\le\alpha_j+(1-\alpha_j)\eta+\beta.
\]
Skipped support levels can be padded by identity transitions.  Without
resets, iteration from \(j=K-1\) to \(M\) gives
\[
 \widehat h
 =1-\prod_{j=M}^{K-1}\left(1-\frac{1+\theta}{j^2}\right),
 \qquad
 h_{K,M}
 =1-\prod_{j=M}^{K-1}\left(1-\frac1{j^2}\right).
\]
Telescoping the difference and using
\(\sum_{j=M}^{K-1}j^{-2}\le2h_{K,M}\) yields
\[
 \widehat h-h_{K,M}\le2\theta h_{K,M}
 =\frac{\eps}{4}h_{K,M}.
\]
The at most \(R\) resets add at most
\(R\beta C_0=\eps h_{K,M}C_0/4\).  Therefore
\[
 \Cov(Q)\preceq(1+\eps/2)h_{K,M}C_0
 \preceq(1+\eps)h_{K,M}C_0.
\]

After every reset the realized state has encoding length
\(\poly(k,a,L+\enc(\eps))\), independently of the incoming denominator.
\Cref{lem:rational-minimax} then gives one uniform polynomial height bound
for the next transition.  There are at most \(k\) stages, and every
categorical draw has expected polynomial cost by
\cref{lem:rational-categorical}.  Thus the total expected runtime and every
realized output length have the bounds stated in \cref{thm:sampler}.  The
finite terminal law may still have exponentially many leaves.

\section{Query-aware complexity proofs}
\label{app:hardness}

\Cref{tab:interfaces} summarizes what \cref{thm:minuncut} and this appendix
establish for the query-aware problem, by output interface and instance
family.

\begin{table}[h]
\centering
\footnotesize
\setlength{\tabcolsep}{2pt}
\caption{Established query-aware complexity by output interface.  ``No
PTAS'' and ``no FPTAS'' entries assume \(\mathsf P\ne\mathsf{NP}\), and
``open'' marks a claim outside the present reductions.  The columns are the
pair-protected PSD \(W_G\) of \cref{thm:minuncut}, the pair-protected
positive definite \(W_{G,\eta}\) with polynomial condition number
(\cref{app:pd-variant}), and the unprotected case \(A=0\) with near-identity
\(W\) (\cref{lem:direct-a0-gap}).}
\label{tab:interfaces}
\begin{tabular}{@{}lccc@{}}
\toprule
Output interface & Pair-prot., PSD & Pair-prot., PD, poly-cond. & \(A=0\), near-identity PD \\
\midrule
Exact threshold decision & NP-hard & NP-hard (known offset) & NP-hard \\
Additive value estimate & NP-hard below \(2/n^2\) & NP-hard (known offset) & NP-hard below \(1/k^2\) \\
Explicit-law approximation & no PTAS & no PTAS & no FPTAS, PTAS open \\
Value estimate, fixed ratio & open & open & open \\
Black-box sampler & needs randomized assumpt. & open & open \\
\bottomrule
\end{tabular}
\end{table}

\subsection{\texorpdfstring{Proof of \cref{thm:minuncut}}{Proof of the Minimum-Uncut theorem}}

Index the \(2n\) coordinates by \((i,+),(i,-)\), set
\(p=\ones/(2n)\), and let the rows of \(A\) be the first \(n-1\) pair sums
\(q_{(i,+)}+q_{(i,-)}\).  The simplex equation supplies the last, so
\(d=n-1\) and every feasible atom gives each pair mass \(1/n\).  Because a
feasible atom has support at most \(n\), it selects exactly one coordinate
per pair, with mass \(1/n\).  The feasible sparse atoms are therefore
\(\{q_z:z\in\{-1,+1\}^n\}\) with \(Dq_z=z/n\), where \(D\) is the
pair-difference matrix, and
\begin{equation}
 \tfrac12q_z+\tfrac12q_{-z}=p .
 \label{eq:complement-law}
\end{equation}
Let \(\mathcal S_G=\operatorname{Deg}(G)+\operatorname{Adj}(G)\) be the
signless Laplacian and set \(W_G=D^\top\mathcal S_GD\).  Since \(Dp=0\),
\begin{equation}
 (q_z-p)^\top W_G(q_z-p)
 =\frac1{n^2}z^\top\mathcal S_Gz
 =\frac1{n^2}\sum_{\{i,j\}\in E}(z_i+z_j)^2
 =\frac{4\uncut_G(z)}{n^2},
 \label{eq:atom-uncut}
\end{equation}
where \(\uncut_G(z)\) counts edges with \(z_i=z_j\).  Every atom costs at
least the right side of \cref{eq:minuncut-identity}, and mixing a
minimum-uncut \(z\) with \(-z\) as in \cref{eq:complement-law} is a feasible
law attaining it.  For cubic \(G\), \(\|D\|_{\rm op}^2=2\) and
\(\|\mathcal S_G\|_{\rm op}\le6\).

Minimum Uncut is APX-hard on cubic graphs, by an L-reduction from cubic
Max-Cut \citep[Obs.~1]{cilibrasi2007haplotyping}, itself APX-hard
\citep{alimonti2000some}.  An explicit-law algorithm outputs a rational
\(\sum_t\lambda_t\delta_{q_{z_t}}\) of polynomial total encoding length with
exact pathwise feasibility and exact barycenter.  If its cost is within ratio
\(\rho\) of \(\OPT_{W_G}\), scanning the polynomial-size support and returning
the cheapest atom gives
\[
 \min_t\uncut_G(z_t)
 \le \sum_t\lambda_t\uncut_G(z_t)
 \le \rho\,\MinUncut(G).
\]
Conversely every cut yields the two-atom law \cref{eq:complement-law}.
Approximation ratios therefore transfer exactly in both directions, and
bipartite inputs (zero optimum) are recognized separately.

\subsection{Positive-definite protected variant}
\label{app:pd-variant}

Use the pair-sum and pair-difference matrices \(B,D\) from
\cref{thm:minuncut} and let
\[
 W_{G,\eta}=B^\top B+D^\top(\mathcal S_G+\eta I)D,
 \qquad \eta=n^{-4}.
\]
For every \(x\),
\[
 x^\top W_{G,\eta}x
 =\|Bx\|_2^2+(Dx)^\top(\mathcal S_G+\eta I)(Dx),
\]
which is zero only if \(Bx=Dx=0\), hence \(x=0\).  The spectrum is
\[
 \{2\}_{i=1}^n
 \cup\{2(\lambda_i(\mathcal S_G)+\eta)\}_{i=1}^n.
\]
For cubic \(G\),
\(\kappa_2(W_{G,\eta})\le(6+\eta)/\eta=\poly(n)\).  For every feasible
sparse atom \(q_z\), \(B(q_z-p)=0\) and \(\|Dq_z\|_2^2=1/n\), so
\begin{equation}
 \OPT_{W_{G,\eta}}(p,A,n)
 =\frac\eta n+\frac{4\MinUncut(G)}{n^2}.
 \label{eq:pd-offset}
\end{equation}
If a feasible explicit law has average uncut count \(\bar U\) and ratio
\(r\), then
\[
 \bar U\le rU(G)+(r-1)\frac{\eta n}{4}.
\]
Bipartite inputs can be recognized separately.  On nonbipartite inputs
\(U(G)\ge1\), and the additive ratio loss for \(\eta=n^{-4}\) is at most
\((r-1)/(4n^3)\).  A PTAS for this polynomially conditioned positive-definite
explicit-law family would therefore give a PTAS for Cubic Minimum Uncut.  The
offset construction is not ratio-lossless and does not have a uniformly
bounded condition number.

\subsection{Direct unprotected additive hardness}

\begin{lemma}[Direct unprotected additive gap]
\label{lem:direct-a0-gap}
Let \(G\) be a simple unweighted graph on an even number \(k\ge4\) of
vertices.  Set
\[
 m=k/2,\qquad p=\ones/k,\qquad
 \Lambda=8k^3,\qquad W=\Lambda I+\mathcal L_G .
\]
Then \(W\succ0\),
\[
 \kappa_2(W)
 \le 1+\frac{2(k-1)}{\Lambda}
 <1+\frac1{4k^2},
\]
and, writing \(b(G)\) for the minimum-bisection value,
\begin{equation}
 \frac{\Lambda}{k}+\frac{4b(G)}{k^2}-\frac1{2k^2}
 \le \OPT_W(p,0,k/2)
 \le \frac{\Lambda}{k}+\frac{4b(G)}{k^2}.
 \label{eq:bisection-sandwich}
\end{equation}
Consequently, exact threshold comparison and deterministic
\(1/k^2\)-additive evaluation are NP-hard on this family.  Moreover, there
is no deterministic FPTAS for a certified explicit feasible law, or for a
numerical value under the standard relative-error guarantee, unless
\(\mathsf P=\mathsf{NP}\).
\end{lemma}

\begin{proof}
For the upper bound, let \(I\) be a minimum bisection and define
\[
 u_I=\frac1m\ones_I,\qquad u_{I^c}=\frac1m\ones_{I^c}.
\]
The law assigning probability \(1/2\) to each vector is feasible and has
mean \(p\).  For \(x_I=u_I-p\),
\[
 \|x_I\|_2^2=\frac1k,\qquad
 x_I^\top\mathcal L_Gx_I=\frac{4|E(I,I^c)|}{k^2},
\]
and \(u_{I^c}-p=-x_I\).  This proves the upper bound.

For the lower bound, fix a feasible atom \(q\) and put
\(s=|\supp(q)|\le m\).  If \(s<m\), then
\[
 \|q-p\|_2^2
 =\|q\|_2^2-\frac1k
 \ge\frac1s-\frac1k
 \ge\frac1k+\frac1{m(m-1)}.
\]
Since \(\mathcal L_G\succeq0\),
\[
 (q-p)^\top W(q-p)
 \ge\frac{\Lambda}{k}+\frac{\Lambda}{m(m-1)}.
\]
Here
\[
 \frac{\Lambda}{m(m-1)}
 =\frac{32k^2}{k-2}>1
 \ge \frac{4b(G)}{k^2},
\]
so this is stronger than the desired lower bound.

It remains to consider \(s=m\).  Let \(I=\supp(q)\), and write
\[
 u_I=\frac1m\ones_I,\qquad x_I=u_I-p,\qquad \Delta=q-u_I.
\]
Then \(\Delta\) is supported on \(I\), satisfies
\(\ones^\top\Delta=0\), and is orthogonal to \(x_I\).  Therefore
\(\|q-p\|_2^2=1/k+\|\Delta\|_2^2\).  Completing the square and using
\(\mathcal L_G\succeq0\) gives
\begin{align*}
 (q-p)^\top W(q-p)
 &=
 \frac{\Lambda}{k}
 +x_I^\top\mathcal L_Gx_I
 +\Lambda\|\Delta\|_2^2
 +2\Delta^\top\mathcal L_Gx_I
 +\Delta^\top\mathcal L_G\Delta \\
 &\ge
 \frac{\Lambda}{k}
 +\frac{4|E(I,I^c)|}{k^2}
 -\frac{\|\mathcal L_Gx_I\|_2^2}{\Lambda}.
\end{align*}
Because \(I\) is a bisection, \(|E(I,I^c)|\ge b(G)\).  Moreover,
\[
 \frac{\|\mathcal L_Gx_I\|_2^2}{\Lambda}
 \le
 \frac{\|\mathcal L_G\|_{\rm op}^2\|x_I\|_2^2}{\Lambda}
 \le
 \frac{4(k-1)^2/k}{8k^3}
 \le\frac1{2k^2}.
\]
Thus every feasible atom obeys the lower bound in
\cref{eq:bisection-sandwich}, and averaging preserves it.

For the complexity consequences, reduce an integer-threshold Minimum
Bisection instance \((G,C)\).  Define
\[
 a_C=\frac{\Lambda}{k}+\frac{4C}{k^2},\qquad
 c_C=\frac{\Lambda}{k}+\frac{4C+7/2}{k^2},\qquad
 t_C=a_C+\frac7{4k^2}.
\]
Since \(b(G)\) is integral,
\[
 b(G)\le C\Longrightarrow\OPT\le a_C<t_C,\qquad
 b(G)\ge C+1\Longrightarrow\OPT\ge c_C>t_C.
\]
This proves exact-threshold hardness.  If
\(|\widehat v-\OPT|\le1/k^2\), then the two cases remain separated by
\(t_C\), proving additive hardness.

Finally, \(b(G)\le k^2/4\) gives \(\OPT\le8k^2+1\).  An FPTAS called with
\(\varepsilon_{\rm F}=1/(9k^4)\) would have absolute error at most
\[
 \varepsilon_{\rm F}\OPT
 \le\frac{8k^2+1}{9k^4}\le\frac1{k^2},
\]
while \(1/\varepsilon_{\rm F}\) is polynomial in \(k\).  This contradicts
the additive-hardness consequence.
\end{proof}

This lemma does not prove a constant-factor barrier, a no-PTAS result,
hardness for diagonal \(W\), or randomized hardness under the same
assumption.

\subsection{Output interfaces}

An explicit feasible-law algorithm returns rational
\(\sum_t\lambda_t\delta_{q^t}\) of polynomial total encoding length with all
pathwise constraints and \(\sum_t\lambda_tq^t=p\) exact.  A certified feasible
upper value accompanied by such a law has the same extraction power.  A bare
number, threshold decision, and implicit sampler are different outputs.  For
the pair-protected instance, an implicit sampler PTAS would give a randomized
PTAS for Minimum Uncut by independent repetition and selection of the cheapest
sampled cut, but excluding it requires an explicit randomized-complexity
assumption rather than only \(\mathsf P\ne\mathsf{NP}\).  Constant-factor hardness at \(A=0\),
or under a uniformly bounded condition number, remains open.

\section{Fixed-deployment learning consequences}
\label{app:learning}

\paragraph{Quadratic tasks.}
After an admissible law \(\Law\) is fixed, let a future task choose points
\(V=[v_1,\ldots,v_k]\) and losses
\(f_i^V(\theta)=\frac\mu2\|\theta-v_i\|_2^2\), so that the minimizer of
\(F^V_w=\sum_iw_if_i^V\) is \(\theta_w=Vw\).  Writing
\(B_{\Law}=\Cov_{\Law}(Q)\),
\begin{equation}
 \E_{\Law}\bigl[F_p^V(\theta_Q)-F_p^V(\theta_p)\bigr]
 =\frac\mu2\tr(VB_{\Law}V^\top).
 \label{eq:quadratic-exact}
\end{equation}
Normalizing by the residual task dispersion \(\frac\mu2\tr(V\Cpa V^\top)\),
\cref{thm:main} gives
\begin{equation}
 \sup_{p,A:\,\rank(A\Cp A^\top)=d}\,
 \inf_{\Law}\,
 \sup_{V:\,\tr(V\Cpa V^\top)>0}
 \frac{\tr(VB_{\Law}V^\top)}{\tr(V\Cpa V^\top)}
 =h_{k-d,m-d}.
 \label{eq:quadratic-minimax}
\end{equation}
The law sees \((p,A)\) but not the task, and the task sees the law but not the
realized draw.  The proof is at the end of this appendix.

Let \(B=\Cov(Q)\psdle h\Cpa\) and
\(F_w(\theta)=\sum_iw_if_i(\theta)\).  Suppose every realized \(F_q\) is
\(\mu\)-strongly convex, \(F_p\) is \(\beta\)-smooth, and
\(\theta_w=\arg\min_\theta F_w(\theta)\).  At
\(\theta_p\), write
\(G_*=[\nabla f_1(\theta_p),\ldots,\nabla f_k(\theta_p)]\).  Since
\(G_*p=0\),
\(\nabla F_q(\theta_p)=G_*(q-p)\).  Strong monotonicity and
Cauchy-Schwarz give
\[
 \|\theta_q-\theta_p\|_2
 \le \mu^{-1}\|G_*(q-p)\|_2.
\]
Squaring and averaging yields
\[
 \E\|\theta_Q-\theta_p\|_2^2
 \le \frac1{\mu^2}\tr(G_*BG_*^\top)
 \le \frac h{\mu^2}\tr(G_*\Cpa G_*^\top).
\]
Smoothness of \(F_p\) gives the excess-risk bound of \cref{sec:learning}.

For a fixed draw reused along deterministic full-batch gradient descent,
assume
\(\mu I\preceq\nabla^2F_q(\theta)\preceq\beta I\) for all relevant
\((q,\theta)\), use common initialization and \(0<\eta<2/\beta\), and put
\(\rho=\max\{|1-\eta\mu|,|1-\eta\beta|\}<1\).  Along the deterministic
full-data path, let
\(G_t=[\nabla f_1(\theta_t^p),\ldots,\nabla f_k(\theta_t^p)]\).  Contractivity
of the gradient map gives
\[
 \|\theta_{t+1}^q-\theta_{t+1}^p\|_2
 \le \rho\|\theta_t^q-\theta_t^p\|_2
 +\eta\|G_t(q-p)\|_2.
\]
Unrolling and applying Minkowski in \(L_2(Q)\) yields
\[
 \left(\E\|\theta_T^Q-\theta_T^p\|_2^2\right)^{1/2}
 \le \eta\sqrt h\sum_{t<T}\rho^{T-1-t}
 \sqrt{\tr(G_t\Cpa G_t^\top)}.
\]
This argument relies on uniform strong convexity and deterministic full-batch
updates.  A PL condition alone does not control which minimizer a perturbed
objective selects, and generic nonconvex extensions require additional basin
or error-bound assumptions.

For the quadratic family in \cref{eq:quadratic-exact},
\(F_p^V(\theta_Q)-F_p^V(\theta_p)=\frac\mu2\|V(Q-p)\|_2^2\).  Maximization
over \(V\) gives the largest generalized eigenvalue of \(B\) relative to
\(\Cpa\) on \(\range(\Cpa)\).  Therefore the upper and sharpness parts of
\cref{thm:main} transfer verbatim to \cref{eq:quadratic-minimax}.  If the
denominator is zero, exact protection forces the numerator to be zero as
well.

\section{An exact conic lift}
\label{app:boundaries}

For \(A=0\), let
\(\mathcal D_{k,m}=\{q\in\simplex_k:\|q\|_0\le m\}\) and define the sparse
completely-positive cone
\[
 \mathcal{CP}^{(m)}_k
 =\operatorname{cone}\{zz^\top:z\ge0,\ \|z\|_0\le m\}.
\]
The fixed-\(W\) value has the exact lift
\begin{equation}
 \OPT_W(p,0,m)+p^\top Wp
 =\min\{\langle W,X\rangle:X\ones=p,\ X\in\mathcal{CP}^{(m)}_k\}.
 \label{eq:cp-primal}
\end{equation}
Indeed, a law maps to \(X=\E[QQ^\top]\).  Conversely, write
\(X=\sum_t z^t(z^t)^\top\), put
\(s_t=\ones^\top z^t\), \(q^t=z^t/s_t\) for nonzero factors, and
\(\lambda_t=s_t^2\).  Then \(X\ones=p\) gives
\[
 \sum_t\lambda_tq^t=p,\qquad
 \sum_t\lambda_t=\ones^\top X\ones=\ones^\top p=1,
\]
so the factors normalize into an admissible law supported on
\(\mathcal D_{k,m}\).  Its second moment is \(X\), proving
\cref{eq:cp-primal}.

This is a nonnegative-factor-width reformulation, not a new cone or a
tractability claim.  The support restriction remains encoded in the
generators of \(\mathcal{CP}^{(m)}_k\), and the hardness results in
\cref{sec:hardness,app:hardness} show why the lift alone does not yield a
polynomial-time method for general \((m,W)\).

\section{A larger protected instance}
\label{app:numerics-large}

\begin{figure}[h]
  \centering
  \includegraphics[width=0.56\textwidth]{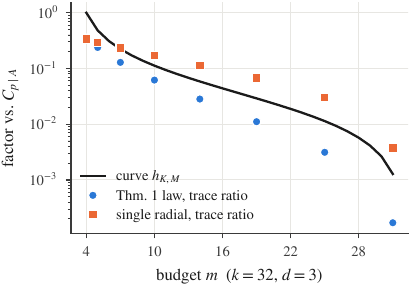}
  \caption{\textbf{Larger instance} (\(k=32\), \(d=3\), \(1000\) paths per law).
  Unbiased trace ratio \(\tr\Cov(Q)/\tr\Cpa\) against the curve.}
  \label{fig:numerics-large}
\end{figure}

\Cref{fig:numerics-large} and \cref{tab:numerics-large} run the construction
on \(k=32\) coordinates with \(d=3\) protected totals (\(K=29\)).  The weights
are a Dirichlet draw with concentration \(0.7\), whose smallest coordinate is
\(2.2\cdot10^{-4}\), and the three rows of \(A\) have entries in
\(\{0,1,2\}\), so the protected slice is a \(28\)-dimensional polytope.  Each
law is run for \(1000\) paths from \(m=k\) down to \(m=d+1\), with \(150\)
multiplicative-weights rounds per step, and every one-step certificate of the
law of \cref{thm:main} is at most \(0.9997\).  The trace ratio of the law of
\cref{thm:main} lies below the curve at every budget, at \(0.13\) to \(0.58\)
of \(h_{29,M}\).  The single radial response crosses the curve at \(m=7\) and
exceeds it by a factor \(2.9\) at \(m=31\).  With coordinates as small as
\(2\cdot10^{-4}\) the comparator is ill conditioned, and the eigenvalue
estimates from \(1000\) draws are noisy, which is why the cross-fitted value
at \(m=5\) exceeds the curve although every step is certified.

\begin{table}[h]
\centering
\footnotesize
\setlength{\tabcolsep}{4pt}
\caption{Larger instance: trace ratio, cross-fitted and plug-in largest
generalized eigenvalue relative to \(\Cpa\).}
\label{tab:numerics-large}
\begin{tabular}{rrrrrrrr}
\toprule
& & & \multicolumn{3}{c}{Thm.~1 law} & \multicolumn{2}{c}{single radial} \\
\cmidrule(lr){4-6}\cmidrule(lr){7-8}
$m$ & $M$ & $h_{29,M}$ & trace & cross-fit & plug-in & trace & cross-fit \\
\midrule
4 & 1 & 1.0000 & 0.3466 & 0.4426 & 0.7078 & 0.3388 & 0.2509 \\
5 & 2 & 0.4821 & 0.2381 & 0.7454 & 0.7803 & 0.2927 & 0.3377 \\
7 & 4 & 0.2232 & 0.1286 & 0.1230 & 0.2546 & 0.2324 & 0.3260 \\
10 & 7 & 0.1122 & 0.0618 & 0.1379 & 0.1512 & 0.1697 & 0.2594 \\
14 & 11 & 0.0584 & 0.0282 & 0.0305 & 0.0396 & 0.1142 & 0.2020 \\
19 & 16 & 0.0290 & 0.0111 & 0.0102 & 0.0161 & 0.0672 & 0.1365 \\
25 & 22 & 0.0114 & 0.0031 & 0.0034 & 0.0048 & 0.0303 & 0.0656 \\
31 & 28 & 0.0013 & 0.0002 & 0.0001 & 0.0004 & 0.0037 & 0.0097 \\
\bottomrule
\end{tabular}

\end{table}

\section{Additional experiments}
\label{app:numerics-revision}

The experiments of this appendix follow two protocols that were written
before the corresponding runs and are included with the code.  Protocol 1
covers the step rules (N3) and the partition instances (N1, N2).  Protocol 2
covers the
one-step factors (E2), the larger instance of \cref{app:numerics-large} (E3),
and the classical designs (E4).

\paragraph{Estimating a Loewner factor.}
The plug-in largest generalized eigenvalue of a sample covariance is biased
upward, by a factor close to \((1+\sqrt{r/n})^2\) for an isotropic covariance
of dimension \(r\) estimated from \(n\) draws.  The cross-fitted estimate
selects the top direction on one half of the draws and evaluates its Rayleigh
quotient on the other half.  It is unbiased for that direction and therefore
biased downward for the largest eigenvalue.  The population trace ratio
\(\tr\Cov(Q)/\tr\Cpa\) bounds the factor from below.  Its sample-covariance
estimate is unbiased, but need not lie below the factor on each run.  The
exact designs of \cref{thm:designs} need no estimate, since their laws are
computed exactly, by enumeration for the ratio design and by a dynamic program
over subsets for sequential VarOpt.

\paragraph{Step rules (N3).}
\Cref{tab:rev-N3} lists the values behind \cref{fig:numerics}(a): \(k=12\), a
Dirichlet draw with concentration \(2\), two protected rows with entries in
\(\{0,1,2\}\), and \(3000\) paths per rule.  The largest one-step certificate
of the law of \cref{thm:main} is \(0.997\), and the single radial response
reaches \(3.0\).

\begin{table}[!htbp]
\centering
\footnotesize
\setlength{\tabcolsep}{2.5pt}
\caption{Step rules on the instance of \cref{fig:numerics}(a): plug-in largest
generalized eigenvalue with \(90\%\) bootstrap interval, and trace ratio.}
\label{tab:rev-N3}
\resizebox{\textwidth}{!}{\begin{tabular}{rrlrlrlr}
\toprule
$m$ & $h_{10,M}$ & Thm.~1: $\hat\lambda_{\max}$ & trace & naive: $\hat\lambda_{\max}$ & trace & flight: $\hat\lambda_{\max}$ & trace \\
\midrule
3 & 1.0000 & 0.7043 [0.6381, 0.8242] & 0.4224 & 0.7022 [0.6021, 0.8155] & 0.4196 & 0.7162 [0.5820, 0.8526] & 0.4205 \\
4 & 0.4444 & 0.3765 [0.3496, 0.4172] & 0.2770 & 0.4312 [0.4211, 0.4441] & 0.3069 & 0.4492 [0.4004, 0.5360] & 0.3252 \\
5 & 0.2593 & 0.2117 [0.2080, 0.2323] & 0.1838 & 0.3231 [0.3143, 0.3340] & 0.2280 & 0.3276 [0.2997, 0.3854] & 0.2538 \\
6 & 0.1667 & 0.1377 [0.1379, 0.1479] & 0.1240 & 0.2398 [0.2330, 0.2476] & 0.1694 & 0.2298 [0.2234, 0.2585] & 0.1975 \\
7 & 0.1111 & 0.0896 [0.0910, 0.0968] & 0.0838 & 0.1775 [0.1711, 0.1844] & 0.1237 & 0.1728 [0.1699, 0.1874] & 0.1512 \\
8 & 0.0741 & 0.0611 [0.0613, 0.0655] & 0.0560 & 0.1270 [0.1227, 0.1312] & 0.0871 & 0.1252 [0.1256, 0.1338] & 0.1149 \\
9 & 0.0476 & 0.0375 [0.0380, 0.0406] & 0.0349 & 0.0843 [0.0812, 0.0879] & 0.0580 & 0.0922 [0.0918, 0.0981] & 0.0835 \\
10 & 0.0278 & 0.0221 [0.0217, 0.0242] & 0.0196 & 0.0518 [0.0499, 0.0539] & 0.0351 & 0.0597 [0.0602, 0.0649] & 0.0549 \\
11 & 0.0123 & 0.0093 [0.0090, 0.0107] & 0.0082 & 0.0249 [0.0238, 0.0266] & 0.0160 & 0.0316 [0.0309, 0.0341] & 0.0275 \\
\bottomrule
\end{tabular}
}
\end{table}

\begin{figure}[!htbp]
  \centering
  \includegraphics[width=0.56\textwidth]{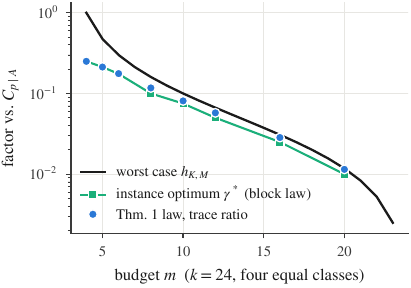}
  \caption{\textbf{Four equal classes} (\(k=24\)).  The block law of
  \cref{prop:partition} attains the instance optimum, a factor \(m/(m-c+1)\)
  below the worst case.}
  \label{fig:partition}
\end{figure}

\paragraph{Partition instances (N1, N2).}
N1 has four classes of six coordinates with uniform weights, so the instance
optimum of \cref{cor:equal-classes} is exact and the block law attains it
(\cref{fig:partition}, \cref{tab:rev-N1}).  The general law of
\cref{thm:main} is close to the optimum at \(m=4\) and, at larger budgets,
has trace ratio between the optimum and the worst case.  N2 has classes of
sizes \(1,2,3,18\) with masses \(0.02,0.08,0.15,0.75\) and Dirichlet weights
inside each class (\cref{tab:rev-N2}).  Its comparator has nonzero eigenvalues
from \(5\cdot10^{-4}\) to \(0.11\), and plug-in eigenvalues from \(2000\)
draws are not reliable there.  At \(m=4\) each class receives one slot, the
block law has covariance \(\sum_j\pi_j^2C_{p^{(j)}}\) in closed form, and its
exact factor is \(0.750\), equal to both bounds of \cref{prop:partition},
while the plug-in estimate is \(1.15\).  The trace ratios of both laws lie
below the upper bound at every budget.

\begin{table}[!htbp]
\centering
\footnotesize
\setlength{\tabcolsep}{3pt}
\caption{N1, four equal classes (\(k=24\), \(d=3\)).  The instance optimum is
exact.  The last column gives the cross-fitted estimate on the two halves of
the draws.}
\label{tab:rev-N1}
\begin{tabular}{rrrrlrl}
\toprule
$m$ & $M$ & $h_{21,M}$ & $\gamma^*$ (block law) & Thm.~1 law: $\hat\lambda_{\max}$ [90\% boot.] & trace & cross-fit (post hoc) \\
\midrule
4 & 1 & 1.0000 & 0.2500 & 0.2916 [0.3038, 0.3232] & 0.2500 & 0.2375, 0.2337 \\
5 & 2 & 0.4750 & 0.2125 & 0.2484 [0.2589, 0.2764] & 0.2125 & 0.2086, 0.2043 \\
6 & 3 & 0.3000 & 0.1750 & 0.2055 [0.2150, 0.2278] & 0.1762 & 0.1780, 0.1690 \\
8 & 5 & 0.1600 & 0.1000 & 0.1387 [0.1430, 0.1529] & 0.1171 & 0.1127, 0.1181 \\
10 & 7 & 0.1000 & 0.0750 & 0.0955 [0.0996, 0.1056] & 0.0809 & 0.0811, 0.0796 \\
12 & 9 & 0.0667 & 0.0500 & 0.0675 [0.0705, 0.0752] & 0.0575 & 0.0557, 0.0569 \\
16 & 13 & 0.0308 & 0.0250 & 0.0341 [0.0352, 0.0375] & 0.0284 & 0.0285, 0.0298 \\
20 & 17 & 0.0118 & 0.0100 & 0.0133 [0.0139, 0.0149] & 0.0115 & 0.0112, 0.0114 \\
\bottomrule
\end{tabular}

\end{table}

\begin{table}[!htbp]
\centering
\footnotesize
\setlength{\tabcolsep}{2.5pt}
\caption{N2, unequal classes (\(k=24\), \(d=3\)): bounds of
\cref{prop:partition}, and plug-in eigenvalue and trace ratio of two laws.}
\label{tab:rev-N2}
\resizebox{\textwidth}{!}{\begin{tabular}{rrrrlrlr}
\toprule
$m$ & lower & upper & $h_{21,M}$ & block: $\hat\lambda_{\max}$ & trace & Thm.~1: $\hat\lambda_{\max}$ & trace \\
\midrule
4 & 0.7500 & 0.7500 & 1.0000 & 1.1539 [0.9744, 2.3073] & 0.6618 & 1.5151 [1.0111, 3.0299] & 0.6606 \\
5 & 0.3314 & 0.3529 & 0.4750 & 0.5605 [0.4361, 1.0807] & 0.3053 & 0.5034 [0.4223, 0.8932] & 0.3134 \\
6 & 0.1918 & 0.2206 & 0.3000 & 0.2411 [0.2405, 0.3012] & 0.1859 & 0.3457 [0.2634, 0.6099] & 0.2064 \\
8 & 0.0975 & 0.1239 & 0.1600 & 0.1322 [0.1337, 0.1955] & 0.0998 & 0.1870 [0.1463, 0.3047] & 0.1127 \\
12 & 0.0237 & 0.0568 & 0.0667 & 0.0583 [0.0570, 0.0629] & 0.0368 & 0.0588 [0.0503, 0.0760] & 0.0404 \\
16 & 0.0000 & 0.0276 & 0.0308 & 0.0293 [0.0264, 0.0322] & 0.0130 & 0.0157 [0.0162, 0.0179] & 0.0132 \\
20 & 0.0000 & 0.0110 & 0.0118 & 0.0101 [0.0096, 0.0117] & 0.0032 & 0.0042 [0.0038, 0.0050] & 0.0029 \\
\bottomrule
\end{tabular}
}
\end{table}

\begin{figure}[!htbp]
  \centering
  \includegraphics[width=\textwidth]{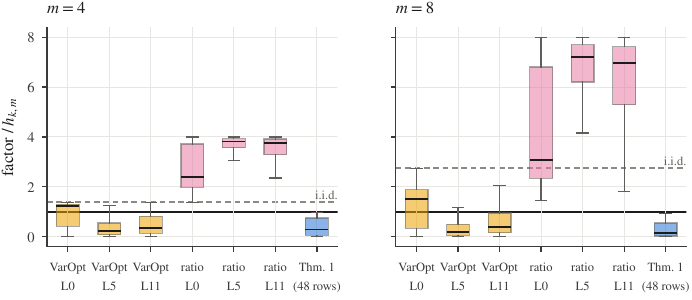}
  \caption{\textbf{GPT-2 attention rows} (\(k=12\), \(720\) rows).  Factor
  relative to \(h_{k,m}\) by layer for VarOpt and the ratio design (exact),
  and for the cross-fitted law of \cref{thm:main} on \(48\) rows.  The solid
  line is the curve and the dashed line is independent sampling.}
  \label{fig:gpt2}
\end{figure}

\paragraph{Classical designs (E4).}
\Cref{tab:e4a} reports the medians behind \cref{fig:numerics}(c) for both
budgets.  On GPT-2 small, the attention row of the twelfth token over its
twelve positions was taken in layers \(0\), \(5\) and \(11\), all heads, and
twenty fixed sentences, giving \(720\) rows (\cref{fig:gpt2},
\cref{tab:e4b}).  Rows in layers \(5\) and \(11\) are dominated by the first
position, with median weight \(0.79\), and those in layer \(0\) are more
spread.  VarOpt exceeds the curve on \(33\%\) of the rows at \(m=4\) and on \(32\%\) at \(m=8\), with median factor \(0.44\) and \(0.41\).  On the first \(48\) rows the cross-fitted law of \cref{thm:main} has median factor \(0.29\) at \(m=4\) and \(0.14\) at \(m=8\).  It lies below VarOpt on \(94\%\) and \(90\%\) of these rows.

\begin{table}[!htbp]
\centering
\footnotesize
\setlength{\tabcolsep}{4pt}
\caption{Median factor relative to \(h_{k,m}\) on \(p=\mathrm{softmax}(\beta z)\),
\(k=12\), for VarOpt and the ratio design (exact, \(20\) draws per \(\beta\))
and the cross-fitted law of \cref{thm:main} (\(5\) draws).}
\label{tab:e4a}
\begin{tabular}{rrrrrrr}
\toprule
& \multicolumn{3}{c}{$m=4$ (i.i.d.\ $1.38$)} & \multicolumn{3}{c}{$m=8$ (i.i.d.\ $2.75$)} \\
\cmidrule(lr){2-4}\cmidrule(lr){5-7}
$\beta$ & VarOpt & ratio & Thm.~1 & VarOpt & ratio & Thm.~1 \\
\midrule
0.0 & 1.00 & 1.00 & 0.98 & 1.00 & 1.00 & 0.97 \\
0.5 & 1.29 & 1.76 & 0.91 & 2.05 & 1.99 & 0.93 \\
1.0 & 1.32 & 2.51 & 0.72 & 1.56 & 3.30 & 0.52 \\
1.5 & 1.04 & 3.05 & 0.63 & 0.85 & 4.52 & 0.31 \\
2.0 & 0.78 & 3.39 & 0.37 & 0.45 & 5.61 & 0.20 \\
2.5 & 0.26 & 3.82 & 0.39 & 0.08 & 7.21 & 0.11 \\
3.0 & 0.34 & 3.72 & 0.04 & 0.06 & 6.78 & 0.03 \\
\bottomrule
\end{tabular}

\end{table}

\begin{table}[!htbp]
\centering
\footnotesize
\setlength{\tabcolsep}{4pt}
\caption{GPT-2 attention rows: median factor relative to \(h_{k,m}\) and the
share of rows where VarOpt exceeds the curve.}
\label{tab:e4b}
\begin{tabular}{lrrrrrr}
\toprule
& \multicolumn{3}{c}{$m=4$} & \multicolumn{3}{c}{$m=8$} \\
\cmidrule(lr){2-4}\cmidrule(lr){5-7}
layer & VarOpt & VarOpt$>1$ & ratio & VarOpt & VarOpt$>1$ & ratio \\
\midrule
0 & 1.24 & 68\% & 2.40 & 1.53 & 62\% & 3.08 \\
5 & 0.25 & 11\% & 3.82 & 0.19 & 10\% & 7.19 \\
11 & 0.36 & 21\% & 3.76 & 0.39 & 25\% & 6.96 \\
\midrule
Thm.~1, 48 rows & \multicolumn{3}{c}{0.29} & \multicolumn{3}{c}{0.14} \\
\bottomrule
\end{tabular}

\end{table}

\paragraph{Protocol outcomes.}
Protocol 1 predicted that the block law of N1 equals the instance optimum
(confirmed), that the plug-in factor of the law of \cref{thm:main} stays
below the curve within bootstrap error (confirmed for N3, not for N1 at
\(m=12,16,20\), where the curve lies between the plug-in and cross-fitted
estimates and every certificate holds), that the naive rule and the flight
exceed the curve (confirmed), and a bracket for N2 (not assessable with the
plug-in estimate, see above).  Protocol 2 predicted a cross-fitted factor of at most \(1.1\) for the law of \cref{thm:main} (largest value \(1.06\) on the synthetic weights and \(0.97\) on the GPT-2 rows, confirmed), that VarOpt exceeds the curve on a majority of the GPT-2 rows at \(m=8\) (\(32\%\), not confirmed), that the ratio design lies above VarOpt at \(\beta=3\) (confirmed), and the identity for independent sampling (confirmed).

\FloatBarrier
\Needspace{8\baselineskip}
\section{Protection on attention rows of a language model}
\label{app:llm}

This appendix measures what protection buys on real attention, in five
experiments with protocols written before their runs (protocols 3 to 7,
after protocols 1 and 2 of \cref{app:numerics-revision}).
\Cref{app:llm-512} compares certified bounds for the protected law with the
actual errors of classical sparsifiers at length \(512\).
\Cref{app:llm-64} runs the samplers themselves at length \(64\) and measures
their errors, \cref{app:llm-tails} examines tails, cost, the instance
optimum, and the protection of value directions, \cref{app:llm-fast}
evaluates the fast sampler of \cref{prop:fast}, and \cref{app:llm-downstream}
measures the effect on the output of a long-context model.

\subsection{Certificates at length 512}
\label{app:llm-512}

We run Qwen2.5-0.5B (fp32, eager attention) on eight disjoint chunks of 512 tokens
of the WikiText-2 test split and take the attention row \(p\) of the last
position in every layer and head, which gives 2688 rows with \(k=512\).  The
response is the value matrix \(V\in\R^{64\times512}\) of the head, so the
error is \(\E\|V(Q-p)\|_2^2\).  VarOpt is the fixed-size threshold design
that keeps the heavy tokens and draws the light ones by systematic sampling,
and its error is the mean over \(200\) draws.

\begin{figure}[!ht]
  \centering
  \includegraphics[width=\textwidth]{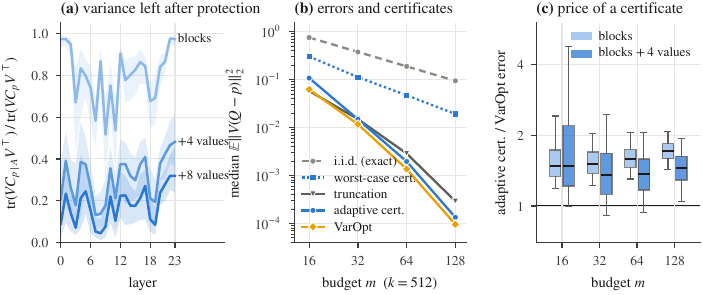}
  \caption{\textbf{Protection on Qwen2.5-0.5B attention, \(k=512\), 2688 rows.}
  (a)~Share of the value variance \(\tr(V\Cp V^\top)\) that remains in the
  comparator after protecting the three block masses, and the block masses
  together with four or eight value directions (median and interquartile band
  per layer).  (b)~Median errors of i.i.d.\ sampling, top-\(m\) truncation and
  VarOpt, and two certified bounds for the protected law: the worst-case
  coefficient of \cref{thm:main} and the instance-adaptive certificate of
  \cref{cor:heavy} with the heaviest tokens kept exactly.
  (c)~Adaptive certificate divided by the actual VarOpt error, per row (boxes
  are quartiles, whiskers the 5th and 95th percentiles).}
  \label{fig:llm}
\end{figure}

\paragraph{What protection removes.}
Protecting the three block masses (the attention sink, the 32 most recent
tokens and the rest, \(d=2\)) leaves a median \(0.83\) of
\(\tr(V\Cp V^\top)\) in \(\tr(V\Cpa V^\top)\).  Adding the top four
eigenvectors of \(V\Cp V^\top\) as proxy features, as in \cref{sec:setup},
leaves a median \(0.31\), and eight directions leave \(0.18\)
(\cref{fig:llm}(a)).

\paragraph{Worst-case certificate against actual errors.}
The certificate \(h_{k-d,m-d}\tr(V\Cpa V^\top)\) with the block masses and
four value directions protected is below the exact i.i.d.\ error on every
row for \(m\ge32\).  It is loose on these rows, however.  Real attention is
peaked, and VarOpt, which keeps heavy tokens exactly, has an actual error 28
times smaller than this certificate at \(m=64\) (median over rows).  The
coefficient of \cref{thm:main} is forced by instances that spread the
residual mass evenly over \(k-d\) coordinates, while a real row puts most of
its mass on a few tokens.

\paragraph{Instance-adaptive certificate.}
Keeping the \(j\) heaviest tokens exactly spends one atom on each and removes
their variance.  \Cref{cor:heavy} then gives
\(\Cov(Q)\psdle\pi_Lh_{k-j-d_L,m-j-d_L}\Cpa\), where \(\pi_L\) is the mass of
the remaining tokens \(L\), and on the value response the sharper bound
\[
 \E\|V(Q-p)\|_2^2\le\pi_L^2\,h_{k-j-d_L,\,m-j-d_L}\,
 \tr\bigl(V_LC_{p_L\mid A_L}V_L^\top\bigr),
\]
where \(V_L\) is \(V\) restricted to \(L\).  We choose \(j\) per row to
minimize the factor \(\pi_Lh_{k-j-d_L,m-j-d_L}\), which depends on \(p\) and
\(A\) only, so the law does not depend on the response.  With the block
masses protected, this factor is a median \(2.8\) to \(15\) times below the
worst case \(h_{k-d,m-d}\) as \(m\) grows from \(16\) to \(128\), and the
certified value error is within a median factor \(1.48\) to \(1.71\) of the
actual VarOpt error.  With four value directions as well, the median factor
is \(1.36\) to \(1.49\), and the certificate is below the actual VarOpt error
on \(4\%\) to \(11\%\) of the rows (\cref{fig:llm}(c)).  At \(m=64\) this
closes the gap of the worst-case certificate from a median factor \(28\) to
\(1.37\).  Choosing \(j\) for the total variance \(\tr\Cov(Q)\) instead gives
\(1.31\) to \(1.35\) with the block masses.  The certificate holds for every
response at once and the law keeps the block masses on every draw, whereas
the VarOpt error is measured on this one response and VarOpt does not keep
the block masses.  The sampler of \cref{thm:main} is not run at \(k=512\)
here, so all values reported for the protected law are certified upper
bounds.

\paragraph{Outcomes of protocol 3.}
Protocol 3 predicted that the block masses with four value directions leave
at most half of \(\tr(V\Cp V^\top)\) in the median (\(0.31\), confirmed), that
the worst-case certificate lies below the i.i.d.\ error on at least \(75\%\)
of the rows at \(m=64\) (every row, confirmed), that it lies below the actual
VarOpt error on at least half of the rows at \(m=64\) (\(1.7\%\), not
confirmed), and that the coefficient of \cref{thm:main} is at most \(9/20\)
of the bound of \citet{haverbeck2026risk} under the same protection (largest
ratio \(0.36\), an identity check of \cref{cor:haverbeck}).  The
instance-adaptive certificate was defined after the third outcome and is
evaluated on the same rows.  The first two versions chose \(j\) with the value
matrix itself, and the rule above was fixed before it was evaluated.

\subsection{Measured errors at length 64}
\label{app:llm-64}

\begin{figure}[!ht]
  \centering
  \includegraphics[width=\textwidth]{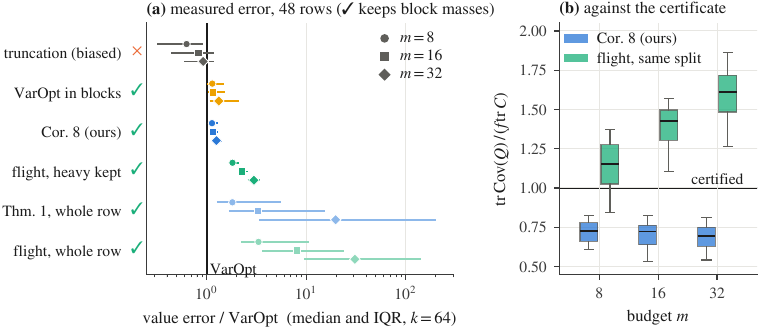}
  \caption{\textbf{Measured errors on Qwen2.5-0.5B attention, \(k=64\), 48
  rows.}  (a)~Value error relative to VarOpt (median and interquartile range)
  for \(m=8,16,32\).  A check mark means that every block mass is kept on every
  draw.  VarOpt moves the block masses by a median \(0.040\), \(0.011\) and
  \(0.0003\) in root mean square, and truncation by \(0.092\), \(0.022\) and
  \(0.0047\).  (b)~Unbiased estimate of the population trace ratio
  \(\tr\Cov(Q)/(f\tr C)\), which bounds the Loewner factor from below
  relative to the certified factor \(f\) of \cref{cor:heavy},
  for the law of \cref{thm:main} and for a random flight with the same
  protection and heavy tokens (boxes are quartiles, whiskers the 5th and 95th
  percentiles).}
  \label{fig:e6}
\end{figure}

The certificates of \cref{app:llm-512} bound the error of the protected law.
To measure the error itself, protocol 4 runs the samplers on rows short
enough for the real-arithmetic construction of \cref{sec:numerics}.  It uses
the
last-position rows of eight new chunks of \(64\) tokens, draws \(48\) rows
with two per layer, protects the three block masses (the sink, the last \(8\)
tokens and the rest), and takes \(m\in\{8,16,32\}\).  Each path law is run
\(64\) times, and its covariance is estimated without bias by averaging the
sums of exact step covariances over these paths.  Every one-step factor of the
minimax steps is at most \(1.0004\).  \Cref{fig:e6} reports the results.

With the heaviest tokens kept exactly as in \cref{cor:heavy} and \(j\) chosen
as above, the law of \cref{thm:main} keeps every block mass on every draw at
a measured value error \(1.13\), \(1.15\) and \(1.25\) times that of VarOpt
(medians for \(m=8,16,32\)), and its measured error lies below its
certificate on every row, with median ratio \(0.72\).  The same law on the
whole row has \(1.8\) to \(20\) times the VarOpt error, so keeping the heavy
tokens matters.  A cube-style random flight with the same protection and the
same heavy tokens has \(1.6\) to \(2.4\) times the error of the minimax step.
Its estimated trace ratio exceeds one in \(130\) of the \(144\) cases
(about \(90\%\)), providing empirical evidence that the flight can exceed
the certified factor.  The estimate for the law of \cref{thm:main} never
exceeds \(0.89\).  VarOpt run inside each block
with the allocation of \cref{prop:partition} also keeps the block masses and
has a similar median error with a wider spread.  Truncation, which is biased
and moves block mass, has a smaller median error on these short rows
(\(0.63\) to \(0.92\) times VarOpt), whereas at \(k=512\) its error is
\(1.4\) to \(3.2\) times that of VarOpt for \(m\ge32\).

\paragraph{Outcomes of protocol 4.}
Protocol 4 predicted that the measured error of the law of \cref{thm:main}
lies below its certificate (confirmed on every case), that keeping the heavy
tokens lowers it on at least \(75\%\) of the rows at each budget (\(88\%\),
\(92\%\) and \(100\%\), confirmed), that the random flight has a larger median
error (\(1.95\) times, confirmed), that the price of protection is at most
\(1.5\) times the VarOpt error in median at each budget (confirmed), and that
VarOpt inside blocks is
within a factor \(1.25\) of the law of \cref{thm:main} in median
(confirmed).  It also predicted that every protected method keeps the block
masses (largest deviation \(2\cdot10^{-15}\)) while VarOpt and truncation move
them in every case.  This failed for VarOpt on \(17\) of the \(144\) cases,
rows so peaked that every token it randomizes carries less than \(10^{-12}\)
of mass.  A prediction on the largest generalized eigenvalue could not be
assessed, because its plug-in estimate is dominated by noise in directions of
tiny comparator variance, with attention weights down to \(10^{-34}\).  The
trace-ratio estimate, added to the protocol after a pilot on separate rows
and before the evaluation rows were run, is reported instead (prediction
confirmed by the point estimates).

\subsection{Tails, cost, the instance optimum, and value directions}
\label{app:llm-tails}

The first three analyses below reuse the rows and results of protocol 4 and
were defined after its run.  The fourth follows protocol 5, written before
its run.

\begin{figure}[!ht]
  \centering
  \includegraphics[width=\textwidth]{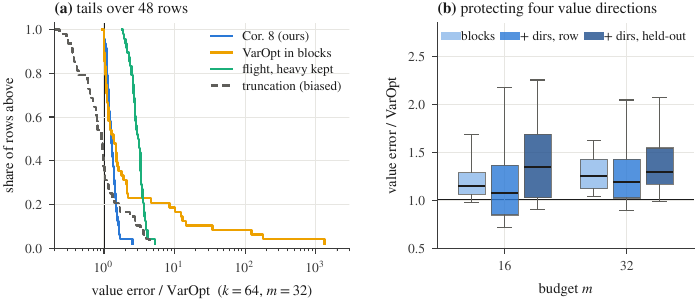}
  \caption{\textbf{Tails and value directions on Qwen2.5-0.5B attention,
  \(k=64\), 48 rows.}  (a)~Share of the rows whose value error exceeds a given
  multiple of the VarOpt error, at \(m=32\).  (b)~Value error relative to
  VarOpt with the block masses protected (protocol 4) and with four value
  directions added (protocol 5), taken from the row itself or estimated on
  held-out text (boxes are quartiles, whiskers the 5th and 95th
  percentiles).}
  \label{fig:tails}
\end{figure}

\paragraph{Tails.}
The protected law and VarOpt inside the blocks have close median errors, but
their worst rows differ (\cref{fig:tails}(a)).  Over the 48 rows the value
error of the protected law is at most \(2.8\), \(1.8\) and \(2.6\) times that
of VarOpt for \(m=8,16,32\).  VarOpt inside the blocks reaches \(26\), \(19\)
and \(1330\) times, with 90th percentiles \(2.9\), \(3.1\) and \(13\).  Its
allocation suits weights that are uniform inside each block, and on these rows
it gives a block fewer atoms than VarOpt would keep exactly, so heavy tokens
are randomized.  The ratios are large because VarOpt is almost exact on these
rows: there the error of VarOpt inside the blocks stays below \(5\%\) of the
i.i.d.\ error.  On these 48 rows, the measured error of the protected law
stays within the reported factors of VarOpt, while the cheap design has a
longer relative-error tail.

\paragraph{Cost.}
Each draw of the protected law takes a median \(1.5\) seconds at \(k=64\) on
one CPU core in our implementation, with one minimax step per removed
coordinate and \(60\) rounds of matrix multiplicative weights per step.  The
random flight takes \(0.01\) seconds and VarOpt microseconds.  A fast minimax
step is the main obstacle to use at attention scale.

\paragraph{The instance optimum.}
Block protection is a partition, so \cref{prop:partition} brackets the
instance-optimal constant on every row, and the factor of \cref{cor:heavy} is
a further upper bound.  At \(m=8\) the lower bound is informative on \(90\%\)
of the rows, the bracket is within a median factor \(2\), and the trace of the
measured covariance of the protected law is within a median factor \(1.11\)
of the lower bound.  The lower bound vanishes once the budget covers the
effective number of tokens \(1/\|p^{(j)}\|_2^2\) of every block, which happens
on \(42\%\) of the rows at \(m=16\) and \(94\%\) at \(m=32\).  Lower bounds for
peaked weights remain open.

\paragraph{Protecting value directions (protocol 5).}
Protocol 5 adds four value directions to the protected block masses at
\(m\in\{16,32\}\), taken either from the row itself or from the same head on
a held-out chunk, and runs the protected law and the random flight with
\(32\) paths per row.  It predicted that the row's own directions bring the
median value error to at most the VarOpt error at \(m=32\) (\(1.19\) times,
not confirmed) and to at most \(1.25\) times at \(m=16\) (\(1.07\) times,
confirmed), that they improve on the block masses alone on at least \(75\%\)
of the rows at \(m=32\) (\(67\%\), not confirmed), and that held-out
directions improve on the block masses alone in median at \(m=32\) (\(1.05\)
times their error, not confirmed).  It also predicted that the random flight
has a larger error (\(1.66\) times in median, confirmed), that protection is
exact (largest relative deviation \(4\cdot10^{-16}\), confirmed), and that the
measured error lies below the certificate (every case, confirmed).  Each
protected direction costs one atom, and at these budgets the variance that
four directions remove barely pays for the four atoms, while directions
estimated on other text do not pay at all (\cref{fig:tails}(b)).

\subsection{The fast sampler (protocol 6)}
\label{app:llm-fast}

Protocol 6 runs the sampler of \cref{prop:fast} on the \(48\) rows of
\cref{app:llm-64} at \(k=64\), with \(4000\) draws per row and budget, and on
\(384\) rows at \(k=512\) (heads 0 and 7 of every layer on the eight chunks of
\cref{app:llm-512}, block masses protected with a window of \(32\)), with
\(2000\) draws each, together with VarOpt and VarOpt inside the blocks.  The
heavy set takes the \(j\) heaviest tokens, with \(j\) chosen to minimize the
factor of \cref{prop:fast}, plus any token whose inclusion probability would
exceed one.  The light budget is split over the blocks in proportion to their
light mass, capped by block size, and each block is sampled by pivotal
sampling (dependent rounding) in a random order.  The code was developed on
the two pilot rows of \cref{app:llm-64} only.

\begin{figure}[!ht]
  \centering
  \includegraphics[width=\textwidth]{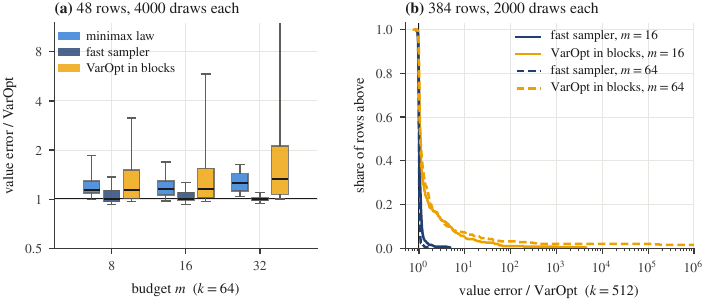}
  \caption{\textbf{The fast sampler on Qwen2.5-0.5B attention.}  (a)~Value
  error relative to VarOpt at \(k=64\) for the minimax law of
  \cref{cor:heavy}, the fast sampler of \cref{prop:fast}, and VarOpt inside the
  blocks (boxes are quartiles, whiskers the 5th and 95th percentiles).
  (b)~Share of the \(384\) rows at \(k=512\) whose value error exceeds a given
  multiple of the VarOpt error.}
  \label{fig:fps}
\end{figure}

\paragraph{Outcomes of protocol 6.}
The protocol predicted that the Monte Carlo estimate of the trace ratio
\(\tr\Cov(Q)/(f\tr C_{p\mid A'})\) with \(f=2\max_b\pi_b/s_b\), twice the
factor of \cref{prop:fast}, is at most one on at least \(99\%\) of the cases
(every case, largest value \(0.47\), confirmed), that the fast sampler has a median value error at most that of
the minimax law at \(k=64\) (medians \(0.88\), \(0.91\) and \(0.80\) of it,
confirmed), and that its median value error is at most \(1.15\) times that of
VarOpt at every budget and length (approximately \(1.00\) across cases, confirmed).  It
predicted bounded tails (largest ratio to VarOpt \(2.0\), \(1.4\) and \(1.2\)
at \(k=64\), and 99th percentile at most \(1.9\) at \(k=512\), confirmed),
while VarOpt inside the blocks has a 99th percentile above \(50\) at every
budget at \(k=512\), confirmed.  It predicted block masses exact to
\(10^{-12}\) of the block mass, which failed on one block of mass
\(4\cdot10^{-14}\), where the floating-point error is \(5\cdot10^{-23}\), and
a support of at most \(m\) on every draw (confirmed).  It predicted at most
\(2\) milliseconds per draw at \(k=512\) in NumPy on one core (median
\(0.6\), confirmed).  The held-out response \(V'\) and the identity behave
like \(V\).  The fast sampler thus keeps the guarantee of a protected law,
has a median error close to that of VarOpt, and avoids both the cost of the
minimax steps and the tails of the cheap design.

\paragraph{Analyses defined after the run.}
Against the factor \(\max_b\pi_b/s_b\) of \cref{prop:fast} the estimated trace ratio is
at most \(0.89\) at \(k=64\) and \(0.94\) at \(k=512\).  This factor is a
median \(0.94\), \(1.10\) and \(1.36\) times the factor of \cref{cor:heavy} at
\(k=64\) and \(m=8,16,32\), and \(0.98\), \(1.02\), \(1.07\) and \(1.15\) times
it at \(k=512\) and \(m=16,32,64,128\), with a largest ratio of \(1.82\).
The setup chooses the heavy set and the budgets from \(p\) and \(A\) by
searching \(j\), and on \(48\) random rows at \(k=512\) it takes a median of
\(0.8\), \(2.5\), \(7.7\) and \(53\) milliseconds at \(m=16,32,64,128\) on one
core, against \(0.6\) milliseconds per draw.  When each row is sparsified
once, this setup dominates the cost at large \(m\).

\subsection{Downstream decoding (protocol 7)}
\label{app:llm-downstream}

Protocol 7 asks whether these differences reach the output of a long-context
model.  Qwen2.5-7B-Instruct \citep{yang2024qwen25} in bf16 receives prompts of
about \(16\)k tokens.  The prompt except its last token is prefilled with dense
attention.  From the last prompt token on, every attention row of every layer
and query head is replaced by the sparse weights of one method at budget
\(m\in\{64,256\}\), which is \(0.4\%\) and \(1.6\%\) of the context, and the
head output is computed from these weights and the values.  The methods are
dense attention, top-\(m\) truncation, the first \(4\) and the last \(m-4\)
positions (sink and window), \(m\) independent draws, VarOpt with systematic
sampling, VarOpt with ordered pivotal sampling (one block), and the fast sampler of
\cref{prop:fast} with the first \(4\) positions, the last \(64\) positions and
the rest as blocks.  Its heavy set is initialized from the VarOpt heavy set
and adjusted to reserve an atom for every nonempty light block and keep
inclusion probabilities feasible.  This rule is cheaper than the search
over \(j\) of protocol 6 and was fixed before the runs.  A minimal
implementation of the model matches the reference implementation to
\(7\cdot10^{-4}\) in the logits in float32 on Qwen2.5-0.5B, and the thirteen
configurations decode in one batch from the same prefill.

The tasks are the six RULER tasks \citep{hsieh2024ruler} that need no external
corpus, generated by its own scripts at length \(16384\) with seed \(42\):
retrieval of one needle among noise (S1) and among other needles (MK2, MK3),
variable tracking (VT), and common and frequent word extraction (CWE, FWE).
Answers are decoded greedily and scored by the string match of RULER.
Perplexity is measured on the first \(12\) books of the PG-19 test split
\citep{rae2019compressive} with at least \(16384\) tokens, on \(256\) tokens
after a context of \(16128\).  The protocol fixed the number of RULER samples
from timing alone.  The projected cost of \(50\) samples per task exceeded
three GPU hours, so each task has \(30\).  One crash before any output, a
synchronization call in the driver that called itself, was fixed and the run
restarted, as recorded in the protocol.

\begin{table}[!ht]
\centering
\footnotesize
\caption{\textbf{Protocol 7: RULER scores and PG-19 perplexity of
Qwen2.5-7B-Instruct with sparse attention rows at decoding time.}  Scores are
in percent over \(30\) samples per task, RULER is their mean over the six
tasks, and PPL is the perplexity over \(12\) books.}
\label{tab:e9}
\begin{tabular}{llrrrrrrrr}
\toprule
Method & $m$ & S1 & MK2 & MK3 & VT & CWE & FWE & RULER & PPL \\
\midrule
dense & all & 76.7 & 93.3 & 66.7 & 78.7 & 67.0 & 80.0 & 77.1 & 13.12 \\
\midrule
top-$m$ & 64 & 80.0 & 86.7 & 73.3 & 84.0 & 66.0 & 77.8 & 78.0 & 13.83 \\
sink+window & 64 & 0.0 & 0.0 & 0.0 & 0.0 & 1.3 & 0.0 & 0.2 & 18.63 \\
i.i.d. & 64 & 73.3 & 86.7 & 56.7 & 87.3 & 64.0 & 78.9 & 74.5 & 18.94 \\
VarOpt & 64 & 73.3 & 93.3 & 53.3 & 80.7 & 73.3 & 72.2 & 74.4 & 13.70 \\
pivotal, no blocks & 64 & 76.7 & 93.3 & 46.7 & 82.0 & 66.3 & 77.8 & 73.8 & 13.60 \\
\cref{prop:fast}, blocks & 64 & 76.7 & 93.3 & 60.0 & 80.7 & 67.0 & 76.7 & 75.7 & 13.83 \\
\midrule
top-$m$ & 256 & 80.0 & 90.0 & 70.0 & 83.3 & 75.7 & 76.7 & 79.3 & 13.20 \\
sink+window & 256 & 3.3 & 3.3 & 0.0 & 2.0 & 12.3 & 65.6 & 14.4 & 15.00 \\
i.i.d. & 256 & 80.0 & 90.0 & 60.0 & 82.0 & 69.3 & 80.0 & 76.9 & 13.35 \\
VarOpt & 256 & 76.7 & 93.3 & 63.3 & 78.7 & 74.0 & 73.3 & 76.6 & 13.22 \\
pivotal, no blocks & 256 & 76.7 & 93.3 & 50.0 & 76.7 & 61.7 & 78.9 & 72.9 & 13.18 \\
\cref{prop:fast}, blocks & 256 & 76.7 & 93.3 & 53.3 & 77.3 & 72.0 & 77.8 & 75.1 & 13.14 \\
\bottomrule
\end{tabular}

\end{table}

\begin{figure}[!ht]
  \centering
  \includegraphics[width=\textwidth]{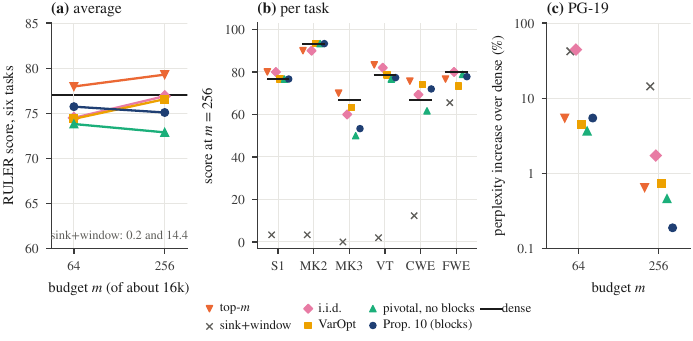}
  \caption{\textbf{Protocol 7.}  (a)~RULER score averaged over six tasks.
  Sink and window score \(0.2\) and \(14.4\).  (b)~Score per task at
  \(m=256\).  (c)~Increase of the PG-19 perplexity over dense attention.}
  \label{fig:e9}
\end{figure}

\paragraph{Outcomes of protocol 7.}
Dense attention averages \(77.1\) on the six tasks, above the predicted floor
of \(75\).  The protocol predicted that the fast sampler beats top-\(m\)
truncation on the aggregation tasks CWE and FWE at \(m=256\), where top-\(k\)
degrades in the experiments of \citet{chen2024magicpig}.  This failed
(difference \(-1.3\), \(95\%\) bootstrap interval \([-7.4,4.9]\)).  It
predicted that the fast sampler is at most two points below VarOpt on RULER
at both budgets, which holds for the point estimates (\(+1.4\) at \(m=64\)
and \(-1.5\) at \(m=256\)), whose intervals \([-2.8,5.6]\) and
\([-4.5,1.4]\) reach below \(-2\).  It predicted that the fast sampler is at least as good as
independent draws on RULER and in perplexity at both budgets.  This failed on
RULER at \(m=256\) (\(-1.8\), \([-5.5,1.7]\)), while the perplexity is lower at
both budgets, by \(0.31\) nats per token at \(m=64\) (\([0.19,0.44]\)) and by
\(0.015\) at \(m=256\) (\([-0.0003,0.030]\)).  Sink and window fail on
the three needle tasks, with \(2.2\) on average at \(m=256\) (confirmed).
At \(m=256\) the fast sampler has the smallest perplexity of the six
sparse methods, \(0.2\%\) above dense, against \(0.5\%\) for pivotal sampling
without blocks, \(0.6\%\) for top-\(m\), \(0.7\%\) for VarOpt, \(1.7\%\) for
independent draws and \(14\%\) for sink and window (confirmed).

\paragraph{What the outcomes show.}
Top-\(m\) truncation is the most accurate sparse method on these RULER tasks,
\(2.2\) and \(4.2\) points above the fast sampler at \(m=64\) and \(m=256\)
(intervals \([-2.7,7.5]\) and \([0.1,8.6]\)), and unbiasedness does not help
there.  The guarantees of this paper concern responses linear in \(Q\),
whereas the logits depend on the attention outputs of all layers through
nonlinear maps, so unbiased rows do not give unbiased logits, and their
variance reaches every later layer.  In perplexity, variance matters.
Independent draws, whose covariance is the with-replacement benchmark of
\cref{prop:fast}, raise the perplexity by \(44\%\) at \(m=64\), against
\(3.6\%\) to \(5.4\%\) for top-\(m\) and the three designs without
replacement.  At \(m=256\) every method except sink and window is within
\(1.8\%\) of dense.  Protecting the block masses has no detectable cost against
VarOpt in either measure.  Pivotal sampling without blocks is \(3.7\) points
below VarOpt on RULER at \(m=256\) (\([-6.8,-1.0]\)), mostly on MK3 and CWE,
which the protocol did not predict.  A decoding step of the thirteen
configurations takes about \(0.9\) seconds in our implementation, so the
protocol measures accuracy, not speed.  SANTA \citep{lee2026santa} samples
value rows from exact scores as here and reports up to a \(1.5\)-fold faster
decoding attention kernel at \(32\)k tokens.  Its basic estimator is the mean
of independent draws, which raises the perplexity by \(44\%\) at \(m=64\)
here.

\paragraph{Analyses defined after the run.}
With a paired bootstrap over the \(12\) books (\(10^4\) resamples), the
\(95\%\) intervals of all pairwise perplexity differences among top-\(m\),
independent draws, VarOpt, pivotal sampling without blocks and the fast
sampler contain zero at \(m=256\), so these five methods cannot be told apart
there.  Against dense attention the intervals exclude zero for VarOpt and for
pivotal sampling without blocks and contain it for the other three.  Top-\(m\)
is \(2.2\) points above dense attention on RULER at \(m=256\), with interval
\([-2.2,6.6]\).  MK3 asks to copy a UUID.  There, at \(m=256\), the fast
sampler answers \(4\) of \(30\) questions fewer than dense attention
(\(-13.3\) points, \([-26.7,-3.3]\)) and pivotal sampling without blocks \(5\)
fewer, while top-\(m\) answers one more, and at \(m=64\) every sampling design
loses \(6.7\) to \(20\) points on MK3.  This is one task, without a correction
for multiple comparisons.

\paragraph{Scope.}
The protocol uses one model, one context length and \(30\) samples per task.
Each sample is decoded once, so the randomness of the samplers is part of the
sampling error, and only the decoding steps are sparsified.

\end{document}